\documentclass[10pt]{article} % For LaTeX2e
\PassOptionsToPackage{table}{xcolor}
\usepackage[accepted]{tmlr}
\usepackage[english]{babel}

\usepackage{amsmath, amssymb, mathtools, bbm}

\usepackage{graphicx}       % For including images
\usepackage{caption}        % For figure captions
\usepackage{subcaption}     % For subfigures
\usepackage{wrapfig}        % For wrapping text around figures
\usepackage{rotating}       % For sideways figures
\usepackage{adjustbox}      % For scaling tables or figures
\usepackage{makecell}       % For multi-line cells
\usepackage{array, booktabs} % For professional tables

\usepackage[table]{xcolor}
\definecolor{LightGray}{gray}{0.85} % light gray for lines
\definecolor{TagColor}{HTML}{1DAD3F} % tag color for metric box

\usepackage{listings}

\definecolor{myref}{HTML}{DB1604}
\definecolor{mycite}{HTML}{0B8C0B}
\usepackage[
colorlinks=true,
citecolor=mycite, linkcolor=myref
]{hyperref}
\usepackage{url}

\usepackage{enumitem}
\setlist[itemize]{noitemsep, topsep=5pt}

\usepackage{pdflscape}

\usepackage{titlesec}

\usepackage[most]{tcolorbox}
\usepackage{needspace}

\newcounter{metriccounter}
\newcommand{\MetricPrefix}{Default} % Default prefix if not set manually
\newcommand{\setMetricPrefix}[1]{\renewcommand{\MetricPrefix}{#1}}

\newtcolorbox[auto counter]{Metric}[6]{%
    breakable, %  this line allows the box to break across pages
    enhanced jigsaw, %
    colback=gray!05, % Background color
    colframe=gray!30, % Border color
    coltitle=black, % Title color
    fonttitle=\bfseries, % Title font
    title={\makebox[\linewidth][l]{Metric \MetricPrefix.\number\numexpr\value{metriccounter}+1\relax: #2 \hfill }}, % Metric number and title
    before upper={%
        \refstepcounter{metriccounter}
         \IfValueTF{#1}{%
            \ifstrequal{#1}{nobreak}{\needspace{0\baselineskip}}{\Needspace{8\baselineskip}}%
        }{}%
        \textbf{Desiderata:} #3 \\
        \textbf{Explanation Type:} #4 \hfill \textbf{References:} (#5)\\
        {\footnotesize  #6 }       
        \vspace*{-5pt} % space before main content

        \makebox[\linewidth][c]{\color{gray!20}\rule{\linewidth + 1cm}{1.5pt}\color{black} }% horizontal line divider matching box border
        
        \vspace*{2pt} % space before main content
    },
}

\providecommand{\metref}[1]{%
    \hyperref[#1]{Metric~\ref*{#1}}}

\newtcolorbox{terminologybox}[1]{
  colback=gray!10,
  colframe=gray!70,
  boxrule=0.5pt,
  arc=4pt,
  left=4pt,
  right=4pt,
  top=2pt,
  bottom=2pt,
  fonttitle=\bfseries,
  title=#1,
  before upper={%
    \hspace*{-0.25cm}
    }
}

\newtcolorbox{definitionbox}[1]{
  colback=gray!10,
  colframe=gray!50,
  boxrule=0.5pt,
  arc=4pt,
  left=4pt,
  right=4pt,
  top=2pt,
  bottom=2pt,
}

\newcommand{\citenote}[1]{\unskip\textsuperscript{[}\footnote{#1}\textsuperscript{]}}

\let\originalautoref\autoref
\renewcommand{\autoref}[1]{%
  \begingroup%
  \def\chapterautorefname{Chapter}%
  \def\sectionautorefname{Section}%
  \def\subsectionautorefname{Subsection}%
  \def\subsubsectionautorefname{Subsection}%
  \originalautoref{#1}%
  \endgroup%
}

\usepackage{tikz}
\usetikzlibrary{decorations.pathreplacing}
\usepackage{lmodern}

\newcommand{\lightcheckmark}{\textcolor{gray!90}{(\checkmark)}}

\newcommand{\repeatfootnote}[1]{%
  \textsuperscript{\hyperref[#1]{\ref*{#1}}}%
}

\usepackage[acronym]{glossaries}

\glsdisablehyper

\newacronym{vxai}{VXAI}{eValuation of Explainable Artificial Intelligence}
\newacronym{prisma}{PRISMA}{Preferred Reporting Items for Systematic Reviews and Meta-Analyses}

\newacronym[user1={def:FA}]{FA}{FA}{Feature Attribution}
\newacronym[user1={def:CE}]{CE}{CE}{Concept Explanation}
\newacronym[user1={def:ExE}]{ExE}{ExE}{Example Explanation}
\newacronym[user1={def:WBS}]{WBS}{WBS}{White-Box Surrogate}
\newacronym[user1={def:NLE}]{NLE}{NLE}{Natural Language Explanation}
\newacronym{aa}{AA}{Adversarial Attacks }

\newglossaryentry{explanandum}{
    name ={explanandum},
    plural = {explananda},
    description = {What is to be explained, i.e. a model and its prediction},
    user1={def:explan}
}
\newglossaryentry{explanation}{
    name ={explanation},
    plural = {explanations},
    description = {The process of explaining, i.e. the XAI algorithm},
    user1={def:explan}
}
\newglossaryentry{explanans}{
    name ={explanans},
    plural = {explanantia},
    description = {The explaining information, i.e. the output of an explanation},
    user1={def:explan}
}

\let\originalgls\gls
\let\originalglspl\glspl

\definecolor{GlossaryLinkColor}{HTML}{3D3D3D}
\usepackage{xspace}
\renewcommand*{\gls}[1]{%
  \ifglshasfield{user1}{#1}%
    {\protect\hyperlink{\glsentryuseri{#1}}{\textcolor{GlossaryLinkColor}{\originalgls{#1}}}}%
    {\originalgls{#1}}%
}
\renewcommand*{\glspl}[1]{%
  \ifglshasfield{user1}{#1}%
    {\protect\hyperlink{\glsentryuseri{#1}}{\textcolor{GlossaryLinkColor}{\originalglspl{#1}}}}%
    {\originalglspl{#1}\xspace}%
}

\title{Unifying VXAI: A Systematic Review and Framework\\for the Evaluation of Explainable AI}

\author{%
\name David Dembinsky \email david.dembinsky@dfki.de \\
\addr German Research Center for Artificial Intelligence (DFKI) GmbH\\
      RPTU University Kaiserslautern-Landau, Department of Computer Science
\AND
\name Adriano Lucieri \email adriano.lucieri@dfki.de \\
\addr German Research Center for Artificial Intelligence (DFKI) GmbH\\
      RPTU University Kaiserslautern-Landau, Department of Computer Science
\AND
\name Stanislav Frolov \email stanislav.frolov@dfki.de \\
\addr German Research Center for Artificial Intelligence (DFKI) GmbH\\
      RPTU University Kaiserslautern-Landau, Department of Computer Science
\AND
\name Hiba Najjar \email hiba.najjar@dfki.de \\
\addr German Research Center for Artificial Intelligence (DFKI) GmbH\\
      RPTU University Kaiserslautern-Landau, Department of Computer Science
\AND
\name Ko Watanabe \email ko.watanabe@dfki.de \\
\addr German Research Center for Artificial Intelligence (DFKI) GmbH\\
      RPTU University Kaiserslautern-Landau, Department of Computer Science
\AND
\name Andreas Dengel \email andreas.dengel@dfki.de \\
\addr German Research Center for Artificial Intelligence (DFKI) GmbH\\
      RPTU University Kaiserslautern-Landau, Department of Computer Science
}

\def\month{01}
\def\year{2026}
\def\openreview{\url{https://openreview.net/forum?id=wAvFLe7o0E}}

\newcommand{\vxailink}{\url{https://vxai.dfki.de/}}

\begin{document}
\maketitle

%%%%%%%%%%%%%%
%%

\begin{abstract}%
Modern AI systems frequently rely on opaque black-box models, most notably Deep Neural Networks, whose performance stems from complex architectures with millions of learned parameters.
While powerful, their complexity poses a major challenge to trustworthiness, particularly due to a lack of transparency.
Explainable AI (XAI) addresses this issue by providing human-understandable explanations of model behavior.
However, to ensure their usefulness and trustworthiness, such explanations must be rigorously evaluated.
Despite the growing number of XAI methods, the field lacks standardized evaluation protocols and consensus on appropriate metrics.
To address this gap, we conduct a systematic literature review following  the  \textit{Preferred Reporting Items for Systematic Reviews and Meta-Analyses (PRISMA)} guidelines and introduce a unified framework for the \textit{eValuation of XAI (VXAI)}.
We identify $362$ relevant publications and aggregate their contributions into $41$ functionally similar metric groups.
In addition, we propose a three-dimensional categorization scheme spanning explanation type, evaluation contextuality, and explanation quality desiderata.
Our framework provides the most comprehensive and structured overview of \glsentryshort{vxai} to date.
It supports systematic metric selection, promotes comparability across methods, and offers a flexible foundation for future extensions.

\end{abstract}
\section{Introduction}
\label{sec:intro}

% Importance of XAI
Explainable AI (XAI) is a research area of growing interest to both AI researchers and practitioners.
It aims to alleviate the black-box issue of current deep-learning models, which can reach stunning performances at the expense of their interpretability \citep{vilone2021notions}. 
Government-affiliated initiatives, such as the \citet{european2019ethics}, the \citet{nist2023airmf}, and the DARPA initiative \citep{gunning2019darpa}, identified XAI as a crucial part of Trustworthy AI.
Especially as it helps AI systems in serving the ``right to explain'' its decisions \citep{goodman2017european} and fosters user trust through understanding of the system \citep{morandini2023examining}.
XAI already plays a fundamental role in making high-stakes AI systems more trustworthy \citep{saarela2024recent,xua2024interpretability}, with broad applications in areas such as healthcare, finance, autonomous driving, natural disaster detection, energy management, military and remote sensing \citep{adadi2018peeking, markus2021role, saraswat2022explainable, kadir2023evaluation, hosain2024explainable, hoehl2024opening}. 
Furthermore, explainability is used to help with other dimensions of trustworthiness like privacy, robustness, or fairness \citep{doshi2017towards, yang2019evaluating, arrieta2020explainable, das2020opportunities, markus2021role,
rawal2021recent, agarwal2022openxai}.

% XAI needs evaluation

However, XAI is not a silver bullet.
\Citet{van2021evaluating} point out that humans tend to trust predictions more readily when an explanation is provided, often without carefully examining the explanation itself.
This lack of critical scrutiny can lead to unwarranted trust, especially when decisions are taken based on incorrect or misleading explanations \citep{eiband2019impact, jesus2021can}.
To make matters worse, different XAI algorithms may result in conflicting explanations for the same model and sample \citep{krishna2022disagreement}.
Therefore, simply providing \textit{any} explanation is not sufficient, but it is important to assess the quality of the explanation at hand \citep{sovrano2021survey}. 
Unfortunately, while there is a plethora of XAI methods, evaluation of explanations is still an immature research area \citep{ribera2019can}, with many studies relying on the notion of a good explanation as ``You'll know it when you see it'', providing anecdotal evidence (i.e. small-scale qualitative validation) \citep{doshi2017towards, nauta2023anecdotal, saarela2024recent}. 
Especially in computer vision tasks, evaluation through qualitative inspection of a few examples can be appealing \citep{ibrahim2023explainable}.
However, unstructured qualitative examination yields highly subjective results, as humans struggle at judging the value of XAI explanations \citep{adebayo2018sanity,buccinca2020proxy, hase2020evaluating}. 
In addition, such evaluations risk cherry-picking favorable examples and offer no reliable foundation for comparing different explanation methods across studies or practitioners.
For the same explanation, human ratings vary depending on both the task itself \citep{franklin2022human} and the participant's cultural background \citep{peters2024cultural}. 
Evaluation is further complicated by the lack of ground-truth for the explanations, as it requires knowledge about the model's internal reasoning process  \citep{samek2019explainable, markus2021role, samek2021explaining,  bommer2024finding, ortigossa2024explainable}.

% Evaluation is not done consistently and needs more systematic approaches
Because evaluation is still not performed consistently and seldom systematically \citep{
adadi2018peeking, lipton2018mythos, payrovnaziri2020explainable, messina2022survey, lopes2022xai, de2023implementation, kadir2023evaluation, nauta2023anecdotal, mohamed2024decoding, naveed2024overview, saarela2024recent, salih2024review}, the community frequently calls to develop comprehensive and unified evaluation standards \citep{pinto2024towards, saarela2024recent, xua2024interpretability}. 
A central motivation behind such efforts is to enable the comparison of explanations and asses whether explainability is achieved \citep{markus2021role, zhou2021evaluating}

\begin{figure}
    \centering
    \includegraphics[width=0.7\linewidth]{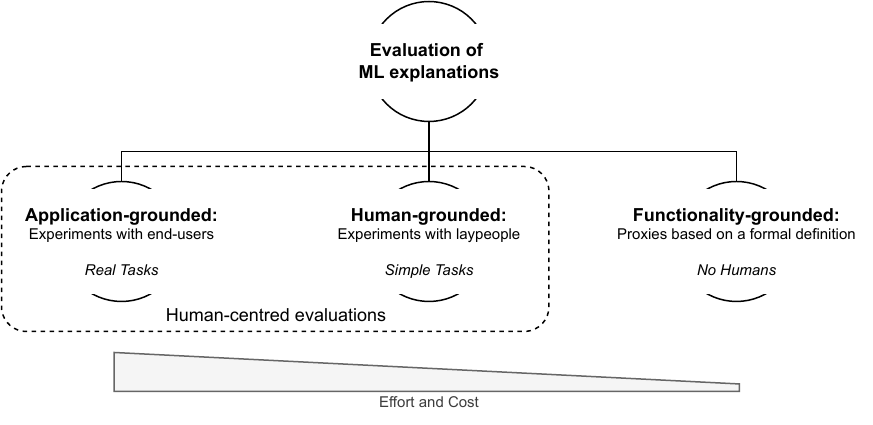}
    \caption{XAI evaluation classified into human-grounded and functionality-grounded evaluation, adapted from the classification framework by \citet{doshi2017towards} and its visualization by \citet{zhou2021evaluating}.}
    \label{fig:evaluation_doshi}
\end{figure}

One of the most prevalent taxonomies reported in the literature \citep{vilone2021notions, zhou2021evaluating, elkhawaga2023evaluating}, and illustrated in \autoref{fig:evaluation_doshi}, is the distinction proposed by \citet{doshi2017towards} between human-grounded and functionality-grounded evaluation methods. The former includes qualitative and quantitative evaluations by laypeople and experts, while the latter consists of (semi-)automatic metrics.

\begin{figure}
    \centering
    \includegraphics[width=0.99\linewidth]{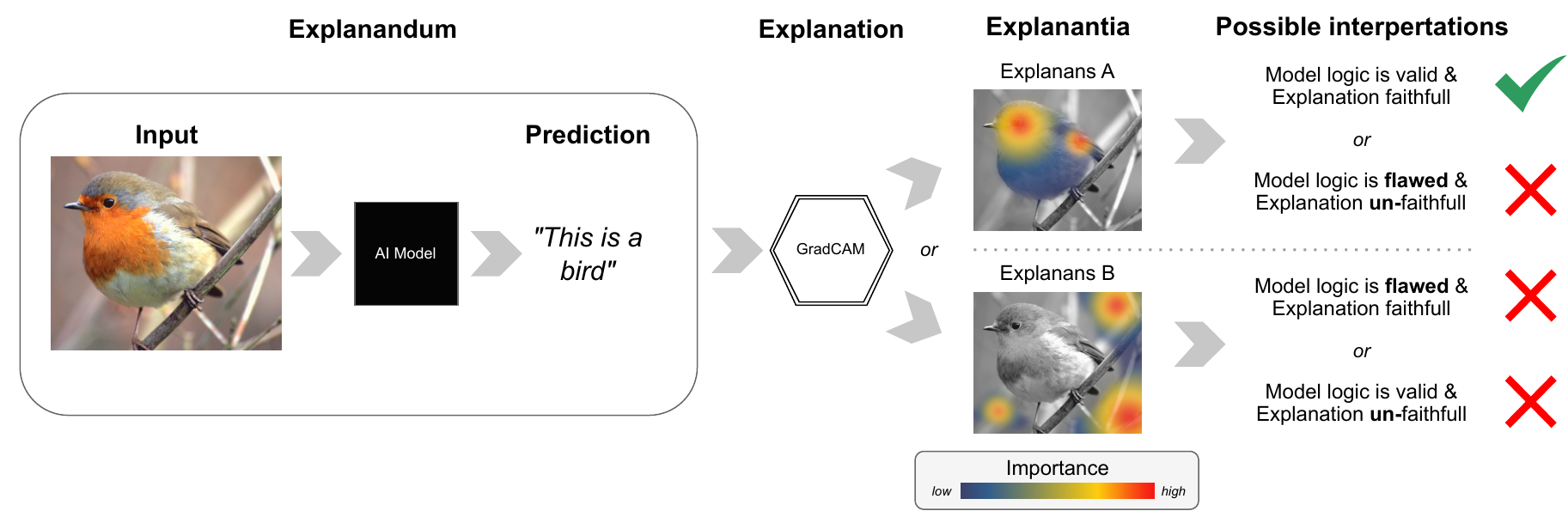}
    \caption{The heatmaps show two alternative example \glspl{explanans}\repeatfootnote{fn:terminology}, indicating which input regions were deemed decisive for the model's decision (the \gls{explanandum}\repeatfootnote{fn:terminology}).
    A qualitative inspection allows for multiple interpretations, as it is unclear whether a) both the model and the explanation process (\gls{explanation}\repeatfootnote{fn:terminology}) are correct or flawed (top), or b) one is correct and the other failed (bottom).
     The green checkmark marks the only scenario in which a human relying on visual plausibility would arrive at a correct conclusion; the red crosses indicate cases where such qualitative judgment would be misleading.
    }
    \label{fig:intro_badexplain}
\end{figure}

Since explanations are meant to aid humans, human-grounded evaluation remains the gold standard to assess their effectiveness in assisting humans \citep{doshi2017towards, gunning2019darpa, miller2017explainable}.
However, the faithfulness (i.e., technical correctness) of an explanation and the plausibility to humans do not necessarily correlate \citep{wiegreffe2019attention, jacovi2020towards, atanasova2024diagnostic}.
Therefore, human-grounded evaluation of  comprehensibility should be distinguished from functionality-grounded evaluation of faithfulness \citep{nauta2023anecdotal}.
Especially humans cannot confidently attribute whether an unexpected \gls{explanans} (i.e., the information provided to explain a decision\footnotemark) is caused by a faulty \gls{explanation} (process\repeatfootnote{fn:terminology}) or a flawed black-box model 
\citep{robnik2018perturbation, zhang2019towards}; see \autoref{fig:intro_badexplain} for an illustration. %
\footnotetext{\label{fn:terminology}The exact definitions of \gls{explanandum}, \gls{explanation}, and \gls{explanans} are provided at the end of \autoref{sec:intro}.}%
In both cases, the consequences can be severe, either reducing trust in a well-functioning model or, more critically, reinforcing trust in a flawed one.
Further, human evaluation, especially through the system's developers, is prone to confirmation bias \citep{doshi2017towards, lipton2018mythos}.

There exist a number of surveys and guidelines that address human-centered evaluations \citep{hoffman2018metrics, miller2019explanation, chromik2020taxonomy, holzinger2020measuring, franklin2022human, hsiao2021roadmap, jesus2021can, langer2021we, mohseni2021multidisciplinary, van2021evaluating, silva2023explainable}.
Unfortunately, the range of reviews dedicated to functionality-grounded evaluation is still limited. 
This is despite the advantage of offering objective, quantitative metrics without requiring human experiments, which can save both time and cost \citep{doshi2017towards, samek2019explainable, zhou2021evaluating}.
Most existing studies are narrow and restricted to a specific application domain \citep{ giuste2022explainable, arreche2024xai}, including cybersecurity \citep{pawlicki2024evaluating}, medical image classification \citep{patricio2023explainable, chaddad2024generalizable}, electronic health record data \citep{payrovnaziri2020explainable}, data and knowledge engineering \citep{li2020survey}, or time-series classification \citep{theissler2022explainable}.
Others focus on particular XAI approaches, such as visual explanations in CNNs \citep{mohamed2022review} or instance-based explanations \citep{bayrak2024evaluation}.
Moreover, many surveys dedicate only limited attention to evaluation metrics, mainly focusing on the XAI methods themselves \citep{carvalho2019machine, ding2022explainability, minh2022explainable, mohamed2022review, ali2023explainable, clement2023xair, patricio2023explainable, chaddad2024generalizable, gongane2024survey, xua2024interpretability}.
By contrast,  this review focuses exclusively on functionality-grounded evaluation across domains and is applicable to a wide range of XAI approaches.
We focus on both in-hoc and post-hoc explanations for black-box models.
Although this work does not consider directly interpretable (white-box) models, many of the presented metrics can also be applied to such models.

\subsubsection*{Contributions}
Despite the growing number of proposed metrics, a comprehensive and unified framework for functionality-grounded evaluation is still missing.
Further, the inconsistent use of terms such as interpretability, comprehensibility, understandability, transparency, and explainability \citep{koh2017understanding, guidotti2018survey, arrieta2020explainable, markus2021role} hampers conceptual clarity and comparability across approaches.
To address this gap, we introduce a framework called \textbf{\gls{vxai}}, aimed at unifying functionality-grounded evaluation for XAI.
An interactive version of the framework is available at \vxailink.

Our contributions are as follows:
\begin{itemize}
\item We perform a systematic literature review based on the \gls{prisma} guidelines by \citet{page2021prisma}, identifying $362$ relevant publications that introduce or utilize evaluation metrics.
\item We aggregate these into $41$ functionally similar metric groups, capturing common methodological patterns across the literature.
\item We propose a three-dimensional categorization scheme consisting of desiderata, explanation type, and evaluation contextuality, and use it to organize the identified metrics.
\item To our knowledge, this results in the most comprehensive and unified VXAI framework to date and provides an extensible foundation for future research.
\end{itemize}

\begin{figure}
    \centering
    {\includegraphics[width=0.8\linewidth, trim = 50 20 30 0, clip]{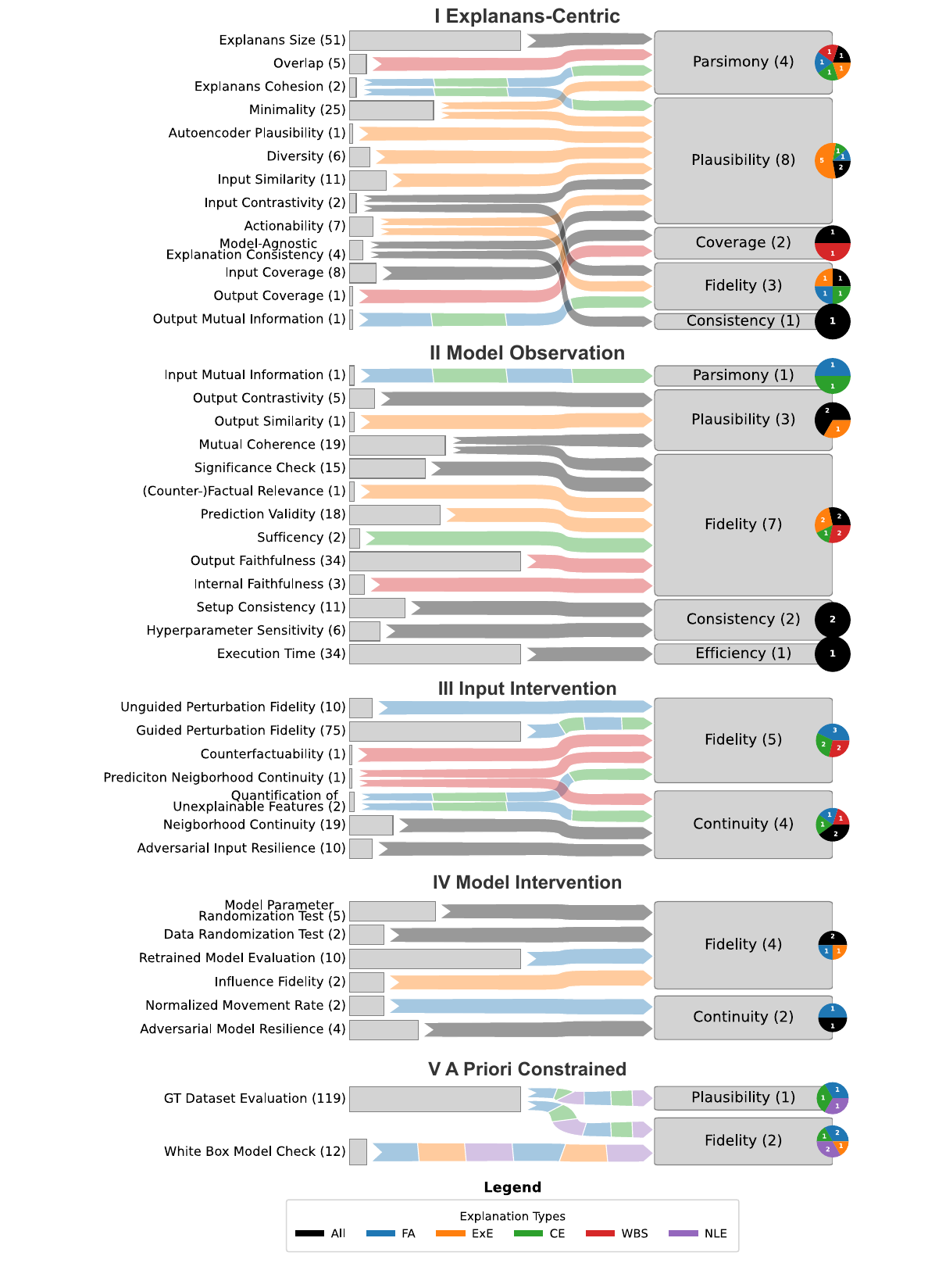}}
    \captionsetup{width=1\linewidth} % limit caption width
    \caption{Overview of the $41$ identified metrics, grouped by contextuality. Each metric is represented by a horizontal bar indicating the number of supporting references. Metrics are mapped to their associated desiderata, with arrow colors denoting the corresponding \glspl{explanation} type. These associations do not distinguish between full (\checkmark) and partial \lightcheckmark ~alignment (unlike \autoref{tab:metric_overview}). For each desideratum, a pie chart summarizes the distribution of linked metrics by explanation type.}
    \label{fig:metric_overview}
\end{figure}

The remainder of this review is structured as follows:
In \autoref{sec:related_work}, we first present related studies on the topic of \gls{vxai} to motivate the need for this systematic review.
Further, \autoref{sec:method} outlines our literature research, with full details provided in \autoref{app:review_method}.
The \gls{vxai} framework is presented in  \autoref{sec:results}, where we introduce our new categorization scheme (\autoref{sec:categorization_scheme}) and summarize the identified metrics (\autoref{sec:identified_metrics}), complemented by a visual overview in \autoref{fig:metric_overview}.
A deeper discussion of these findings is provided in \autoref{sec:discussion}, while comprehensive descriptions of the metrics alongside references are listed in \autoref{app:metrics}. We conclude the review in \autoref{sec:conclusion}, discussing the results and future paths for the area of VXAI.

\subsubsection*{Terminology}
To avoid ambiguous language, throughout the paper we stick to the terminology of the XAI Handbook by \citet{palacio2021xai}: 
The goal of XAI is to facilitate understanding by providing insights into an \textit{\gls{explanandum}} (``What is to be explained''), usually a model or a model's decision.
To accomplish this, we leverage an \textit{\gls{explanation}}, which is the process of getting insight into the \gls{explanandum}. 
The resulting output of this process is the \textit{\gls{explanans}}, which provides the user with information about the model's inner workings.
In a mathematical sense, the \gls{explanation} can be viewed as a function that maps an \gls{explanandum} to an \gls{explanans}.
For example, the \gls{explanandum} could be a CNN's classification of a given input image.
The \gls{explanation} might be an algorithm such as GradCAM \citep{selvaraju2017grad}, and the resulting heatmap is the \gls{explanans}, which highlights important features.
We use the Latin plural forms \glspl{explanandum} (\gls{explanandum}) and \glspl{explanans} (\gls{explanans}) throughout.
When we refer to \gls{vxai}, we include both the evaluation of the method (\gls{explanation}) and its output (\gls{explanans}), since most evaluation metrics necessarily assess the quality of \glspl{explanation} through the quality of their generated outputs.

As defined by the XAI Handbook, \textit{interpretation} (or \textit{interpretability}) refers to the subsequent assignment of meaning to an \gls{explanation}.
It describes the process through which a human infers knowledge about the \gls{explanandum} using the \gls{explanans}.
This step significantly influences the success of the \gls{explanation} and also depends on the receiving human's (the \textit{explainee}'s) mental model.

\hypertarget{def:explan}{}
\begin{terminologybox}{Terminology}
    \centering
    \begin{tabular}{l}
        \textbf{\gls{explanandum}}  (pl. \textbf{\glspl{explanandum}}):   \glsdesc{explanandum}. \\
        \textbf{\gls{explanation}}  (pl. \textbf{\glspl{explanation}}):  \glsdesc{explanation}. \\
        \textbf{\gls{explanans}}  (pl. \textbf{\glspl{explanans}}):  \glsdesc{explanans}. \\
    \end{tabular}

\end{terminologybox}

%%%%%%%%%%%%%%%%%%%%%%%%%%%%%%%%%%%%%%%%%%%%%%%%%%%%%%%%%
%
%
%
%%%%%%%%%%%%%%%%%%%%%%%%%%%%%%%%%%%%%%%%%%%%%%%%%%%%%%%%%

\section{Related Work}
\label{sec:related_work}

\begin{table}
    \centering
    \renewcommand{\arraystretch}{1.75} % Adjust row height for better readability
    \setlength{\tabcolsep}{6pt} % Adjust column spacing

    \newlength{\countwidth}
    \setlength{\countwidth}{3.4cm}
    \newlength{\deswidth}
    \setlength{\deswidth}{6.7cm}

    \rowcolors{2}{gray!10}{white}
    
    \begin{adjustbox}{max width=0.95\linewidth} % Ensure table fits within the page
    \begin{tabular}{c|c|c|c|c|p{\deswidth}|p{3.7cm}|c}
    \toprule
        \textbf{Work} & 
        % \rotatebox{90}{\makecell{\textbf{Primary \gls{vxai} Focus}}} &
        % \rotatebox{90}{\parbox{4.5cm}{\centering \textbf{Primarily Functionality- Grounded VXAI}}} &
        % \rotatebox{90}{\makecell{\textbf{(Semi-)Systematic Review}}} &
        \rotatebox{90}{\makecell{\textbf{\gls{vxai} Focus}}} &
        \rotatebox{90}{\parbox{3cm}{\centering \textbf{Functionality- Grounded VXAI}}} &
        \rotatebox{90}{\makecell{\textbf{(Semi-)Systematic}}} &
        \textbf{~Date~$\downarrow$} &
        \centering\textbf{Desiderata} &
        \centering\textbf{Limited to} &
        \parbox{2cm}{\centering \textbf{Reported Metrics}}\\
        \midrule
        This work & \checkmark & \checkmark & \checkmark & Jan 2025  & \parbox{\deswidth}{ Parsimony, Plausibility, Coverage,\\ Fidelity, Continuity, Consistency, Efficiency}   &   & \parbox{\countwidth}{\centering $41$ metrics \\ from $362$ sources} \\
        \midrule
        \citeauthor{klein2024navigating} & \checkmark & \checkmark & ~ & Jan 2025 & Faithfulness, Robustness, Complexity & Feature Attributions; Computer Vision & 20 metrics \\  
        \citeauthor{pawlicki2024evaluating} & \checkmark & \checkmark & \checkmark & Oct 2024 & ~ &  Cybersecurity & 86 metrics \\  
        \citeauthor{awal2024evaluatexai} & \checkmark & \checkmark & ~ & Jun 2024 & Reliability, Consistency & Rule Explanations & 6 metrics \\  
        \citeauthor{bayrak2024evaluation} & \checkmark & \checkmark & \checkmark & Apr 2024 &  ~ & Counterfactuals & 66 metrics \\  
        \citeauthor{bommer2024finding} & \checkmark & \checkmark & ~ & Mar 2024 & Robustness, Faithfulness, Complexity, Localization, Randomization & Climate Science & 10 metrics \\  
        \citeauthor{li2023m4} & \checkmark & \checkmark & ~ & Dec 2023 & Faithfulness  & Feature Attributions & 6 metrics \\  
        \citeauthor{alangari2023exploring} & \checkmark & ~ & ~ & Aug 2023 & Correctness, Comprehensibility, Stability &  ~ & ~59 metrics \\  
        \citeauthor{le2023benchmarking} & \checkmark & \checkmark & \checkmark & Aug 2023 & Co-12$^\dagger$ &  ~ & 
        \parbox{\countwidth}{\centering 86 metrics\\ from 17 toolkits} \\  
        \citeauthor{salih2024review} & \checkmark & ~ & \checkmark & Aug 2023 &  ~ & Cardiology & 27 metrics \\  
        \citeauthor{kadir2023evaluation} & \checkmark & \checkmark & \checkmark & Jul 2023 &  ~ & ~ & ~80 metrics \\  
        \citeauthor{hedstrom2023quantus} & \checkmark & \checkmark & ~ & Apr 2023 & Faithfulness, Robustness, Localization, Complexity, Axiomatic, Randomization & Feature Attribution & 27 metrics \\  
        \citeauthor{schwalbe2023comprehensive} & ~ & ~ & \checkmark & Jan 2023 &  ~ &  ~ & \parbox{\countwidth}{\centering 11 metrics \\ (already grouped)} \\  
        \citeauthor{agarwal2022openxai} & \checkmark & \checkmark & ~ & Nov 2022 & Faithfulness, Stability & Feature Attributions & 11 metrics \\  
        \citeauthor{coroama2022evaluation} & \checkmark & ~ & ~ & Nov 2022 &  ~ &  ~ & 26 metrics \\  
        \citeauthor{verma2024counterfactual} & ~ & \checkmark & ~ & Nov 2022 &  ~ & Counterfactuals & \parbox{\countwidth}{\centering 9 metrics \\ (already grouped)} \\  
        \citeauthor{belaid2022we} & \checkmark & \checkmark & ~ & Oct 2022 & Fidelity, Fragility, Stability, Simplicity, Stress, Other & Feature Attributions & 22 metrics \\  
        \citeauthor{cugny2022autoxai} & \checkmark & \checkmark & ~ & Oct 2022 &  ~ &  ~ & 6 metrics \\  
        \citeauthor{lopes2022xai} & \checkmark & ~ & \checkmark & Aug 2022 & Fidelity (Completeness, Soundness), Interpretability, Broadness, Simplicity, Clarity) &  ~ & 43 metrics \\  
        \citeauthor{yuan2022explainability} & ~ & \checkmark & \checkmark & Jul 2022 & Fidelity, Sparsity, Stability, Accuracy & Graph Neural Networks & 7 metrics \\  
        \citeauthor{lofstrom2022meta} & \checkmark & ~ & \checkmark & Mar 2022 &  ~ &  ~ & 10 metrics \\  
        \citeauthor{vilone2021notions} & \checkmark & ~ & \checkmark & Dec 2021 &  ~ &  ~ & 36 metrics \\  
        \citeauthor{bodria2023benchmarking} & ~ & \checkmark & \checkmark & Nov 2021 &  ~ &  ~ & 6 metrics \\  
        \citeauthor{sovrano2021survey} & \checkmark & ~ & ~ & Oct 2021 & Similarity, Exactness, Fruitfulness &  ~ & 22 metrics \\  
        \citeauthor{ras2022explainable} & ~ & ~ & ~ & Sep 2021 &  ~ &  ~ & \parbox{\countwidth}{\centering \textit{13 sources in text}\\ (metrics not listed) }  \\  
        \citeauthor{mohseni2021multidisciplinary} & ~ & ~ & \checkmark & Aug 2021 & Fidelity, Trustworthiness &  ~ & 15 metrics \\  
        \citeauthor{yeh2021objective} & \checkmark & \checkmark & ~ & Jun 2021 &  ~ &  ~ & 7 metrics \\  
        \citeauthor{nauta2023anecdotal} & \checkmark & \checkmark & \checkmark & May 2021 & Co-12$^\dagger$ &  ~ & \parbox{\countwidth}{\centering 28 metrics \\ (already grouped)}  \\  
        \citeauthor{zhou2021evaluating} & \checkmark & \checkmark & ~ & Jan 2021 & Fidelity (Completeness, Soundness), Interpretability, Broadness, Simplicity, Clarity) &  ~ & 17 metrics \\  
        \citeauthor{samek2019towards} & ~ & ~ & ~ & Sep 2019 &  ~ &  ~ & \parbox{\countwidth}{\centering \textit{16 sources in text}\\ (metrics not listed) } \\  
        \citeauthor{yang2019evaluating} & \checkmark & \checkmark & \checkmark & Aug 2019 & Generalizability, Fidelity, Persuasibility &  ~ & \parbox{\countwidth}{\centering \textit{40 sources in text}\\ (metrics not listed) } \\  
        \bottomrule
        \rowcolor{white} \multicolumn{8}{l}{$^\dagger${\footnotesize Co-12: Correctness, Output-Completeness, Consistency, Continuity, Contrastivity, Covariate Complexity, Compactness, Composition, Confidence, Context, Coherence, Controllability}}\\
    \end{tabular}

    \end{adjustbox}
    
    \caption{
    Overview of recent XAI reviews, sorted by date. 
    The table indicates whether each survey primarily focused on evaluation metrics, whether it reported mainly functionality-grounded metrics, and whether a (semi-)structured review was conducted.
    The date refers to the earliest available point in the article's timeline;  either the database query, submission, or publication, depending on what was reported.
    For each survey, we also report the desiderata used to classify the metrics and any limitations regarding \gls{explanation} type or application domain.
    Note that not all surveys systematically listed their assessed metrics, so the reported metric count may vary depending on the method of extraction.}
    \label{tab:related_work}
\end{table}

%\subsection{Surveys on VXAI}
Although the field of XAI has gained popularity over the past years, there is still no extensive and unified evaluation framework for XAI metrics.
Various surveys have explored XAI and \gls{vxai} from different angles, ranging from human-grounded evaluation to technical metrics. 
\autoref{tab:related_work} gives an overview over $30$ such XAI reviews from the past years.

While evaluation of XAI is frequently given less attention in XAI surveys, $23$ of these reviews directly focus on the topic of \gls{vxai}. 
Besides functionality-grounded evaluation, a second school of thought is concerned with human-grounded evaluation of \glspl{explanation} through qualitative expert evaluations or quantitative user studies, with representative surveys for this domain available as well \citep{sokol2020explainability, rawal2021recent, naveed2024overview}.
Nevertheless, a considerable number of $19$ reports focus specifically on the topic of functionality-grounded evaluation.
Unfortunately, most of these surveys focus on a subset of well-known metrics, whereas only $14$ surveys gathered \gls{vxai} metrics in a systematic or semi-systematic literature review.
Further, numerous of the referenced reviews either lack an extensive list of desiderata and focus only on a subset of them, or limit their research to specific types of \glspl{explanation}\footnotemark ~or application domains.

There are five reviews, that we consider most similar to this work, as they present systematic functionality-grounded \gls{vxai} surveys. 
\citet{le2023benchmarking} and \citet{nauta2023anecdotal} both categorize the identified metrics based on a scheme of $12$ properties, namely the Co-12 framework, which we discuss in more detail in \autoref{app:xai_desiderata_compare}. 
However, \citet{le2023benchmarking} restrict their analysis to metrics available through public XAI or \gls{vxai} toolkits (e.g., Quantus \citep{hedstrom2023quantus}) and do not report on metrics introduced in the literature but not implemented in such libraries.
Although there is some overlap with the study by \citet{nauta2023anecdotal}, particularly in the inclusion of some identical metrics, their review also incorporates studies that merely apply \gls{vxai} metrics rather than introducing them. In contrast, our work provides detailed descriptions and categorizations of each identified metric.
The review by \citet{kadir2023evaluation} covers a broad range of domains and \gls{explanation} types\repeatfootnote{fn:explanation_type} and reports a wide variety of metrics. It also groups several metrics by method, a strategy shared by our work. However, it does not adopt a categorization scheme based on the desiderata fulfilled by individual metrics.
Notably, the recent reviews from \citet{bayrak2024evaluation} and \citet{pawlicki2024evaluating} report a high number of individual metrics for VXAI.
However, both limit the scope of their review considerably, either in terms of application domain \citep{pawlicki2024evaluating} or explanation type\repeatfootnote{fn:explanation_type} \citep{bayrak2024evaluation}.
In contrast, our work includes all metrics reported to date and introduces a categorization scheme based on three individual dimensions.
Finally, many of the metrics we identified were introduced only recently, underlining the need for this more recent literature review.

\footnotetext{\label{fn:explanation_type}The \textit{explanation type} refers to both the design of the explanation algorithm and, consequently, the nature of the resulting explanans, as introduced in \autoref{sec:explanation_types}.}

While previous reviews report between $10$ and $90$ individual metrics, our work introduces a unified structure by aggregating over $360$ individual metrics into $41$ conceptually related groups. 
This enables clearer comparison and interpretation across metrics.
Unlike most surveys, we do not limit our analysis to specific \gls{explanation} types or application domains, ensuring broader applicability across the XAI landscape.

\section{Method}
\label{sec:method}
\begin{figure}
    \centering
    \includegraphics[width=0.97\linewidth]{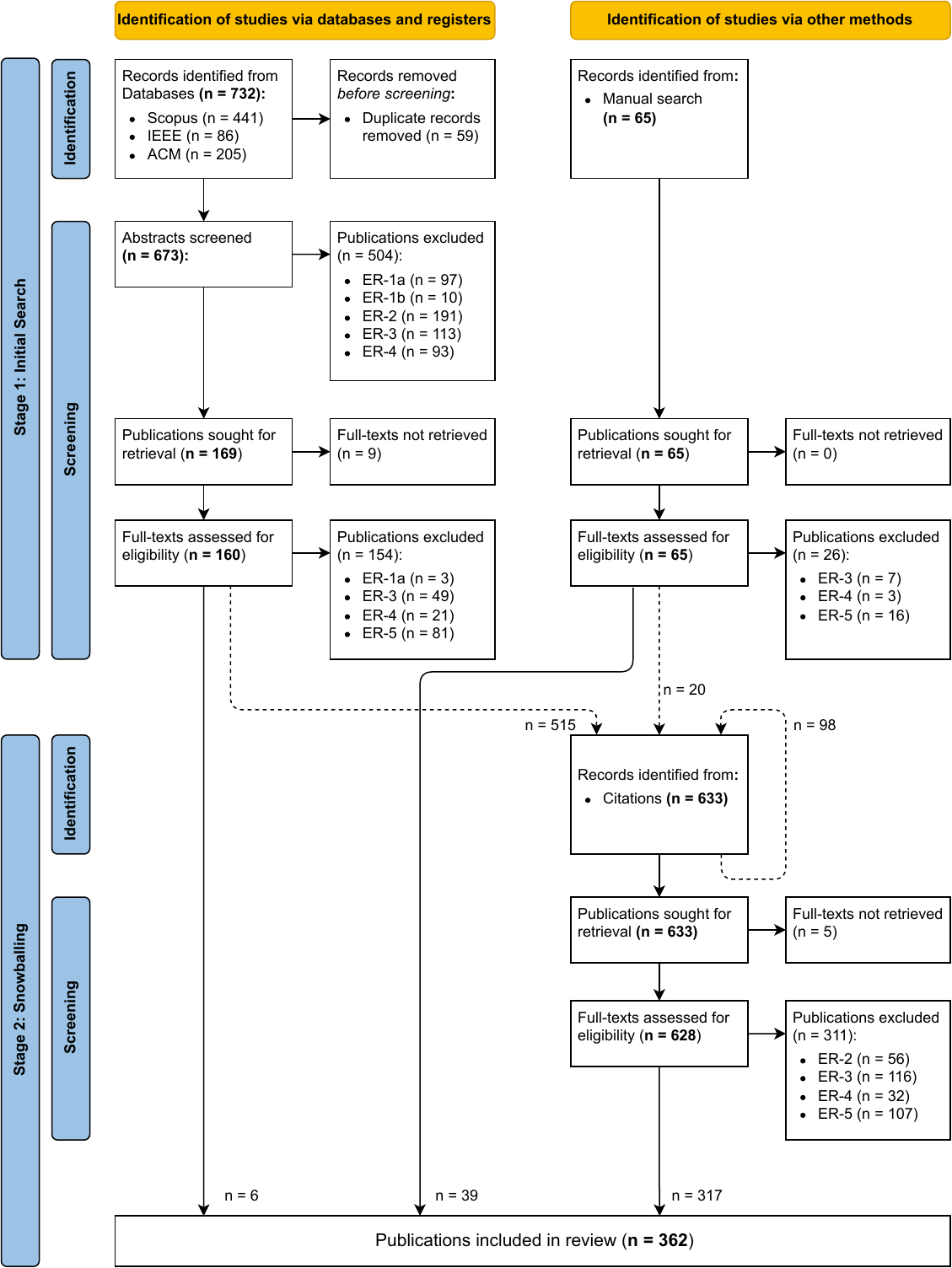}
    \caption{Our search strategy building upon the \gls{prisma} guidelines \citep{page2021prisma}. Notably, we split the process into an initial database search phase and a snowballing phase, identifying the most relevant literature in the second phase.  }
    \label{fig:prisma}
\end{figure}

In our review, we aim to systematically collect and classify all functionality-grounded metrics relevant to evaluating \glspl{explanation} in the context of XAI. 
We base our review on the \gls{prisma} \citep{page2021prisma} guidelines to make the process transparent.
\autoref{fig:prisma} gives an overview of our process.

The overall procedure consisted of two stages:
an initial structured database search to identify secondary literature (e.g., surveys and reviews), followed by a recursive backward snowballing stage targeting primary sources that introduced new metrics. 
In total, we reviewed $1{,}459$ papers, screened $866$ in full, and included $362$ that proposed an original \gls{vxai} metric or one of its variants. 
Comprehensive details of the database queries, screening criteria, and inclusion statistics are reported in \autoref{app:review_method}.

\section{The VXAI Framework}
\label{sec:results}
In this section, we present the twofold contribution that constitutes the \gls{vxai} framework: 
(i) a three-dimensional categorization scheme for evaluating explainability metrics, and 
(ii) a comprehensive overview of the metrics identified through our systematic literature review. 
The categorization scheme provides the conceptual structure used to organize and analyze the collected metrics.

From the literature review, we identified a total of $362$ individual references, some of which introduce or modify multiple metrics simultaneously. 
We grouped these metrics based on methodological similarity, aggregating those that are functionally similar and measure the same underlying property into distinct \textit{aggregated metrics}. 
For each aggregated metric, we determined the associated desiderata, the metric's contextuality, and the suitable \gls{explanation} types as defined in our categorization scheme below.

We first formalize the dimensions of this categorization scheme before presenting the overview of aggregated metrics in \autoref{sec:categorization_scheme}.
To maintain readability, this section this section mainly focusses on a broader overview of the desiderata, and meta-results and statistics of the metrics. 
\autoref{tab:metric_overview} summarizes the \gls{vxai} framework by listing all aggregated metrics structured along the categorization scheme, while detailed descriptions and corresponding references are provided in \autoref{app:metrics}. 
An interactive version of the framework is available at \vxailink.

\subsection{Categorization Scheme}
\label{sec:categorization_scheme}
\subsubsection{Overview}
To facilitate the selection of evaluation metrics for testing explanations, we propose a categorization scheme that groups the identified metrics into functionally similar approaches. 
Serving both as a conceptual overview and a practical guide, it is structured along three orthogonal dimensions:

\begin{enumerate}[label=(\roman*), itemsep = -3pt, topsep = 5pt]
    \item the \textbf{Desiderata} each metric contributes to,
    \item  the suitable \textbf{\Gls{explanation} Types} to which a metric applies, and
    \item  the \textbf{Contextuality}, given by the degree of dependency on the model and data.
\end{enumerate}

\subsubsection{Desiderata}
We annotate each metric with the functional desiderata it addresses. 
To justify our selection, we compared several of the most common desiderata formulations proposed in related works (see \autoref{app:xai_desiderata_compare}). 
We find substantial overlap among these frameworks in their core desiderata, yet also notable gaps: 
some desiderata are inconsistently covered, while others included in prior work do not directly translate to functionality-grounded evaluation. 
Building on this comparison, we derive a coherent set of desiderata that capture both interpretability and technical soundness.

Specifically, we follow a two-stage view of explaining: 
an \textit{Interpretability dimension (I)}, concerned with how the \gls{explanation} is perceived and understood by humans, 
and a \textit{Technical dimension (T)}, focused on the reliability and rigor of the \gls{explanans}. 
The seven desiderata (fully defined in \autoref{app:xai_desiderata_our}) are:

\begin{itemize}[left=10pt, itemsep=0pt, topsep=2pt]
  \item \textbf{Parsimony (I):} The \gls{explanation} should keep the \gls{explanans} concise to support interpretability.
  \item \textbf{Plausibility (I):} The \gls{explanation} should shape the \gls{explanans} to align with human expectations.
  \item \textbf{Coverage (T):} The \gls{explanation} should provide an \gls{explanans} for every \gls{explanandum}.
  \item \textbf{Fidelity (T):} The \gls{explanation} should make the \gls{explanans} reflect the model's true reasoning.
  \item \textbf{Continuity (T):} The \gls{explanation} should ensure that similar \glspl{explanandum} yield similar \glspl{explanans}.
  \item \textbf{Consistency (T):} The \gls{explanation} should produce stable \glspl{explanans} across repeated evaluations.
  \item \textbf{Efficiency (T):} The \gls{explanation} should compute the \gls{explanans} efficiently and broadly.
\end{itemize}

Because various metrics contribute to multiple desiderata, we do not enforce a one-to-one mapping between metrics and desiderata.

\subsubsection{Explanation Types}
\label{sec:explanation_types}
We categorize \gls{vxai} metrics based on the accepted input. 
Apart from a few exceptions, most metrics are agnostic to the underlying black-box model or data format (e.g., tabular, image, or graph). 
Therefore, we do not consider this dimension separately.
Instead, we follow prior XAI classification schemes that organize methods based on the \gls{explanation} approach or resulting \gls{explanans} \citep{carvalho2019machine, markus2021role, zhang2021survey, speith2022review, nauta2023anecdotal}. 

For XAI algorithms, a distinction is often made between local and global methods \citep{zhang2021survey, speith2022review, bedi2024explainable}. 
We do not translate this distinction to \gls{vxai} metrics (unlike \citet{robnik2018perturbation}), as most metrics can be adapted accordingly, for instance, by computing changes in logits for a single instance rather than accuracy over an entire dataset. 
Conversely, metrics that provide local scores can be aggregated across \glspl{explanans} to obtain global results. 
Another frequent distinction is made between \glspl{explanation} that can be built directly into a model (in-hoc) or derived after training (post-hoc) \citep{carvalho2019machine, zhang2021survey, speith2022review, nauta2023anecdotal, bedi2024explainable}. 
The metrics in the \gls{vxai} framework are largely agnostic to this distinction and can generally be applied to either approach.

We differentiate between five principal \gls{explanation} types: 
\textbf{Feature Attributions}, \textbf{Concept Explanations}, \textbf{Example Explanations}, \textbf{White-Box Surrogates}, and \textbf{Natural Language Explanations}.
We introduce each of these types below. 
Similar to the formulation of desiderata, our categorization is extensible and not mutually exclusive. 
Because many XAI methods combine multiple forms of explanations, metrics designed for one type often transfer to others. 
Examples include LIME, which produces a local surrogate yielding attribution scores \citep{ribeiro2016should}, or the generation of counterfactuals by leveraging prior surrogates \citep{pornprasit2021pyexplainer} or attributions \citep{ge2021counterfactual, albini2022counterfactual}. 
This overlap enhances the framework's flexibility and supports systematic comparison across different \gls{explanation} families.

\paragraph{\glspl{FA}}%
\hypertarget{def:FA}{}%
\hspace*{-0.5em}%
return a vector $e \in \mathbb{R}^d$, typically (but not necessarily) matching the dimensionality of the input $x$. 
Each element $x_j$ represents an input feature, such as a column in tabular data, a (super-)pixel in an image, or a node in a graph, with value $e_j$ indicating its relevance to the prediction. 
Attribution values may be positive or negative and can be continuous, discrete, or thresholded, depending on the underlying method. 
Saliency maps form a structured subtype commonly used in computer vision, where features exhibit spatial relations \citep{bach2015pixel, ribeiro2016should, lundberg2017unified, shrikumar2017learning, sundararajan2017axiomatic}. 
Although most \glspl{FA} are local, aggregating individual explanations can yield global feature-importance estimates \citep{lundberg2017unified, molnar2020interpretable}.

\paragraph{\glspl{CE}}%
\hypertarget{def:CE}{}%
\hspace*{-0.5em}%
capture higher-level, human-interpretable properties beyond individual features, such as visual patterns or abstract semantic ideas. 
They are typically extracted from intermediate model representations, resulting in a reduced and interpretable set of dimensions compared to the input space. 
Unlike feature-level relevance, concepts are meaningful on their own and may appear across multiple inputs. 
Consequently, they establish a middle ground between local and global \glspl{explanation}, as their detection in a single instance is local, while their overall contribution to model behavior is global \citep{kim2018interpretability}. 

\paragraph{\glspl{ExE}:}%
\hypertarget{def:ExE}{}%
\hspace*{-0.5em}%
reside directly in the input space and explain models through representative instances. 
They include counterfactuals identifying minimal input changes that alter the prediction outcome \citep{wachter2017counterfactual, karimi2020model, mothilal2020explaining, verma2024counterfactual}, as well as prototypes or factuals representing typical or minimally changed samples within a class \citep{kim2016examples, dhurandhar2019model, koh2017understanding, molnar2020interpretable}. 
Local \glspl{ExE} describe instance-specific changes, while global ones summarize model behavior through sets of influential or representative samples.

\paragraph{\glspl{WBS}}%
\hypertarget{def:WBS}{}%
\hspace*{-0.5em}%
approximate the behavior of a black-box model using interpretable surrogate models that themselves serve as the \gls{explanans}. 
These surrogates can capture global behavior by reconstructing the overall decision logic of the black box, as commonly done with decision trees or rule sets \citep{craven1995extracting, friedman2008predictive}, 
or describe local neighborhoods that provide \glspl{explanans} for specific subsets of inputs \citep{ribeiro2016should, ribeiro2018anchors}. 

\paragraph{\glspl{NLE}}%
\hypertarget{def:NLE}{}%
\hspace*{-0.5em}%
provide textual justifications that accompany or follow model predictions. 
They include models that generate explanations jointly during inference \citep{ras2022explainable, camburu2018snli, wei2022chain} as well as post-hoc generation through large language models \citep{bills2023language}. 
Template-based approaches that merely verbalize existing \glspl{explanans} (e.g., \glspl{FA}) are not regarded as genuine \glspl{NLE}, as they reformulate rather than generate new content \citep{lucieri2022exaid, das2023state2explanation}. 
We therefore propose evaluating the underlying \gls{explanans} and resulting \gls{explanation} instead.

\subsubsection{Contextuality}
Finally, we distinguish metrics by their evaluation context, which defines how strongly they depend on or intervene in the underlying model or data. 
We identify five levels, each introducing progressively deeper contextual interaction:

\begin{enumerate}[label=\textbf{\Roman*)}, itemsep=-3pt, topsep=5pt]
    \item \textbf{\Gls{explanans}-Centric:} Evaluates only the \gls{explanans} in relation to the raw input instance, fully independent of the model.
    \item \textbf{Model Observation:} Relies on access to model outputs or internal activations to assess behavior.
    \item \textbf{Input Intervention:} Perturbs input data and observes resulting changes in predictions or \glspl{explanans}.
    \item \textbf{Model Intervention:} Alters the model itself, e.g., by retraining or parameter randomization.
    \item \textbf{A Priori Constrained:} Requires specific data, architectures, or experimental setups.
\end{enumerate}

We conceptualize Contextuality as a technical property describing how a metric interacts with or intervenes in the evaluated model or explanation. 
This dimension was derived inductively from the surveyed metrics, where recurring technical dependencies naturally formed five distinct contextual levels.

These levels reflect a gradual shift from \textbf{In-Situ} to \textbf{Ex-Situ} evaluation,\footnote{From Latin for ``in place'' and ``off site'', respectively.} 
moving from metrics that operate directly on given \glspl{explanans} and predictions (Levels~I–III) 
to those that evaluate explanation methods under modified or constrained conditions (Levels~IV–V). 
This gradual structure loosely parallels the distinction between ante-hoc, in-hoc, and post-hoc explainability \citep{carvalho2019machine, zhang2021survey, speith2022review, nauta2023anecdotal, bedi2024explainable}, 
as both progress from intrinsic to externally applied processes; though our formulation pertains specifically to evaluation rather than explanation generation. 
While Ex-Situ evaluations support method-level benchmarking, only In-Situ metrics inform the quality of explanations in their specific deployment context. 
By distinguishing these levels, Contextuality assists practitioners in selecting metrics suited to their evaluation setup and interpreting results consistently. 
Since all surveyed metrics align with these five levels, we currently consider the Contextuality dimension comprehensive.

\subsection{Identified Metrics}
\label{sec:identified_metrics}

\begin{table}
    \centering
    \centering
    \renewcommand{\arraystretch}{1.56} % Adjust row height for better readability
    \setlength{\tabcolsep}{6pt} % Adjust column spacing

    \rowcolors{2}{gray!10}{white}
    
    \begin{adjustbox}{max width=0.95\linewidth} % Ensure table fits within the page

\begin{tabular}{
c|l|c
!{\color{LightGray}\vrule width 0.4pt}
c
!{\color{LightGray}\vrule width 0.4pt}
c
!{\color{LightGray}\vrule width 0.4pt}
c
!{\color{LightGray}\vrule width 0.4pt}
c
!{\color{LightGray}\vrule width 0.4pt}
c
!{\color{LightGray}\vrule width 0.4pt}
c
|
c
!{\color{LightGray}\vrule width 0.4pt}
c
!{\color{LightGray}\vrule width 0.4pt}
c
!{\color{LightGray}\vrule width 0.4pt}
c
!{\color{LightGray}\vrule width 0.4pt}
c
|
r
}
\toprule
&& \multicolumn{7}{c|}{\textbf{Desiderata}} & \multicolumn{5}{c|}{\textbf{Explanation Type}} &\\

  \textbf{Contextuality} &
  \textbf{Aggregated Metric} & 
  \multicolumn{1}{c}{\rotatebox{90}{\makecell{\textbf{Parsimony}}}} & 
 \multicolumn{1}{c}{\rotatebox{90}{\makecell{\textbf{Plausibility}}}} 
 & 
 \multicolumn{1}{c}{\rotatebox{90}{\makecell{\textbf{Coverage}}}} 
 & 
 \multicolumn{1}{c}{\rotatebox{90}{\makecell{\textbf{Fidelity}}}} 
 & 
 \multicolumn{1}{c}{\rotatebox{90}{\makecell{\textbf{Continuity}}}} 
 & 
 \multicolumn{1}{c}{\rotatebox{90}{\makecell{\textbf{Consistency}}}} 
 & 
 \rotatebox{90}{\makecell{\textbf{Efficiency}}}
 & 
 \multicolumn{1}{c}{\textbf{\gls{FA}}}
 & \multicolumn{1}{c}{\textbf{\gls{ExE}}}
 & \multicolumn{1}{c}{\textbf{\gls{CE}}}
 & \multicolumn{1}{c}{\textbf{\gls{WBS}}}
 & \textbf{\gls{NLE}} 
 & 
 \multicolumn{1}{c}{\rotatebox{90}{\makecell{\textbf{$\#$References}}}}
 \\
\midrule
        \cellcolor{white} & \hyperref[met:explanans_size]{(1) Explanans Size}  & \checkmark & ~ & ~ & ~ & ~ & ~ & ~ & \checkmark & \checkmark & \lightcheckmark & \checkmark & \lightcheckmark & 51 \\
        \cellcolor{white} &
        \hyperref[met:overlap]{(2) Overlap} & \checkmark & ~ & ~ & ~ & ~ & ~ & ~ & ~ & ~ & ~ & \checkmark & ~ & 5 \\
        \cellcolor{white} & 
        \hyperref[met:explanans_cohesion]{(3) Explanans Cohesion} & \checkmark & \checkmark & ~ & ~ & ~ & ~ & ~ & \checkmark & ~ & \lightcheckmark & ~ & ~ & 2 \\
        \cellcolor{white} &
        \hyperref[met:minimality]{(4) Minimality} & \checkmark & \checkmark & ~ & ~ & ~ & ~ & ~ & ~ & \checkmark & ~ & ~ & ~ & 25 \\
        \cellcolor{white} &
        \hyperref[met:autoencoder_plausibility]{(5) Autoencoder Plausibility} & ~ & \checkmark & ~ & ~ & ~ & ~ & ~ & ~ & \checkmark & ~ & ~ & ~ & 1 \\
        \cellcolor{white} &
        \hyperref[met:diversity]{(6) Diversity} & ~ & \checkmark & ~ & ~ & ~ & ~ & ~ & ~ & \checkmark & ~ & ~ & ~ & 6 \\
        \cellcolor{white} &
        \hyperref[met:input_similarity]{(7) Input Similarity} & ~ & \checkmark & ~ & ~ & ~ & ~ & ~ & ~ & \checkmark & ~ & ~ & ~ & 11 \\
        \cellcolor{white} &
        \hyperref[met:input_contrastivity]{(8) Input Contrastivity} & ~ & \checkmark & ~ & \lightcheckmark & ~ & ~ & ~ & \checkmark & \lightcheckmark & \lightcheckmark & \lightcheckmark & \lightcheckmark & 2 \\
        \cellcolor{white} &
        \hyperref[met:actionability]{(9) Actionability} & ~ & \checkmark & ~ & \checkmark & ~ & ~ & ~ & ~ & \checkmark & ~ & ~ & ~ & 7 \\
        \cellcolor{white} &
        \hyperref[met:ma_explanation_consistency]{(10) Model-Agnostic Explanation Consistency} & ~ & \checkmark & ~ & ~ & ~ & \lightcheckmark & ~ & \checkmark & \checkmark & \lightcheckmark & \lightcheckmark & \lightcheckmark & 4 \\
        \cellcolor{white} &
        \hyperref[met:input_coverage]{(11) Input Coverage} & ~ & ~ & \checkmark & ~ & ~ & ~ & ~ & \checkmark & \checkmark & \lightcheckmark & \checkmark & \lightcheckmark & 8 \\
        \cellcolor{white} &
        \hyperref[met:output_coverage]{(12) Output Coverage} & ~ & ~ & \checkmark & ~ & ~ & ~ & ~ & ~ & ~ & ~ & \checkmark & ~ & 1 \\
        \cellcolor{white} &
        \hyperref[met:output_MI]{(13) Output Mutual Information} & ~ & ~ & ~ & \checkmark & ~ & ~ & ~ & \checkmark & ~ & \checkmark & ~ & ~ & 1 \\
        \midrule
        \cellcolor{white} &
        \hyperref[met:input_MI]{(14) Input Mutual Information} & \checkmark & ~ & ~ & ~ & ~ & ~ & ~ & \checkmark & ~ & \checkmark & ~ & ~ & 1 \\
        \cellcolor{white} &
        \hyperref[met:output_contrastivity]{(15) Output Contrastivity} & ~ & \checkmark & ~ & ~ & ~ & ~ & ~ & \checkmark & \lightcheckmark & \lightcheckmark & \lightcheckmark & \lightcheckmark & 5 \\
        \cellcolor{white} &
        \hyperref[met:output_similarity]{(16) Output Similarity} & ~ & \checkmark & ~ & ~ & ~ & ~ & ~ & ~ & \checkmark & ~ & ~ & ~ & 1 \\
        \cellcolor{white} & 
        \hyperref[met:mut_coherence]{(17) Mutual Coherence} & ~ & \checkmark & ~ & \lightcheckmark & ~ & ~ & ~ & \checkmark & \lightcheckmark & \lightcheckmark & \lightcheckmark & \lightcheckmark & 19 \\
        \cellcolor{white} &
        \hyperref[met:significance]{(18) Significance Check} & ~ & ~ & ~ & \checkmark & ~ & ~ & ~ & \checkmark & \lightcheckmark & \checkmark & \lightcheckmark & \lightcheckmark & 15 \\
        \cellcolor{white} &
        \hyperref[met:CF_relevance]{(19) (Counter-)Factual Relevance} & ~ & ~ & ~ & \checkmark & ~ & ~ & ~ & ~ & \checkmark & ~ & ~ & ~ & 1 \\
        \cellcolor{white} &
        \hyperref[met:pred_validity]{(20) Prediction Validity} & ~ & ~ & ~ & \checkmark & ~ & ~ & ~ & ~ & \checkmark & ~ & ~ & ~ & 18 \\
        \cellcolor{white} & 
        \hyperref[met:suff]{(21) Sufficency} & ~ & ~ & ~ & \checkmark & ~ & ~ & ~ & ~ & ~ & \checkmark & ~ & ~ & 2 \\
        \cellcolor{white} &
        \hyperref[met:out_faith]{(22) Output Faithfulness} & ~ & ~ & ~ & \checkmark & ~ & ~ & ~ & ~ & ~ & ~ & \checkmark & ~ & 34 \\
        \cellcolor{white} & 
        \hyperref[met:int_faith]{(23) Internal Faithfulness} & ~ & ~ & ~ & \checkmark & ~ & ~ & ~ & ~ & ~ & ~ & \checkmark & ~ & 3 \\
        \cellcolor{white} &
        \hyperref[met:setup_consistency]{(24) Setup Consistency} & ~ & ~ & ~ & ~ & ~ & \checkmark & ~ & \checkmark & \checkmark & \lightcheckmark & \checkmark & \lightcheckmark & 11 \\
        \cellcolor{white} &
        \hyperref[met:hyperparam_sens]{(25) Hyperparameter Sensitivity} & ~ & ~ & ~ & ~ & ~ & \checkmark & ~ & \checkmark & \lightcheckmark & \lightcheckmark & \lightcheckmark & \lightcheckmark & 6 \\
        \cellcolor{white} &
        \hyperref[met:extime]{(26) Execution Time} & ~ & ~ & ~ & ~ & ~ & ~ & \checkmark & \checkmark & \checkmark & \lightcheckmark & \checkmark & \lightcheckmark & 34 \\
        \midrule
        \cellcolor{white} & 
        \hyperref[met:unguided_perturb_F]{(27) Unguided Perturbation Fidelity} & ~ & ~ & ~ & \checkmark & ~ & ~ & ~ & \checkmark & ~ & ~ & ~ & ~ & 10 \\
        \cellcolor{white} & 
        \hyperref[met:guided_perturb_F]{(28) Guided Perturbation Fidelity} & ~ & ~ & ~ & \checkmark & ~ & ~ & ~ & \checkmark & ~ & \checkmark & ~ & ~ & 75 \\
        \cellcolor{white} & 
        \hyperref[met:counterfactuability]{(29) Counterfactuability} & ~ & ~ & ~ & \checkmark & ~ & ~ & ~ & ~ & ~ & ~ & \checkmark & ~ & 1 \\
        \cellcolor{white} & 
        \hyperref[met:pred_neigborhood]{(30) Prediction Neighborhood Continuity} & ~ & ~ & ~ & \checkmark & \checkmark & ~ & ~ & ~ & ~ & ~ & \checkmark & ~ & 1 \\
        \cellcolor{white} &
        \hyperref[met:quant_unexplainable]{(31) Quantification of Unexplainable Features} & ~ & ~ & ~ & \lightcheckmark & \checkmark & ~ & ~ & \checkmark & ~ & \lightcheckmark & ~ & ~ & 2 \\
        \cellcolor{white} &
        \hyperref[met:neighborhood_continutity]{(32) Neighborhood Continuity} & ~ & ~ & ~ & ~ & \checkmark & ~ & ~ & \checkmark & \checkmark & \lightcheckmark & \checkmark & \checkmark & 19 \\
        \cellcolor{white} & 
        \hyperref[met:adversarial_input]{(33) Adversarial Input Resilience} & ~ & ~ & ~ & ~ & \checkmark & ~ & ~ & \checkmark & \lightcheckmark & \lightcheckmark & \lightcheckmark & \lightcheckmark & 10 \\
        \midrule
        \cellcolor{white} &
        \hyperref[met:MPRT]{(34) Model Parameter Randomization Test} & ~ & ~ & ~ & \checkmark & ~ & ~ & ~ & \checkmark & \lightcheckmark & \lightcheckmark & \lightcheckmark & \lightcheckmark & 5 \\
        \cellcolor{white} &
        \hyperref[met:DRT]{(35) Data Randomization Test} & ~ & ~ & ~ & \checkmark & ~ & ~ & ~ & \checkmark & \lightcheckmark & \lightcheckmark & \lightcheckmark & \lightcheckmark & 2 \\
        \cellcolor{white} & 
        \hyperref[met:roar]{(36) Retrained Model Evaluation} & ~ & ~ & ~ & \checkmark & ~ & ~ & ~ & \checkmark & ~ & ~ & ~ & ~ & 10 \\
        \cellcolor{white} & 
        \hyperref[met:influence]{(37) Influence Fidelity} & ~ & ~ & ~ & \checkmark & ~ & ~ & ~ & ~ & \checkmark & ~ & ~ & ~ & 2 \\
        \cellcolor{white} & 
        \hyperref[met:NMR]{(38) Normalized Movement Rate} & ~ & ~ & ~ & ~ & \checkmark & ~ & ~ & \checkmark & ~ & ~ & ~ & ~ & 2 \\
        \cellcolor{white} & 
        \hyperref[met:adversarial_model]{(39) Adversarial Model Resilience} & ~ & ~ & ~ & ~ & \checkmark & ~ & ~ & \checkmark & \lightcheckmark & \lightcheckmark & \lightcheckmark & \lightcheckmark & 4 \\
        \midrule
        \cellcolor{white} & 
        \hyperref[met:gt_dataset]{(40) GT Dataset Evaluation} & ~ & \checkmark & ~ & \checkmark & ~ & ~ & ~ & \checkmark & ~ & \checkmark & ~ & \checkmark & 119 \\
        \cellcolor{white} & 
        \hyperref[met:WBM]{(41) White Box Model Check} & ~ & ~ & ~ & \checkmark & ~ & ~ & ~ & \checkmark & \lightcheckmark & ~ & ~ & \lightcheckmark & 12 \\
\bottomrule
\end{tabular}

% Brackets 
\begin{tikzpicture}[remember picture, overlay]

\newcommand{\bracketxpos}{-20.7cm}
\newcommand{\bracketxshift}{-4em}
\newcommand{\bracketamplitude}{25pt}
% Curly bracket A
\draw [decorate, decoration={brace, amplitude=\bracketamplitude, mirror}, thick]
  (\bracketxpos,12.25) -- (\bracketxpos, 3.8) node[midway,xshift=\bracketxshift]{\Large \textbf{I}};
% Curly bracket B
\draw [decorate, decoration={brace, amplitude=\bracketamplitude, mirror}, thick]
  (\bracketxpos,3.55) -- (\bracketxpos, -4.95) node[midway,xshift=\bracketxshift]{\Large \textbf{II}};
% Curly bracket C
\draw [decorate, decoration={brace, amplitude=\bracketamplitude, mirror}, thick]
  (\bracketxpos,-5.2) -- (\bracketxpos, -9.7) node[midway,xshift=\bracketxshift]{\Large \textbf{III}};
% Curly bracket D
\draw [decorate, decoration={brace, amplitude=\bracketamplitude, mirror}, thick]
  (\bracketxpos,-10) -- (\bracketxpos, -13.85) node[midway,xshift=\bracketxshift]{\Large \textbf{IV}};
% Curly bracket E
\draw [decorate, decoration={brace, amplitude=\bracketamplitude, mirror}, thick]
  (\bracketxpos,-14.1) -- (\bracketxpos, -15.35) node[midway,xshift=\bracketxshift]{\Large \textbf{V}};

\end{tikzpicture}
    
    \end{adjustbox}
    
    \caption{The $41$ aggregated metrics identified in our study, as presented in \autoref{app:metrics}.
Each metric is classified according to its associated desiderata, applicable \gls{explanation} types, and level of contextuality. {\small\checkmark} indicates full alignment or reported usage in the literature, while {\small\lightcheckmark} denotes partial contribution or unreported but plausible applicability. The final column shows the number of references per metric. }
    \label{tab:metric_overview}
\end{table}

Now that we have established our classification scheme, we turn to the results of our literature review and report key statistics about the identified metrics.

We identified metrics across $362$ sources.
Since some works proposed multiple variants, the total number of found metrics exceeded $400$.
Many of these metrics followed closely related approaches or represented minor variants of one another, e.g., differing only in hyperparameters, evaluation setup, or dataset-specific details.
To provide a more coherent and interpretable overview, we therefore summarized related metrics into broader, high-level \emph{aggregated metrics} that capture a shared conceptual core.
In total, we derived $41$ functionally distinct aggregated metrics based on shared goals, methods, or assumptions.
Each was then categorized using the three-dimensional scheme introduced above.
Since a single metric may serve multiple desiderata and apply to different \gls{explanation} types, we organize them primarily by their contextuality level. 
\autoref{tab:metric_overview} presents a complete overview of all metrics and their classification.
Key patterns are summarized below; for detailed descriptions of each individual metric, see \autoref{app:metrics}.
Four metrics deemed conceptually flawed or misaligned with any desideratum were excluded and are discussed in \autoref{app:excluded_metrics}.

\paragraph{Metric Popularity and References:}
On average, each metric is supported by $13.4$ references (standard deviation $22.5$), though the median is only $5$. While a few metrics are backed by large reference sets (e.g., $119$, $75$, and $51$ citations), eight metrics are supported by just one.
This should not be interpreted as lack of relevance: our focus was on original proposals, not reuse or popularity.
The most cited metric, Ground-Truth Dataset Evaluation, reflects the ubiquity of using annotated or synthetic datasets for evaluating \gls{explanation} quality; an intuitive strategy with virtually unlimited implementation variants.

\paragraph{Desiderata:}
Most metrics target \textit{Fidelity}, with $18$ doing so directly and $3$ more partially.
\textit{Plausibility} follows with $12$ metrics.
By contrast, \textit{Efficiency}, \textit{Coverage}, and \textit{Consistency} are least represented, with only $1$, $2$, and $2$ metrics respectively.
The majority of metrics ($32$) address a single desideratum, while $4$ target two equally and $5$ contribute partially to a secondary one.

\begin{figure}
    \centering
    \includegraphics[width=1\linewidth]{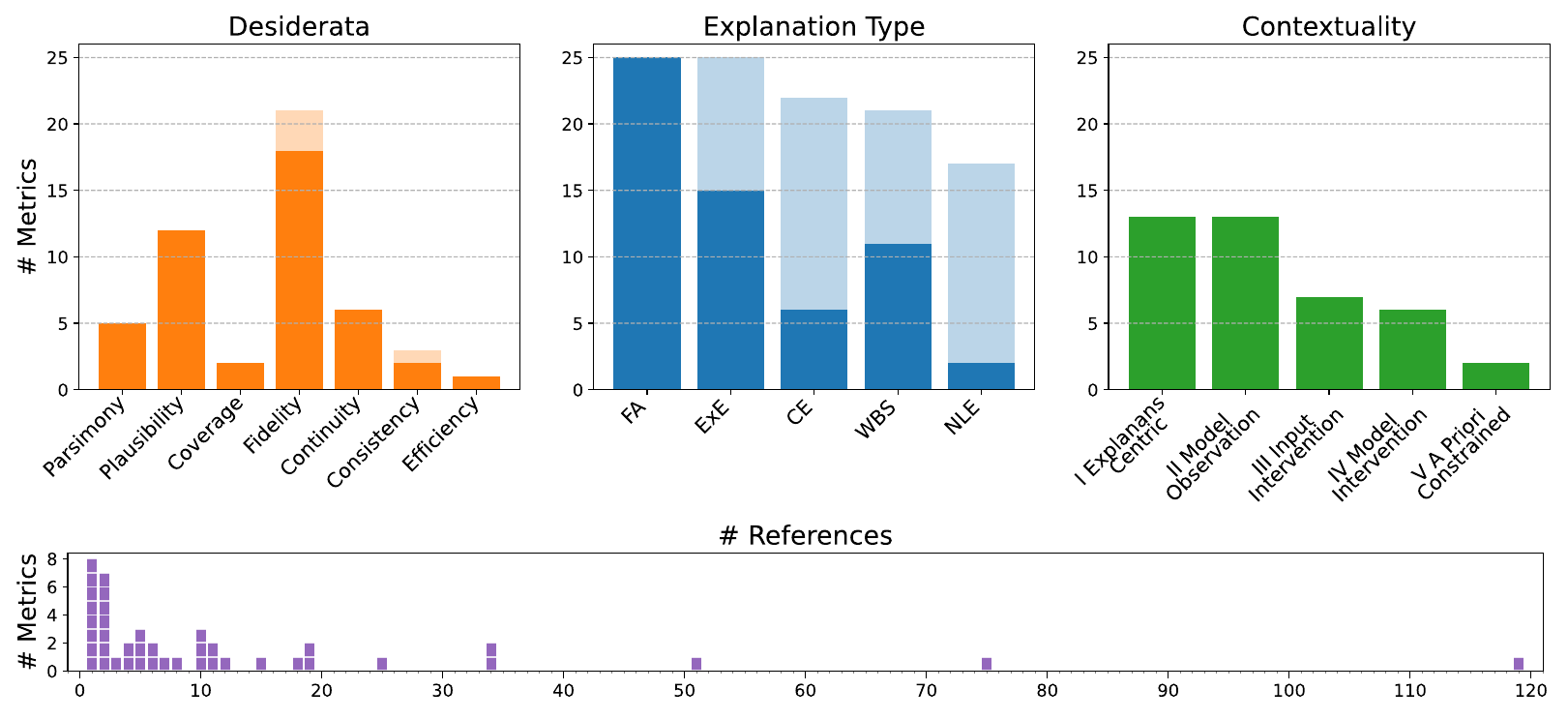}
    \caption{Overview of the distribution of metrics within the categorization scheme. Each metric may be associated with multiple desiderata and \gls{explanation} types, but only a single level of contextuality. Light-colored bars indicate partial alignment with a desideratum. For \gls{explanation} types, light bars denote cases where no usage has been reported in the literature, though the metric is considered adaptable. The bottom histogram shows the number of metrics grouped by their reference count.
 }
    \label{fig:metrics_statistic}
\end{figure}

\paragraph{Explanation Types:}
A total of $25$ metrics are applicable to \gls{FA} methods.
An equal number is available for \gls{ExE}, although only $15$ are directly supported by the literature; the remaining $10$ are marked as potentially applicable (see \lightcheckmark~in \autoref{tab:metric_overview} and the opaque bars in \autoref{fig:metrics_statistic}).
For other \gls{explanation} types, direct literature support is more limited: $11$ metrics for \gls{WBS}, $6$ for \gls{CE}, and just $2$ for \gls{NLE}.
However, we consider many metrics adaptable even in the absence of published usage, increasing the totals to $22$ for \gls{CE}, $21$ for \gls{WBS}, and $17$ for \gls{NLE}.
In total, $15$ metrics are applicable (or adaptable) across all \gls{explanation} types, and $19$ are proprietary to a single type.
Notably, no metric is exclusive to \gls{NLE}, and only one is unique to \gls{CE}.
The largest overlap exists between \gls{FA} and \gls{CE}, with $21$ metrics covering both.
FA also has the highest number of adaptable metrics: while only $3$ are exclusive to \gls{FA} alone, $22$ are shared with other types.

\paragraph{Contextuality:}
Most metrics fall into the less restrictive categories. 
Specifically, $13$ are classified as \textit{Contextuality I} (Explanans-Centric) and another $13$ as \textit{Contextuality II} (Model Observation), both of which do not require altering model or input.
As contextual demands increase, metric availability declines: $7$ fall into \textit{Contextuality III} (Input Intervention), and $6$ into \textit{Contextuality IV} (Model Intervention).
The most restrictive category, \textit{Contextuality V} (A Priori Constrained), includes only two metrics that require specific setups with known ground-truth rationales, either via synthetic data or interpretable white-box models.

\textbf{Desideratum-Contextuality Interactions:}
Metrics targeting  the interpretability desiderata  \textit{Parsimony} and \textit{Plausibility} are found almost exclusively in Contextualities I and II. 
Similarly, the technical desiderata \textit{Coverage}, \textit{Consistency}, and \textit{Efficiency} are addressed only through observational strategies belonging to the same Contextualities of I and II.
In contrast, all metrics measuring \textit{Continuity} require intervention, either through modified inputs (Contextuality III) or altered model internals (Contextuality IV). 
\textit{Fidelity}, in turn, is represented across all contextuality levels.

\textbf{Contextuality-Explanation Interactions:}
There are no striking anomalies in how metrics available for specific \gls{explanation} types are distributed across contextuality levels.
The only exception is that no metric in Contextuality V targets \gls{WBS}, likely because it is difficult to define a single ground-truth surrogate when multiple equivalent surrogates may exist.

\textbf{Desideratum-Explanation Interactions:}
Every \gls{explanation} type is covered by at least one metric for every desideratum, ensuring that all evaluation dimensions are, in principle, addressable regardless of the \gls{explanation} form.
The overall distribution across types is balanced.
The only notable concentration is \textit{Plausibility} in \gls{ExE}, with $10$ metrics.
This may reflect the inherently interpretable nature of many \gls{ExE} techniques (e.g., counterfactuals, prototypes), which naturally suit human-aligned plausibility assessments.

These patterns highlight the breadth of our framework.
Overall, the proposed categorization encompasses a wide range of metrics across \gls{explanation} types, desiderata, and evaluation contexts, offering a comprehensive foundation for structured \gls{vxai} assessment.

\section{Discussion}
\label{sec:discussion}

Finally, we reflect on trends observed in the identified \gls{vxai} metrics and offer considerations for their interpretation and future application.

\paragraph{General observations:}
Our framework reveals a wide and diverse landscape of \gls{vxai} metrics. 
While each aims to quantify a specific property of \glspl{explanation}, few metrics offer clear thresholds or benchmarks to determine what constitutes ``good'' \gls{explanation} quality. 
This lack of consensus limits interpretability and comparability across studies.
Furthermore, although several metrics address similar desiderata, \gls{explanation} types, or contextuality levels, each typically serves a distinct niche. 
These differences often stem from variations in what aspect of a desideratum is targeted, or from adapting to specific \gls{explanation} structures.

\paragraph{Emphasis on Fidelity:}
Fidelity stands out as the most frequently addressed desideratum, both in terms of metric count and methodological variety. 
This supports our perspective of Fidelity as a fundamental aspect of \gls{explanation} quality (see \autoref{sec:desiderata_considerations}).
However, the range of proposed metrics also demonstrates that there is no universally accepted approach to quantifying it.
The same holds for the desiderata Parsimony, Plausibility, and Continuity, each offering a range of possible evaluation metrics.
In contrast, Coverage, Consistency, and Efficiency are addressed by fewer metrics. 
This is likely due to their more straightforward quantifiability; for instance, Coverage is trivially satisfied if \glspl{explanans} exists, and Efficiency can be assessed via computation time. 
Portability, a part of Efficiency, is similarly important, but we identified no functionality-grounded metrics that are associated with it.
Hence, it remains a descriptive property of the method rather than something measurable on individual \glspl{explanans}.
More broadly, the number of metrics associated with a desideratum does not necessarily reflect how well it can be quantified. 
Some desiderata may be better captured by a single strong metric than others by a diverse set of weaker proxies.

\paragraph{Explanation type focus:}
Among \gls{explanation} types, \gls{FA} is the most extensively studied, which is reflected both in literature prevalence and metric availability.
This focus has also enabled the adoption of classical techniques from adjacent fields. 
For example, treating FA \glspl{explanation} as feature selectors allows the use of established stability metrics from feature selection research, such as those used by \citet{nogueira2018stability}. 
Conversely, the \gls{explanation} type \gls{NLE} is still underrepresented in dedicated metrics.
Nonetheless, our framework is flexible enough to incorporate emerging metrics in this and further areas.

\paragraph{Transferability across \gls{explanation} types:}
A notable pattern is that many metrics are rarely applied beyond one or two \gls{explanation} types.
In our analysis, several were marked as ``potentially applicable'' to additional types, but lacked evidence of actual transfer.
This suggests an isolated view of \gls{explanation} types in current research.
We view this as a missed opportunity, as generalizing and adapting metrics across different \gls{explanation} types could foster a more integrated evaluation practice.

\paragraph{Metrics embedded in the \gls{explanation} process:}
Some \gls{explanation} types, notably \gls{ExE} and \gls{WBS}, commonly embed evaluation metrics as part of their generation objectives.
For example, counterfactuals often optimize for plausibility and proximity directly during their construction \citep{wachter2017counterfactual, dandl2020multi, kanamori2020dace}.
While this tight integration may enhance the generated \gls{explanans}, it raises concerns about circularity in evaluation, especially in light of Goodhart's Law \citep{strathern1997improving}: If a metric is optimized for, it may no longer serve as a reliable post-hoc assessment.

\paragraph{Subjectivity and the limits of automation:}
Metrics for Parsimony and Plausibility, while quantifiable, are ultimately shaped by subjective human interpretation.
Functionality-grounded scores offer useful proxies that can be evaluated automatically and without costly human studies.
However, they cannot replace human-centered assessments of whether an \gls{explanation} is truly accessible or helpful in practice.

\paragraph{Caveats in specific metric designs:}
Several metric designs warrant closer scrutiny. Ground-truth-based metrics, specifically those relying on human-annotated rationales (see \metref{met:gt_dataset}), are widely used but have notable limitations.
These include their tendency to reflect plausibility rather than \gls{explanation} fidelity \citep{camburu2019can, carmichael2023well}, vulnerability to trivial baselines like central focus \citep{gu2019understanding}, and their inability to capture model reliance on context or background \citep{arras2022clevr, brandt2023precise}.
There's no reason to assume that the human rationale and the models' rationale are aligned \citep{khakzar2022explanations}, resulting in low scores for truthful \glspl{explanation} generated from an implausible model \citep{samek2019towards}.
Likewise, metrics based on input perturbation or retraining (e.g., \metref{met:guided_perturb_F}, \metref{met:neighborhood_continutity}, or \metref{met:roar}) can yield misleading results. Feature removal may not affect model output if redundant features are present \citep{d2022underspecification}, and retraining can alter the model in uncontrolled ways, obscuring meaningful evaluation \citep{hooker2019benchmark}.

\paragraph{Metric aggregation:}
While most studies report multiple metric scores separately, some recent work proposes aggregating them into a single composite score \citep{farruque2021explainable, margot2021new,  poppi2021revisiting, bommer2024finding}. Suggested methods include scaling by theoretical bounds, comparison to random or perfect baselines, and aggregation via weighted sums or harmonic means. Despite these proposals, the field lacks consensus on how to combine metrics meaningfully and robustly.

\paragraph{The role of white-box models in VXAI:}
These limitations prompt reflection on whether post-hoc \gls{explanation} of black-box models is always justified. Some authors recommend comparing black-box performance to that of directly trained white-box models \citep{zhang2019interpreting, rosenfeld2021better,  margot2021new}. 
If performance is comparable, choosing an interpretable model may be preferable, especially in high-stakes settings, as advocated by \citet{rudin2019stop}. 
The same logic applies when considering surrogate models. They should be benchmarked against white-box models trained directly on the data \citep{barakat2010intelligible}.

Although not the main focus of this work, many of the identified metrics can also be extended to directly interpretable white-box models. 
Here, matching metrics can usually be found in the White-Box Surrogate \gls{explanation} type.
The primary exception are metrics under the Fidelity desideratum, as Fidelity measures the alignment between the \gls{explanans} and the \gls{explanandum}, which presupposes a black-box reference. 
In such cases, the Fidelity desideratum could instead be replaced by a Performance desideratum, where standard predictive metrics such as accuracy or mean squared error serve as proxies for model quality. 
\autoref{app:excluded_desiderata} points to other frameworks in related work, that have such desiderata.

\paragraph{Beyond discriminative tasks:}
While the majority of existing XAI algorithms were developed for discriminative settings, the metrics compiled within the \gls{vxai} framework are formulated independently of the underlying task. 
They evaluate \gls{explanation} properties rather than the model's predictive objective and can, in principle, be applied to generative or other non-discriminative tasks. 
Some adaptations may still be required; for instance, metrics relying on predictive performance (e.g., accuracy or related scores) should employ task-appropriate alternatives such as likelihood-based or reconstruction-based measures.

\subsection{Practical Guidance}

While this work enables researchers to gain a comprehensive overview of the field of \gls{vxai} and the metrics proposed in the literature, we acknowledge that the vast number and diversity of existing metrics pose a major challenge for practitioners when selecting suitable ones. 
As comprehensive selection guidelines remain an open research topic, we offer the following practical considerations derived from our analysis.

First, we encourage researchers and practitioners to address the issue of anecdotal evaluation by systematically applying quantitative metrics rather than relying solely on qualitative or illustrative examples.
Using established metrics supports comparability across studies and strengthens the validity of evaluation results.

To select suitable metrics, practitioners should begin by assessing their test case according to the categorization scheme (\autoref{sec:categorization_scheme}) to identify the appropriate explanation type, as this determines the subset of applicable metrics.
Next, they should identify the key desiderata that are most relevant for their goals and focus their evaluation on these aspects.
We particularly emphasize that, when in doubt, Fidelity is the most critical property from a trustworthiness perspective, while Parsimony is often the most practical consideration for ensuring the accessibility and interpretability of explanations.

Metrics from higher Contextuality stages are typically of greater interest to researchers aiming to evaluate an XAI method in general.
In contrast, such metrics (especially those belonging to Contextuality stage V) are usually less suitable for practitioners operating within a fixed application setup.
Once the relevant subset of metrics has been identified, the final choice should be guided by a careful comparison of their detailed descriptions, as multiple metrics within the same category may still follow different rationales and address distinct aspects.
The metric summaries in \autoref{app:metrics} also include the corresponding references, which can serve as a valuable starting point for further reading and understanding of practical use cases.

To achieve a more comprehensive evaluation, we recommend employing several complementary metrics, ideally with different but reasonable hyperparameter configurations. 
Even in the absence of established benchmarks or threshold values, practitioners can still gain valuable insights by comparing their explanations against na\"ive or random baselines, as it is generally known for each metric whether higher or lower scores indicate better performance.

\section{Conclusion, Limitations \& Future Work}
\label{sec:conclusion}

\subsubsection*{Conclusion}
In this survey, we introduced a unified and comprehensive framework for the evaluation of XAI.
We conducted a systematic review of the literature, identifying original metrics designed to assess \gls{explanation} quality. 
These were grouped into $41$ aggregated metrics and categorized using a three-dimensional scheme based on \textbf{desideratum}, \textbf{explanation type}, and \textbf{contextuality}.

Our analysis reveals a broad range of available metrics, covering diverse use cases and goals.
At the same time, it highlights the lack of consensus regarding when and how specific metrics should be applied.
Moreover, we find that many existing metrics are not yet adapted for all \gls{explanation} types, indicating potential for extension and generalization.

Beyond providing the most extensive synthesis of evaluation metrics to date, the \gls{vxai} framework advances the field conceptually and organizationally. 
It introduces a structured and extensible taxonomy that enables practitioners and researchers to systematically analyze, compare, and select suitable metrics for their evaluation goals.
Unlike prior reviews, \gls{vxai} supports both \emph{horizontal} extension (i.e., incorporating new desiderata, explanation types, or contextuality levels) and \emph{vertical} extension (adding new metrics within the existing structure).
This flexibility allows the framework to evolve in step with emerging explanation paradigms and evaluation needs.

Overall, \gls{vxai} provides a foundation for more transparent, comprehensive, and methodologically grounded evaluation practices in XAI.
It facilitates comparability across studies, encourages reproducible metric application, and paves the way toward standardized evaluation strategies.

\subsubsection*{Limitations and Future Work}
This work focuses exclusively on surveying and categorizing existing metrics.
Although we provide structured comparisons and conceptual insights, we do not conduct empirical benchmarking. 
Future work should investigate how different metrics behave in practice, under which conditions they agree or contradict, and what trade-offs they impose.
This would support the development of a more standardized benchmarking suite for \gls{explanation} evaluation.

Establishing practical thresholds, studying alignment between metrics and desiderata, and exploring aggregation strategies remain open and important questions.
Ideally, this could lead to a broadly accepted set of evaluation standards and protocols for verifying \glspl{explanation} across different models and tasks.

Lastly, while many metrics are labeled  by us as potentially applicable to additional \gls{explanation} types, their use remains unvalidated in literature. 
Future work should adapt and extend these metrics to underexplored \gls{explanation} types, particularly for natural language and concept-based \glspl{explanation}.

\subsection*{Acknowledgment}
This research is funded by the German Federal Ministry for Digital Transformation and Government Modernisation (BMDS) as part of the project \textit{MISSION KI - Nationale Initiative f\"ur Künstliche Intelligenz und Daten\"okonomie} with the funding code 45KI22B021.

We would like to thank our colleagues Kevin Iselborn and Jayanth Siddamsetty for reviewing this survey and sharing helpful comments and insights throughout its development.
We also gratefully acknowledge the anonymous TMLR reviewers, whose valuable feedback helped improve the clarity and quality of this work.

\clearpage
\appendix

\section{Detailed Literature Search Procedure}
\label{app:review_method}
This appendix provides the complete methodological details of our systematic literature review, expanding upon the summary presented in \autoref{sec:method}. 
It includes a full description of the search strategy, databases and queries, screening procedure, inclusion and exclusion criteria, and intermediate statistics. 
The review followed the \gls{prisma} guidelines \citep{page2021prisma} to ensure transparency and reproducibility.

\subsubsection*{Preliminary Consideration}
The research was preceded by two key observations: First, searching for general XAI terms is not feasible, as the database is intractable (searching Google Scholar for ``XAI'' gives over $200{,}000$ results).
Second, \gls{vxai} metrics are usually introduced alongside an XAI method rather than in a dedicated publication, and there exists no unified vocabulary, making it hard to identify relevant sources from keyword searches in titles or abstracts alone.
Hence, we decided to split our research into two stages.
First, we performed a database search using a limited set of search terms to build an initial body of potentially relevant sources.
The first stage was designed primarily to identify \emph{secondary literature} (e.g., reviews and surveys) that could serve as an entry point for the second, snowballing stage. 
In the second stage, we expanded this body through recursive backward snowballing by reviewing references of the already identified publications.
This design explains why comparatively few papers were directly included from Stage~1, while Stage~2, which targeted the \emph{primary literature} containing original metric proposals, yielded a substantially larger number of inclusions.

\subsubsection*{Stage 1: Initial Search}
We started with a database meta-search using 
\textbf{Scopus}\footnote{\href{https://www.scopus.com/search/form.uri?display=advanced}{https://www.scopus.com/search/form.uri?display=advanced}},
\textbf{IEEE}\footnote{\href{https://ieeexplore.ieee.org/search/advanced/command}{https://ieeexplore.ieee.org/search/advanced/command}},
and \textbf{ACM}\footnote{\href{https://dl.acm.org/search/}{https://dl.acm.org/search/}},
to identify existing reviews and surveys that point towards further evaluation metrics.
Using the advanced search features of each database, we designed research queries based on the following key terms:
\begin{lstlisting}[breaklines, escapechar={\%}]
    [Explain*%~%|%~%Interpret*] 
             %$\times$\footnote{The operator $\times$ denotes the concatenation of terms into composite search phrases (e.g., “Explain* AI”, “Interpret* ML”).}%  
    [AI%~%|%~%Artificial Intelligence%~%|%~%ML%~%|%~%Machine Learning]
            AND
    [Evaluation%~%|%~%Metric%~%|%~%Quantification]
            AND
    [Survey%~%|%~%Review]
\end{lstlisting}%
The exact search strings for each database are provided in \autoref{sec:queries}.
The last search was conducted on January 15, 2025.

This resulted in a total of $673$ identified articles after de-duplication. 
We first screened titles and abstracts, excluding sources that were not research articles, did not involve XAI or VXAI, or focused on human-based evaluation.
Through this first screening, we excluded $504$ articles, using the following exclusion criteria:

\begin{enumerate}[topsep=3pt,itemsep=-1ex,partopsep=1ex,parsep=1ex, label=\textbullet\hspace*{3pt}ER-\arabic*:]
    \setlength{\itemindent}{.5in}
    \item General Issues
    \begin{enumerate}[topsep=-10pt,itemsep=-1ex,partopsep=1ex,parsep=1ex, label=\textbullet\hspace*{3pt}ER-1\alph*:]
    \setlength{\itemindent}{.5in}
        \item Not a research article
        \item Not English
    \end{enumerate}
    \item No XAI (or VXAI) at all, or with explainability outside our scope (e.g., ante-hoc data analysis)
    \item Contains XAI, but no systematic evaluation of XAI (e.g., ``anecdotal evidence'')
    \item Contains systematic VXAI, but only human-centered methods
\end{enumerate}

Out of the $169$ articles that passed the initial title and abstract screening, $9$ could not be retrieved. 
For the remaining $160$ articles, we conducted full-text screening, applying the exclusion criteria described above. 
To avoid an intractable number of limited-value sources, we further excluded works which used functionality-grounded \gls{vxai} metrics already introduced by other articles, introducing the following exclusion criterion:
\begin{enumerate}[topsep=3pt,itemsep=-1ex,partopsep=1ex,parsep=1ex]
    \setlength{\itemindent}{.5in}
    \item[\textbullet\hspace*{3pt}ER-5:] Contains functionality-grounded evaluation, but does not introduce a new metric itself.
\end{enumerate}
Instead, we added the cited metric to our corpus for the snowballing phase, yielding $515$ potentially relevant sources. 
Of the articles retrieved through the initial database search, only $6$ were included in our review.

To complement the structured database search, we also conducted an unstructured manual search using Google Scholar and other sources such as personal literature databases, citation alerts, and colleague recommendations. 
This additional search yielded $65$ articles, of which $39$ were included directly in the review. 
A further $20$ references from these articles were added to the corpus for the snowballing phase. 
Duplicates already identified during the database search were not counted again, and duplicates among citations from the manual search were likewise removed. 
We did not specifically search preprint repositories such as arXiv, but relevant preprints that appeared through other channels were included if they met our inclusion criteria.

In total, during the first stage, we included $45$ articles in our review ($6$ from the database search and $39$ from additional sources).
We also compiled $535$ additional references for the second-stage backward snowballing ($515$ from the database search and $20$ from the additional sources).

\subsubsection*{Stage 2: Snowballing}
Starting with the corpus of $535$ references identified in the first stage, we conducted full-text screening on all records as described above.
During the recursive backward snowballing process, we identified and assessed an additional $98$ references for eligibility.
We observed that the set of relevant records quickly began to converge: most newly encountered papers that did not present original metrics (ER-5) cited articles already included in our corpus.
We did not perform forward snowballing (i.e., identifying articles that cite our included works), as our focus was on original metric proposals. 
Including such follow-up papers would have significantly increased the number of records beyond a manageable scope.

In total, we assessed $628$ articles during the second phase ($535$ from the first stage and $98$ from recursive snowballing).
Out of these, we included $317$. 
Compared to the $20\%$ inclusion rate in the first stage, the nearly $50\%$ inclusion rate in the second phase confirms the effectiveness of our strategy.
The initial database search primarily yielded secondary sources (e.g., XAI and \gls{vxai} reviews), which helped identify original metric proposals during snowballing.

Overall, we reviewed $1,459$ articles, screened $866$ in full, and included $362$ that originally proposed a \gls{vxai} metric or one of its variants.

\clearpage
\subsection{Queries}
\label{sec:queries}

\subsubsection*{Scopus}
\begin{lstlisting}[breaklines]
TITLE-ABS-KEY (
  ("Explain* AI" OR "Explain* Artificial Intelligence" OR "Explain* ML" OR "Explain* Machine Learning" OR
  "Interpret* AI" OR "Interpret* Artificial Intelligence" OR "Interpret* ML" OR "Interpret* Machine Learning")
  AND
  ("Evaluation" OR "Metric" OR "Quantification")
  AND 
  ("Survey" OR "Review")
)
\end{lstlisting}

\subsubsection*{IEEE Access}
\begin{lstlisting}[breaklines]
("All Metadata":"Explain* AI" OR "All Metadata":"Explain* Artificial Intelligence" OR "All Metadata":"Explain* ML" OR "All Metadata":"Explain* Machine Learning" OR
"All Metadata":"Interpret* AI" OR "All Metadata":"Interpret* Artificial Intelligence" OR "All Metadata":"Interpret* ML" OR "All Metadata":"Interpret* Machine Learning")
AND
("All Metadata":"Evaluation" OR "All Metadata":"Metric" OR "All Metadata":"Quantification")
AND
("All Metadata":"Survey" OR "All Metadata":"Review")
\end{lstlisting}

\subsubsection*{ACM Digital Library}
\begin{lstlisting}[breaklines]
ANYWHERE:[
    ("Explain* AI" OR "Explain* Artificial Intelligence" OR "Explain* ML" OR "Explain* Machine Learning" OR
    "Interpret* AI" OR "Interpret* Artificial Intelligence" OR "Interpret* ML" OR "Interpret* Machine Learning")
    AND
    ("Evaluation" OR "Metric" OR "Quantification")
    AND 
    ("Survey" OR "Review")
    ]
\end{lstlisting}

\clearpage
\section{Desiderata of XAI}
\label{app:xai_desiderata}
\textit{This appendix expands on the concise outline of desiderata used in the categorization scheme (see \autoref{sec:categorization_scheme}) and provides the full background and definitions.}
A well-founded evaluation of XAI methods requires clearly defined criteria for what constitutes a good \gls{explanation}.
To establish such criteria, we must first reflect on the role of \glspl{explanation} in the context of XAI.
According to the definition in the XAI Handbook \citep{palacio2021xai}, explaining a model and its behavior is a two-stage process:
first, factual information about the model's decision process is generated (the \gls{explanans}); this is then interpreted by the human user.
The first stage can be evaluated using technical criteria that assess whether the model's reasoning has been captured truthfully and reliably.
The second stage depends on the interpretability of the \gls{explanation}, which can be assessed using general cognitive principles, even in the absence of a specific user model.

To capture the multifaceted nature of explanation quality, a number of desiderata have been proposed in the literature.
We interpret these as functionality-grounded expectations that reflect the demands of both stages of the explanation process.
We first provide an overview of existing desiderata proposed in prior work, then introduce a unified set that systematically describes the requirements for ensuring technical soundness and for bridging the interpretation gap. 

\subsection{Common Formulation of Desiderata}
\label{app:xai_desiderata_compare}
Several XAI surveys report that there is no ubiquitous consensus on appropriate desiderata, with some of the categories related to goals pursued \textit{through} XAI, rather than standalone desiderata of XAI, e.g., Trustworthiness, Acceptance, or Fairness \citep{doshi2017towards, langer2021we, vilone2021notions, elkhawaga2023evaluating}.
Hence, we conduct a scoping review, reporting the main desiderata used by different authors and analyzing the similarities as well as differences in their formulations.
For the sake of brevity, we exclude some of the papers listed in \autoref{tab:related_work}, as the missing ones either overlap considerably (e.g., \citet{awal2024evaluatexai} and \citet{klein2024navigating}), rely on a different notion of desiderata (e.g., \citet{sovrano2021survey}), or use no desiderata at all. 
We first present these frameworks using the authors' original terminology before introducing our own categorization scheme.

The famous \textbf{Co-12} properties, introduced by \citet{nauta2023anecdotal} and reused by \citet{le2023benchmarking}, form one of the most extensive existing frameworks for categorizing XAI metrics.
They group properties along three dimensions: Content (\textit{Correctness, Completeness, Consistency, Continuity, Contrastivity, Covariate Complexity}), Presentation (\textit{Compactness, Composition, Confidence}), and User (\textit{Context, Coherence, Controllability}).
While the first dimension focuses on information contained in the \gls{explanans}, the second and third dimensions concern how this information is conveyed.
Although some of these human-centered properties can be measured through proxies, others may mainly be assessed through human-grounded evaluation.

\citet{zhou2021evaluating}, based on the taxonomy of \citet{markus2021role}, define \textit{Interpretability} and \textit{Fidelity} as the two major components of explainability.
The former is concerned with providing understandable \glspl{explanans} and includes the properties of \textit{Clarity}, \textit{Broadness}, and \textit{Parsimony}.
Fidelity, refers to how accurately an \gls{explanans} reflects the model's behavior, and consists of the properties of \textit{Completeness} and \textit{Soundness}.

The framework proposed by \citet{robnik2018perturbation}, which was adopted by \citet{carvalho2019machine} and \citet{molnar2020interpretable}, differentiates between properties of \glspl{explanation} and individual \glspl{explanans}.
Investigating the properties of methods (i.e., \glspl{explanation}), they consider \textit{Translucency}, \textit{Portability}, and \textit{Algorithmic Complexity}, which can all be interpreted as desiderata, while \textit{Expressive Power} is a descriptive property.
The properties of individual \glspl{explanans} include \textit{Comprehensibility}, \textit{Importance}, \textit{Representativeness}, \textit{Fidelity}, and \textit{Stability}.
However, their categorization encompasses further properties, which we do not consider as proper desiderata of XAI: \textit{Accuracy}, \textit{Novelty}, \textit{Certainty}, and \textit{Consistency}.
Accuracy is a measurement of the underlying black-box model, while Novelty and Certainty are rather \glspl{explanans} themselves than properties of general \glspl{explanation}. 
Further, Consistency between different black-box models is not necessarily a useful measure, as different models may derive similar predictions based on different reasoning (see Rashomon Effect \citep{breiman2001statistical, leventi2022rashomon}).

The famous XAI review by \citet{guidotti2018survey}, inspired by earlier works such as \citet{andrews1995survey} and \citet{johansson2004truth}, reports three less fine-grained desiderata: \textit{Interpretability}, \textit{Fidelity}, and \textit{Accuracy}.
Similar to previously discussed reviews, Interpretability describes human understandability, while Fidelity measures how well the \gls{explanans} imitates the black box, and Accuracy focuses on predictive performance, which is outside the scope of XAI in our context.
Additionally, \textit{Consistency} is introduced by \citet{andrews1995survey}, expecting reproducible \glspl{explanation}, while \citet{johansson2004truth} emphasize the \gls{explanation}'s algorithmic \textit{Scalability} and \textit{Generality}.

\citet{alvarez2018towards}, \citet{jesus2021can}, and \citet{alangari2023exploring} all report a similar set of desiderata.
The understandability of \glspl{explanation} is measured in terms such as \textit{Interpretability}, while \textit{Faithfulness} and the corresponding desiderata give insight into how truthful the \gls{explanation} is to the underlying black-box model.
All three works further report the \textit{Stability} of \glspl{explanation} as a desired property, assessing whether \glspl{explanans} on similar inputs are similar.

In their Quantus toolkit, \citet{hedstrom2023quantus} (and the follow-up study by \citet{bommer2024finding} as well), categorize their metrics partly through desiderata, namely \textit{Faithfulness}, \textit{Robustness}, and \textit{Complexity}.
Simultaneously, part of their metrics are grouped by their conceptual similarity, including \textit{Localization}, \textit{Randomization}, and \textit{Axiomatic}.

Finally, the Compare-xAI benchmark by \citet{belaid2022we} organizes functional tests into six categories, namely \textit{Fidelity}, the robustness-related \textit{Stability} and \textit{Fragility}, the interpretability desideratum \textit{Simplicity}, and the explanation-methods-focused \textit{Stress} and \textit{Portability} (which they integrate under ``Other'').

While many existing frameworks overlap conceptually, a unified and practically usable categorization scheme for \gls{vxai} metrics is still lacking.
This requires a structured set of desiderata that defines what makes a good explanation and supports consistent classification of metrics.
Prior work often enforces a rigid one-to-one mapping between metrics and desiderata; in contrast, we decouple these dimensions, defining a set of mostly independent desiderata to which each metric may contribute individually or jointly.
Lightweight frameworks tend to omit critical aspects of explanation quality, while broader ones sometimes include goals that are not intrinsic to the explanation itself (e.g., accuracy).
We restrict our scope to properties that reflect the explanation rather than the underlying model and clarify excluded cases after presenting our set.
Although all desiderata rely on proxies, we limit ourselves to properties that are quantifiable in principle.
Highly abstract or vague notions lacking empirical grounding are omitted. 
Lastly, our framework is designed to be extensible, allowing the integration of future desiderata as the field evolves.

\subsection{Proposed Set of Desiderata}
\label{app:xai_desiderata_our}
Building on the desiderata established above and our findings on \gls{vxai} metrics, we propose a set of seven desiderata to serve as a categorization scheme for \gls{vxai}.

Our goal is to offer a principled yet practical structure that enables consistent classification while avoiding the limitations of prior definitions. 
These are either too narrow to accommodate relevant metrics or too broad and include properties beyond explainability. 
While properties such as fairness are often measured using XAI methods, we consider them beyond the scope of \gls{vxai}, because they assess the model's behavior rather than the \gls{explanation} itself.

Building on the two-stage view of explaining described by \citet{palacio2021xai} (i.e., presenting factual information followed by human interpretation), we define two complementary dimensions of explanation quality.
The \textbf{Technical} (T) dimension comprises desiderata that assess the factual correctness, robustness, and reliability of the \gls{explanation}, ensuring that it faithfully reflects the model's reasoning. 
In contrast, the \textbf{Interpretability} (I) dimension captures how the \gls{explanation} is conveyed and how accessible, intuitive, and useful it is to a general-purpose user.
This separation is aligned with existing frameworks such as the Co-12 properties \citep{nauta2023anecdotal} and the taxonomy by \citet{zhou2021evaluating}.
The desiderata are designed to be as independent from each other as possible, allowing for reliable quantification of different aspects relevant to trustworthy XAI.

\begin{figure}
    \centering
    \includegraphics[height=0.95\textheight]{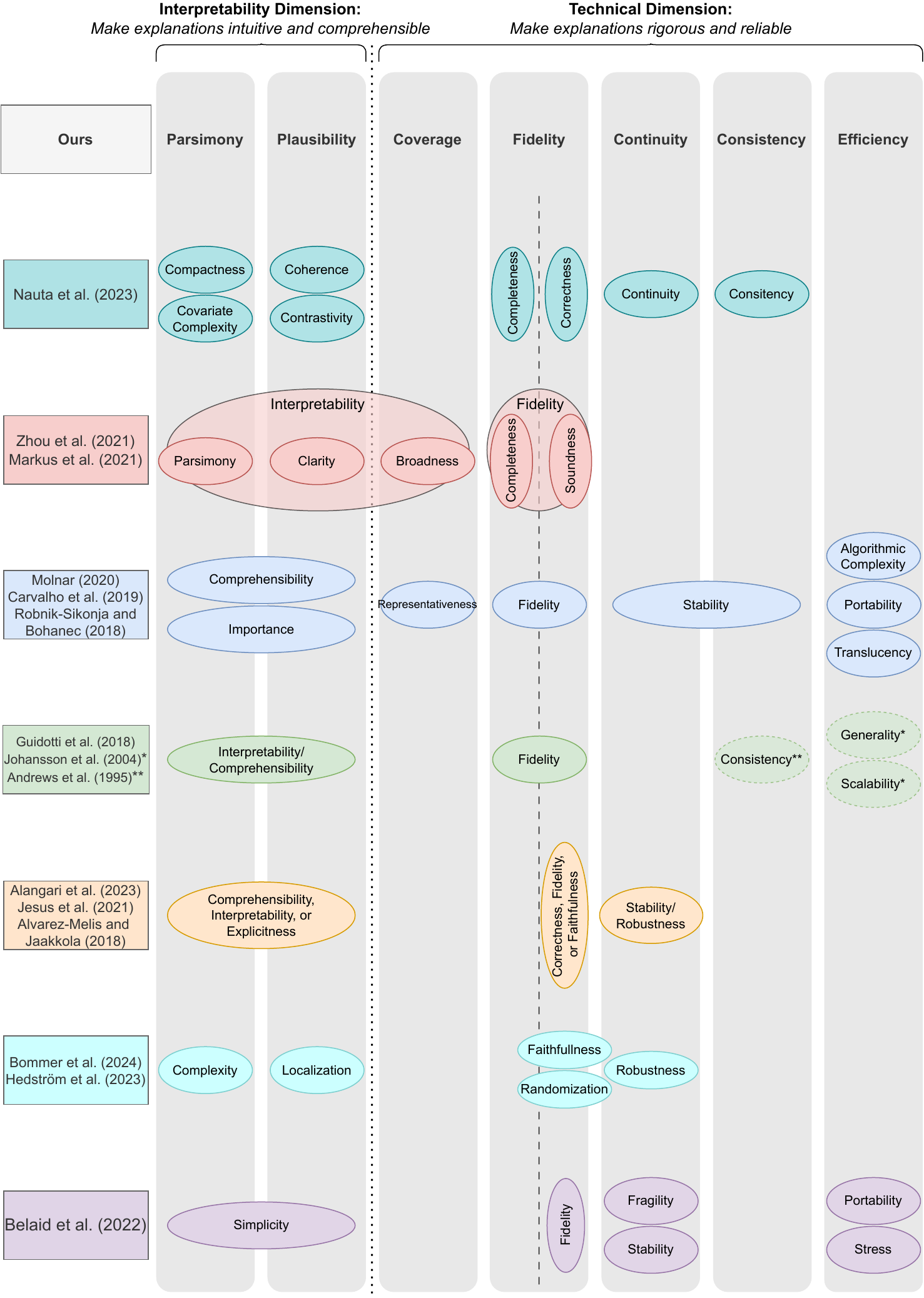}
    \caption{Comparison of our desiderata with those used in related work. Desiderata outside our framework are omitted. Desiderata reported by only individual sources (per row) are denoted by an asterisk~(*).}
    \label{fig:xai_desiderata}
\end{figure}

We present our set of desiderata and its relation to other frameworks in \autoref{fig:xai_desiderata} and introduce them in more detail in the following paragraphs. 
In total, we define seven desiderata, two associated with Interpretability and five belonging to the Technical dimension:

\begin{itemize}[left=10pt, labelsep=5pt]
    \item[(I)] \textbf{Parsimony}: The \gls{explanation} should keep the \gls{explanans} concise to support interpretability.
    \item[(I)] \textbf{Plausibility}: The \gls{explanation} should shape the \gls{explanans} to align with human expectations.
    \item[(T)] \textbf{Coverage}: The \gls{explanation} should provide an \gls{explanans} for every \gls{explanandum}.
    \item[(T)] \textbf{Fidelity}: The \gls{explanation} should make the \gls{explanans} reflect the model's true reasoning.
    \item[(T)] \textbf{Continuity}: The \gls{explanation} should ensure that similar \glspl{explanandum} yield similar \glspl{explanans}.
    \item[(T)] \textbf{Consistency}: The \gls{explanation} should produce stable \glspl{explanans} across repeated evaluations.
    \item[(T)] \textbf{Efficiency}: The \gls{explanation} should compute the \gls{explanans} efficiently and broadly.
\end{itemize}

\subsubsection{Parsimony}
\begin{definitionbox}{Definition: Parsimony}
The \gls{explanation} should keep the \gls{explanans} concise to support interpretability.
\end{definitionbox}%
\noindent
The primary purpose of an \gls{explanation} is to convey information about the black-box model or its decision process to humans.
Therefore, the resulting \gls{explanans} must be expressed in a way that the human mind can grasp easily to increase the \glspl{explanation} success. 
While actual interpretability can only be evaluated through human-grounded evaluation, Parsimony is one of the most prevalent proxies defined to serve as a functionality-grounded approximation.
Since our mental capacity is limited and we tend to struggle with an overload of information \citep{miller1956magical, miller2019explanation, alangari2023exploring}, providing short and simple \glspl{explanans} helps humans understand more effectively.

\citet{nauta2023anecdotal} introduce the property of Compactness, arguing that a briefer \gls{explanans} is easier to understand.
Similarly, they use Covariate Complexity to assess how complex the features are that constitute the \gls{explanans}, where higher interpretability is supported by providing a few high-level concepts in favor of a very granular \gls{explanans}.
Both of these aspects are summarized under Parsimony by \citet{markus2021role} and \citet{zhou2021evaluating}, preferring simpler \glspl{explanans} over longer or more complex ones. 
The scheme used in the Quantus library \citep{hedstrom2023quantus, bommer2024finding} defines a group called Complexity.
It specifically tests for concise \glspl{explanans} and aims to have as few features as possible to be easier to understand.
The associated interpretability desiderata from other authors are defined less explicitly, but similarly favoring simpler \glspl{explanans} \citep{andrews1995survey, johansson2004truth, guidotti2018survey}, proposing that simple \glspl{explanans} should be short \citep{alvarez2018towards, jesus2021can, alangari2023exploring}, promoting small \glspl{explanans} and only focusing on relevant parts \citep{robnik2018perturbation, carvalho2019machine, molnar2020interpretable}, and expecting an \gls{explanans} with concentrated information to facilitate human understanding \citep{belaid2022we}. 

Following the proposed definitions, we include Parsimony as one of our desiderata.
It expects \glspl{explanans} to be as brief and concise as possible, to ensure that rationales can be understood easily and fast. 
We focus Parsimony exclusively on this aspect, as other associated properties are either covered by separate desiderata (such as truthfulness of the \gls{explanation}) or excluded entirely as they are not functionality-grounded (general understandability of the \gls{explanation}).

\subsubsection{Plausibility}
\begin{definitionbox}{Definition: Plausibility}
The \gls{explanation} should shape the \gls{explanans} to align with human expectations.
\end{definitionbox}%
\noindent
To improve the acceptance of \glspl{explanation} and facilitate their interpretation, the Plausibility desideratum is contained in most authors' interpretability desiderata.
Coherence as defined by \citet{nauta2023anecdotal} is the accordance of an \gls{explanans} with the user's previous knowledge and expectations.
The metrics classified under Localization by \citet{hedstrom2023quantus} evaluate whether an \gls{explanation} shows a rationale similar to what humans would expect.
This is concordant with the definition of Comprehensibility \citep{alvarez2018towards, jesus2021can, alangari2023exploring}, which states that an \gls{explanans} should be similar to what a human expert would choose as the correct rationale.
Furthermore, \citet{nauta2023anecdotal} introduce Contrastivity, which supports Plausibility, as an \gls{explanans} should be specific to the given \gls{explanandum}.
Similarly, Clarity is introduced by \citet{markus2021role} and \citet{zhou2021evaluating}, expecting \glspl{explanans} to be unambiguous. 

We include the Plausibility desideratum, which encompasses the idea that \glspl{explanation} should align with human knowledge and intuition.
On one hand, this includes human expectations towards the result (\gls{explanans}), e.g., ``The model focuses on what a human would focus on''.
On the other hand, the XAI methods' behavior (\gls{explanation}) should also be aligned with human intuition, e.g., ``The outputs for individual inputs should differ''.

\subsubsection{Coverage}
\begin{definitionbox}{Definition: Coverage}
The \gls{explanation} should provide an \gls{explanans} for every \gls{explanandum}.
\end{definitionbox}%
\noindent
The extent to which an \gls{explanation} or \gls{explanans} can be applied is considered by two frameworks.
Unfortunately, definitions from both surveys are vague.
\citet{markus2021role} and \citet{zhou2021evaluating} define the Broadness of an \gls{explanation} as ``how generally applicable'' it is, without further elaboration on the implications of this definition.
More concretely, Representativeness is presented by \citet{robnik2018perturbation}, \citet{carvalho2019machine}, and \citet{molnar2020interpretable}. 
It reflects the number of \glspl{explanandum} that are covered by an individual \gls{explanans}, although this definition focuses mainly on the distinction between global and local \gls{explanation} methods.

To add more clarity to these definitions, we include Coverage with an alternative definition.
It defines the amount of \glspl{explanandum} that are covered by the \gls{explanation}, i.e., reflecting whether there exists an \gls{explanans} for every data input or output.

\subsubsection{Fidelity}
\begin{definitionbox}{Definition: Fidelity}
The \gls{explanation} should make the \gls{explanans} reflect the model's true reasoning.
\end{definitionbox}%
\noindent
Fidelity is one of the most frequently discussed concepts in the literature and combines two closely related aspects: Correctness and Completeness. While some works introduce these as separate desiderata, others group them together under the umbrella of Fidelity.

Correctness refers to whether the \gls{explanation} truthfully represents the internal logic and decision process of the black-box model.
It is one of the most frequently emphasized desiderata across the reviewed frameworks. 
Without correctness, even the most interpretable or simple \gls{explanation} may provide no meaningful insight.
Terms like Faithfulness, Truthfulness, and Fidelity are often used interchangeably in literature to describe this idea.
Correctness encompasses both local fidelity for individual \glspl{explanans} and global alignment across the dataset \citep{robnik2018perturbation, carvalho2019machine, molnar2020interpretable}. 
The general consensus is that an \gls{explanation} should reveal what truly drives the model's outputs \citep{alvarez2018towards, markus2021role, zhou2021evaluating, belaid2022we, alangari2023exploring, nauta2023anecdotal}.
It is commonly assessed by how well the \gls{explanation} reflects or mimics the model's behavior \citep{andrews1995survey, johansson2004truth, guidotti2018survey}.

Completeness, in contrast, describes how much of the model's reasoning is captured by the \gls{explanation}.
According to the Co-12 properties by \citet{nauta2023anecdotal}, an \gls{explanation} should ideally include the full scope of the model's rationale.
Some authors treat Completeness as a sub-aspect of Fidelity \citep{markus2021role, zhou2021evaluating}, while others define Fidelity itself as the capacity to capture all of the information embodied in the model \citep{andrews1995survey, johansson2004truth, guidotti2018survey}.

Although it is theoretically possible to have an \gls{explanation} that is partially correct but incomplete (e.g., providing a heatmap that highlights only one of several relevant features), or complete but partially incorrect (e.g., including all the right features alongside irrelevant ones), neither scenario is desirable. 
If key features are missing or irrelevant ones are included, the \gls{explanans} ultimately misrepresents the model's behavior.
While Correctness and Completeness can be distinguished conceptually, they are tightly interwoven in practice and difficult to evaluate in isolation. Since our desiderata are intended to capture orthogonal evaluation dimensions, and these two cannot be meaningfully disentangled, we combine them under the unified criterion of Fidelity.

\subsubsection{Continuity} 
\begin{definitionbox}{Definition: Continuity}
The \gls{explanation} should ensure that similar \glspl{explanandum} yield similar \glspl{explanans}.
\end{definitionbox}%
\noindent
Just as the robustness or stability of a standard AI model is of great interest, similar expectations apply to explainability.
Most frameworks highlight this desideratum, and various metrics have been proposed to assess how stable or reliable \glspl{explanation} are. 
However, the terminology used in literature is inconsistent, at times overlapping and at other times diverging in meaning.

\citet{nauta2023anecdotal} introduce the term Continuity as the smoothness of the \gls{explanation}, i.e., similar \glspl{explanandum} should yield similar \glspl{explanans}.
Others refer to this idea as Stability \citep{robnik2018perturbation, carvalho2019machine, molnar2020interpretable}, describing it as the resilience against slight variations in input features that do not alter the model's prediction.
The term Robustness is used by \citet{alvarez2018towards}, \citet{jesus2021can}, and \citet{alangari2023exploring} to describe the same behavior, referring to it as a key requirement for trustworthy XAI.

The Quantus toolkit reflects the prevalence of this concept, providing a ``Robustness'' metric category \citep{hedstrom2023quantus, bommer2024finding}, which assesses the similarity of \glspl{explanans} under minor changes in input.
Finally, \citet{belaid2022we} cover the same idea under the term Stability.
In addition, they assess Fragility, which they define as the resilience of \glspl{explanation} against malicious manipulation, such as adversarial attacks.

Our Continuity desideratum covers both of the mentioned properties.
It includes the smoothness of \glspl{explanation} with respect to ``na\"ive'' changes in the \gls{explanandum} that ideally do not affect the model's behavior, as well as the resilience of \glspl{explanation} against malicious manipulation attempts.
Note that this includes changes over the input data as well as the model.
We adopt the term \textit{Continuity} instead of Stability or Robustness to reduce possible confusion with model robustness.

\subsubsection{Consistency}
\label{sec:des_consistency}
\begin{definitionbox}{Definition: Consistency}
The \gls{explanation} should produce stable \glspl{explanans} across repeated evaluations.
\end{definitionbox}%
\noindent
While Continuity investigates the smoothness and similarity between similar but different inputs, the Consistency of \glspl{explanation} for identical inputs also needs to be considered. However, few authors explicitly consider this desideratum. 

Consistency is introduced by \citet{nauta2023anecdotal} as a direct measure of the determinism of an XAI algorithm.
Similarly, one part of the definition of Stability by \citet{robnik2018perturbation}, \citet{carvalho2019machine}, and \citet{molnar2020interpretable} considers variations in \glspl{explanation} based on non-determinism.
The oldest formulation of Consistency is given by \citet{andrews1995survey} and considers \gls{explanation} methods to be consistent when they produce equivalent results under repetition.

However, several frameworks additionally consider the similarity of \glspl{explanans} generated from different models trained on the same data \citep{robnik2018perturbation, carvalho2019machine, molnar2020interpretable, nauta2023anecdotal}.
Yet, different models can produce the same prediction while relying on entirely different internal reasoning.
This is especially true as there are often multiple valid reasons for the same event, also known as the \textit{Rashomon Effect} \citep{breiman2001statistical, leventi2022rashomon}. 

We include Consistency using the initial formulations, i.e., \glspl{explanation} should be deterministic or self-consistent, always presenting the same \gls{explanans} for identical \glspl{explanandum}. 
While the latter definition is present in one of the identified metrics, we do not explicitly add it to the definition of our Consistency desideratum, as we do not believe that different \glspl{explanandum} (inputs), i.e., different models, necessarily result in identical \glspl{explanans} (outputs).

\subsubsection{Efficiency}
\begin{definitionbox}{Definition: Efficiency}
The \gls{explanation} should compute the \gls{explanans} efficiently and broadly.
\end{definitionbox}%
\noindent
Finally, out of practical considerations, we want \glspl{explanation} to be conveniently applicable.
This includes considering the range of models or situations in which the algorithm can be effectively applied. Simultaneously, it also includes the time it takes to compute an individual \gls{explanans}.

The first property is introduced as Portability and Translucency throughout literature \citep{robnik2018perturbation, carvalho2019machine, molnar2020interpretable}.
Portability is the variety of models for which an \gls{explanation} can be used, while Translucency is the necessity of the \gls{explanation} algorithm to have access to the internals of the model. 
Similarly, \citet{johansson2004truth} measure Generality, given by the restrictions or overhead necessary to apply an \gls{explanation} to specific models.
\citet{belaid2022we} refer to Portability as the diverse set of models to which the \gls{explanation} can be applied.

Secondly, the Algorithmic Complexity \citep{robnik2018perturbation, carvalho2019machine, molnar2020interpretable} considers the time it takes to generate an \gls{explanans}.
Naturally, the necessary time depends not only on the inherent complexity of the \gls{explanation} algorithm but also on Scalability, i.e., its ability to efficiently handle larger models and input spaces \citep{johansson2004truth}. 
Using the ``Stress test'', \citet{belaid2022we} explicitly evaluate the runtime behavior with respect to increasing input size.

We subsume both of these aspects under a general desideratum called Efficiency.
It includes the algorithmic or computational properties of the \gls{explanation}, which might influence the choice of a specific XAI algorithm over another.

\subsubsection{Excluded Desiderata}
\label{app:excluded_desiderata}
Several of the desiderata introduced in the referenced frameworks were not included in our catalog.
We briefly present each of these and justify why they were excluded.

In the Co-12 properties for XAI, \citet{nauta2023anecdotal} introduce \textbf{Controllability}, which we exclude for two reasons. 
First, it measures the extent to which a user can interact with the \gls{explanans}, which can be relevant in given applications, but is a property of the \textit{presentation} and not necessarily the \gls{explanans} itself.
Second, the only measurement they provide is through user interaction, which we avoid in this study, as we focus on functionality-grounded metrics. 
Their \textbf{Composition} property is not measurable either, as it describes the \gls{explanans}' presentation format. 
Similarly, \textbf{Expressive Power} \citep{robnik2018perturbation, carvalho2019machine, molnar2020interpretable} describes the format of the \gls{explanans} and is therefore not included.

The \textbf{Context} property \citep{nauta2023anecdotal} describes how relevant a given \gls{explanans} is to a user.
This is mostly covered in the Plausibility desideratum, although their definition is more focused on the needs of specific users and hence is very subjective and not measurable through proxies.

Further, both \textbf{Confidence} \citep{nauta2023anecdotal} and the closely related \textbf{Certainty} \citep{robnik2018perturbation, carvalho2019machine, molnar2020interpretable} are excluded. 
They both describe whether confidence scores of the model's decisions or the \gls{explanans} are displayed.
However, this is not a general desideratum for \glspl{explanation}, but can rather be seen as providing assisting information, which is also an \gls{explanans} in itself, and is then subject to all the desiderata of our framework.
Related to this, \textbf{Novelty} \citep{robnik2018perturbation, carvalho2019machine, molnar2020interpretable} refers to whether data instances lie within or outside the training distribution.
While relevant to assessing model behavior or uncertainty, this property does not characterize the quality of the \gls{explanation} itself. Instead, it serves as contextual information that may inform or accompany an \gls{explanans}, but is not an intrinsic requirement for explainability.

As discussed in \autoref{sec:des_consistency}, we do not include some of the aspects of \textbf{Consistency} as defined by \citet{robnik2018perturbation}, \citet{carvalho2019machine}, and \citet{molnar2020interpretable}, who expect several models to generate similar \glspl{explanans}.
This is rooted in the assumption that different models should all follow the same reasoning, which we reject.
In contrast, \citet{anders2020fairwashing} showed that two models exhibiting the same external behavior, such as predictions, can have radically different inner workings.

While it is a standard metric during model training, \textbf{Accuracy} of \glspl{explanation} is also reported by several frameworks \citep{andrews1995survey, johansson2004truth, guidotti2018survey, robnik2018perturbation, carvalho2019machine, molnar2020interpretable}. Importantly, this is only meaningful for specific types of \glspl{explanation}, such as white-box surrogate models (e.g., decision trees; see \autoref{sec:explanation_types} for a definition). 
It is different from ``surrogate fidelity'', since this measures performance against the black-box model's predictions and not the ground truth.
We restrict our framework to \glspl{explanans} derived from black-box models, rather than evaluating interpretable models that are directly trained on the data as predictive models themselves.

Lastly, the Quantus toolkit \citep{hedstrom2023quantus, bommer2024finding} includes a category labeled as \textbf{Axiomatic}, which groups metrics based on their functionality. 
Specifically, it refers to whether they test for formal properties.
This categorization reflects the underlying mechanism of the metric rather than a specific quality criterion.
Accordingly, we reassign these metrics to the corresponding desiderata they evaluate.

\subsection{Considerations}
\label{sec:desiderata_considerations}
We conclude this presentation of desiderata with a few final considerations.

First, all seven desiderata are introduced as independent dimensions.
However, in practice, they cannot always be separated entirely, which is reflected in some metrics being associated with multiple desiderata.
Especially where a single criterion in other authors' frameworks covers multiple of our desiderata, an inherent connection is given.
This can be seen especially in Consistency and Continuity, which both measure some sort of robustness. 
Conversely, some of our desiderata subsume multiple sub-aspects themselves. 
This applies especially to Fidelity, which is comprised of Correctness and Completeness.

Second, a truly useful \gls{explanation} can only be achieved when considering the interplay between the individual desiderata. 
Otherwise, trivial \glspl{explanans} can be constructed to optimize individual desiderata, as we illustrate with two examples: 
a) A saliency map that highlights the entire input image is certainly complete, as it encompasses all relevant features; however, it is hardly parsimonious and unlikely to be correct, as not every region is truly decisive for the prediction.
b) A concept \gls{explanation} that identifies the same concept regardless of the input may be maximally consistent, continuous, and parsimonious; nevertheless, it will unlikely be either correct or complete, as it disregards input-specific variation.

Further, to ensure the usefulness of XAI to humans, we believe that some hierarchy exists between our defined desiderata. 
While an exact ordering is implausible, as it depends on the exact setting and needs, at least some general guidelines can be given:
\textbf{Fidelity} is the foundation of the entire evaluation process.
If there is no necessity for an \gls{explanation} to be correct and complete, it can be completely arbitrary and will not necessarily give any true insights into the underlying model. 
This would undermine the entire purpose of XAI. 
If important parts of the rationale are excluded in the \gls{explanans}, it may result in an incomplete (and hence incorrect) understanding of the decision process.
While not as crucial, but still important, \textbf{Consistency} and \textbf{Continuity} considerably contribute to the trustworthiness of XAI.
Without them, \glspl{explanation} may vary unpredictably, as \glspl{explanans} can change without reasons rooted in the model's behavior.
Given these four desiderata, the \gls{explanation} can be expected to give solid results from a technical viewpoint.
Hence, the interpretability dimensions, \textbf{Parsimony} and \textbf{Plausibility}, can be considered next, as they facilitate understanding but do not assess whether an \gls{explanation} is actually reliable. 
Finally, \textbf{Coverage} and \textbf{Efficiency} are useful properties of an \gls{explanation}, but can be seen as a bonus, making them readily applicable to a wide set of \glspl{explanandum}, including data and models.
However, depending on the use case and context, individual desiderata may be weighted differently or even disregarded entirely, particularly when some aspects are less relevant for a specific application or domain.

Finally, the framework is based on the works presented by other authors, as well as the metrics we identified through our systematic survey.
While we consider it extensive for now, it is possible to expand it horizontally in the future.
Depending on the requirements, further desiderata may be added.
We do not rigorously map each metric to a single desideratum, but instead list several desiderata it contributes to. Therefore, our proposed categorization can be readily extended with further desiderata.

%%
%%%%%%%%%%%%%

\clearpage

\setcounter{footnote}{0}
\renewcommand{\thefootnote}{\alph{footnote}}

\section{VXAI Metrics}
\label{app:metrics}

This appendix details all metrics identified in our literature survey and included in the VXAI framework, as summarized in \autoref{tab:metric_overview}.
An interactive version of the framework is available at \vxailink.

\subsection{Notation}
The mathematical expressions used throughout the appendix are intended to clarify concepts. 
We do not enforce strict formalism as long as the notation remains unambiguous.
In cases where clarity is not compromised, we slightly overload notation, but specific meanings are always disambiguated by context.
While the notation is predominantly tailored to classification tasks, where the model assigns scores to discrete class labels, it can be adapted to suit regression settings or, in part, other paradigms.
We define the following symbols, which are used across all metrics:

\begin{itemize}
    \item Let $\mathcal{X}$ and $\mathcal{Y}$ denote the input (data) and output (label) spaces, respectively.
    \item The black-box model is defined as a scoring function $\theta : \mathcal{X} \times \mathcal{Y} \to \mathbb{R}$ that assigns a real-valued predictive score (e.g., logits) to each input–label pair.
    \item For a given input $x \in \mathcal{X}$, we write $\theta(x)$ as shorthand for the vector of scores across all labels in $\mathcal{Y}$, i.e., $\theta(x) := (\theta(x, y_1), \theta(x, y_2), \dots, \theta(x, y_k))$.
    \item The predicted label is then given by $\hat{y}_x := \arg\max_{y \in \mathcal{Y}} \theta(x, y)$.
    \item We denote a second input as $x' \in \mathcal{X}$ and a perturbed version of $x$ as $\dot{x}$ (see \autoref{sec:perturbations}).
    \item Let $\mathcal{X}_y \subseteq \mathcal{X}$ be the set of all instances with ground-truth class $y$.

    \vspace{0.8\baselineskip}
     \item An \gls{explanandum} is a triple $(\theta, x, y)$, where $\theta$ is the model, $x \in \mathcal{X}$ is the input instance, and $y$ is the label to be explained (which is often, but not necessarily, $\hat{y}$).
    \item An \gls{explanans} for a given \gls{explanandum} is denoted $e_{\theta, x, y}$, or simply $e$ when unambiguous.
    \item For \glspl{ExE}, we write $z := e^\text{(\gls{ExE})}_{\theta, x, y^*_z}$, where $z \in \mathcal{X}$ is typically a counterfactual input such that ${\hat{y}_z = y^*_z}$.
    \item For \gls{WBS}, the surrogate model is denoted $\theta^e := e^\text{(\gls{WBS})}_{\theta, x, \hat{y}}$. The surrogate returns a label prediction for arbitrary input via $\hat{y}^e_{x'} := \theta^e(x')$.

    \vspace{0.8\baselineskip}
     \item $\delta(\cdot, \cdot)$ denotes an arbitrary distance or dissimilarity function, which may be applied to inputs, explanations, or predictions depending on context. Similarity measures (see \autoref{sec:res_similarity}) may be applied by inverting them appropriately (e.g., via negation or reciprocal).
    \item $|\cdot|$ indicates size or cardinality; $\|\cdot\|_p$ denotes the $L_p$-norm.
    %\item $\mathbbm{1}[\cdot]$ is the indicator function, returning $1$ if the condition holds and $0$ otherwise.
    \item $k$ refers to a generic parameter (e.g., number of selected features, neighbors, or examples); $\epsilon$ denotes a small threshold or tolerance constant.
    \item $\Phi_y$ denotes an autoencoder trained specifically on inputs from class $y$ (i.e., $x \in \mathcal{X}_y$).

\end{itemize}

\noindent
For convenience, we recall the abbreviations of the \gls{explanation} types introduced in \autoref{sec:categorization_scheme}:

\glsreset{FA}
\glsreset{CE}
\glsreset{ExE}
\glsreset{WBS}
\glsreset{NLE}

\begin{itemize}
    \item \gls{FA}
    \item \gls{CE}
    \item \gls{ExE}
    \item \gls{WBS}
    \item \gls{NLE}
\end{itemize}

\noindent 
Additionally, we adopt the following terminology:
\textbf{Metrics} refer to the individual or aggregated VXAI criteria we evaluate and identified from the literature survey.
\textbf{Measures} denote performance-scoring functions (e.g., accuracy, similarity, overlap), often also called metrics in literature; but we use the term ``measure'' to distinguish them clearly.

To improve readability while maintaining completeness throughout the Appendix, we move extensively long reference lists into footnotes using the notation: \citenote{Author et al.}.

\subsection{Helper Functions}
The following components are recurring elements that serve as functional building blocks across multiple metrics. 
They can be understood as interchangeable parameters that shape how specific evaluation metrics are instantiated and interpreted.

\subsubsection{Perturbation Approach}
\label{sec:perturbations}
Perturbations are small changes typically applied to input features and are recurring components in metrics targeting Fidelity and Continuity. 
There exists a wide range of perturbation strategies, and the choice of approach can significantly affect both metric results and their interpretation \citep{brunke2020evaluating,funke2022zorro, rong2022consistent}.

We distinguish first by the \textbf{Perturbation Scope}, i.e., the parts of the input that are modified.
Perturbations can be applied at a fine-grained level (e.g., individual features) or on more structured, high-level groupings.
For image data, scope definitions may involve aggregating pixels into fixed grids \citep{schulz2020restricting} or segmenting into superpixels \citep{ribeiro2016should, kapishnikov2019xrai, rieger2020irof}.
In time-series data, one may perturb fixed-length windows, with the target time-step at the beginning or middle \citep{schlegel2019towards, schlegel2020empirical}.
In topologically ordered domains (e.g., images, time-series, or graphs), adjacent features can be perturbed together, such as modifying the area surrounding a focal pixel \citep{samek2016evaluating, brahimi2019deep}.
Higher-level approaches include perturbing Concepts \citep{el2021towards} or internal activations associated with object parts \citep{zhang2019interpreting}.

Once the scope is defined, a variety of \textbf{Perturbation Functions} can be applied.
In fact, any perturbation function might be suitable  \citep{hameed2022based, schlegel2023deep}.
However, here we present some of the most common in literature.
At the simplest level, features may be removed by setting them to zero or dropping them entirely, especially in structured domains such as graphs, text, or time-series data, as employed by many authors \citenote{\citet{bach2015pixel, ancona2017towards, alvarez2018towards, chu2018exact, arya2019one, deyoung2019eraser, schlegel2019towards, cong2020exact, singh2020valid, warnecke2020evaluating, bajaj2021robust, faber2021comparing, singh2021extracting, jin2023guidelines}}.
Alternatively, features can be replaced by a fixed value, e.g., the per-channel or per-instance mean \citep{petsiuk2018rise, schlegel2019towards, schlegel2020empirical, hameed2022based, jin2023guidelines}.

To generate less deterministic perturbations, authors propose adding random noise (e.g., Gaussian) or drawing values from uniform distributions \citep{yeh2019fidelity, bhatt2020evaluating, sturmfels2020visualizing, bajaj2021robust, funke2022zorro, veerappa2022validation}.
Other strategies leverage the spatial structure of the data: for instance, applying blurring or interpolation \citep{sturmfels2020visualizing, rong2022consistent}, or reordering spatial regions \citep{schlegel2019towards, chen2020generating}.
Instead of applying synthetic noise, values can be resampled from the marginal distribution of a feature, from its nearest neighbor, or even from an opposite-class example \citep{guo2018lemna, hameed2022based}.
Where influence regions are known, perturbations can be constrained to lie inside or outside these intervals \citep{velmurugan2021developing}.

In the NLP domain, word embeddings can be noised, or tokens substituted using synonym sets and domain knowledge \citep{yin2021sensitivity}. 
For image inputs, another strategy is cropping and resizing to emphasize or suppress local information \citep{dabkowski2017real}.

Finally, when perturbations are guided by a \gls{FA}, their intensity can be scaled proportionally to the assigned importance scores \citep{chattopadhay2018grad, guo2018lemna, jung2021towards}.

\subsubsection{Normalization of Explanantia}
Since \glspl{FA} and \glspl{CE} are typically represented as real-valued vectors, computed through various mechanisms, their value ranges are not inherently standardized. 
However, many metrics either explicitly require the \gls{explanans} to lie within a fixed range or implicitly assume comparability across \glspl{explanans}, making normalization a necessary preprocessing step.
A widely used normalization method is Min-Max Scaling \citep{binder2023shortcomings, brandt2023precise}, which maps all values into a fixed interval (typically $[0, 1]$). 
Alternative strategies include normalization based on the square root of the average second-moment estimate, offering robustness to outliers and variance shifts \citep{binder2023shortcomings}.
This limited selection can be extended through any suitable normalization approach.

\subsubsection{Similarity Measures}
\label{sec:res_similarity}
Across various metrics, it is necessary to calculate similarities or distances between two \glspl{explanans}, especially when concerned with the desideratum of Continuity and in metrics relying on Ground-Truth evaluations.
Throughout the reported literature, various approaches have been reported, which differ based on the type of \gls{explanation}.
While some measures directly compute similarity, others quantify distance or disparity. 
In this work, we adopt a similarity-based framing, either directly or by transforming distance measures, to ensure that higher values uniformly indicate greater explanatory agreement.
Analogously, we can compute the similarity between \glspl{explanandum}, adopting measures that are presented in the following.

For \textbf{\glspl{FA}}, similarity may be measured using arbitrary inverted distance or loss measures (e.g., $L_p$, MSE, cosine distance, JS-Divergence, or Bhattacharyya Coefficient), potentially normalized (e.g., by standard deviation) \citenote{\citet{alvarez2018robustness, alvarez2018towards, chu2018exact, wu2018faithful, jain2019attention, jia2019improving, mitsuhara2019embedding, pope2019explainability, trokielewicz2019perception, yeh2019fidelity, zhang2019towards, jia2020exploiting, agarwal2022rethinking, atanasova2022diagnostics, dai2022fairness, fouladgar2022metrics, agarwal2023evaluating, huang2023safari, nematzadeh2023ensemble}}.
Alternatively, rank correlation measures such as Spearman's or Kendall's Tau can be applied \citenote{\citet{das2017human, adebayo2018sanity, chen2019robust, dombrowski2019explanations, ghorbani2019interpretation, nguyen2020quantitative, rajapaksha2020lormika, sanchez2020evaluating, liu2021synthetic, yin2021sensitivity, krishna2022disagreement, huang2023safari}}.
When binarizing \gls{FA} outputs through thresholding, feature-wise evaluation measures such as accuracy, precision, $F_1$, or AUROC are commonly used \citenote{\citet{chen2018neural, yang2018explaining, jia2019improving, jia2020exploiting, sanchez2020evaluating, bykov2021explaining, joshi2021explainable, park2021comparing, amoukou2022accurate, chen2022can, funke2022zorro, tjoa2022quantifying, wilming2022scrutinizing, agarwal2023evaluating}}.
Similarly, IoU can be calculated over binarized features \citenote{\citet{oramas2017visual, fan2020can, kim2021sanity, situ2021learning, vermeire2022explainable}}, or top-$k$ intersection measures may be used \citenote{\citet{ghorbani2019interpretation, mishra2020reliable, rajapaksha2020lormika, warnecke2020evaluating, amparore2021trust, bajaj2021robust}}.
For saliency maps, specialized similarity measures are available, such as SSIM \citenote{\citet{adebayo2018sanity, dombrowski2019explanations, rebuffi2020there, graziani2021evaluation, sun2023improving}}, Earth Movers Distance \citep{park2018multimodal, wu2018faithful}, Normalized Cross Correlation \citep{baumgartner2018visual, bass2020icam}, or Mutual Information \citep{sun2023improving}.
While primarily established for \glspl{FA}, similar similarity functions can be naturally applied to \textbf{\gls{CE}}-based \glspl{explanation} as well.

For \textbf{\glspl{WBS}} and \textbf{\glspl{ExE}}, the choice of similarity measure depends strongly on the underlying model or domain. 
For linear predictive models, coefficient mismatch is a common choice \citep{lakkaraju2020robust}, whereas rule- and tree-based \glspl{explanans} may be compared by their rule overlap, feature usage, or node structures \citenote{\citet{bastani2017interpreting, guidotti2019factual, lakkaraju2020robust, rajapaksha2020lormika, margot2021new}}.

\textbf{\glspl{NLE}} can be compared using standard natural language processing measures \citenote{\citet{camburu2018snli, chuang2018learning, liu2018towards, wu2018faithful, chen2019co, rajani2019explain, wickramanayake2019flex, li2020generate, sun2020dual, jang2021training, atanasova2024generating}}.
Those include for instance BLEU \citep{papineni2002bleu}, METEOR \citep{banerjee2005meteor}, ROUGE \citep{lin2004rouge}, CIDEr \citep{vedantam2015cider}, or SPICE \citep{anderson2016spice}.
 In addition, several measures have been proposed specifically for evaluating natural language \glspl{explanation} \citep{xie2021neural, du2022care, rodis2024multimodal, park2018multimodal}.

When comparing similarities over multiple instances, the most natural aggregation is to compute the mean similarity \citep{fan2020can, fouladgar2022metrics, yeh2019fidelity}.
Depending on the evaluation goal, alternative aggregation strategies may offer more informative insights. 
For example, worst-case stability,  defined as the minimum similarity across inputs,  can be used to quantify robustness \citenote{\citet{alvarez2018robustness, alvarez2018towards, yeh2019fidelity, yin2021sensitivity, fouladgar2022metrics}}.

\clearpage
\subsection{Metrics}
\renewcommand{\thesubsubsection}{\thesubsection.\Roman{subsubsection}}

\subsubsection{Explanans-Centric}
\setMetricPrefix{I}
The following metrics directly evaluate the \gls{explanans}, potentially looking into the data inputs but without having any access to the underlying model.

\begin{Metric}{}{Explanans Size}{Parsimony}{\gls{FA}, \gls{ExE}, \gls{WBS}, (\gls{CE}, \gls{NLE})}{51}{
\citet{craven1995extracting, stefanowski2001induction, nauck2003measuring, alonso2008hilk, augasta2012reverse, lakkaraju2016interpretable, samek2016evaluating, zilke2016deepred, lakkaraju2017interpretable, guidotti2018survey, hara2018making, rustamov2018interpretable, wang2018interpret, wang2018multi, wang2018reinforcement, wu2018sharing, wu2018beyond, deng2019interpreting, evans2019s, fong2019understanding, guidotti2019factual, ignatiev2019abduction, lakkaraju2019faithful, polato2019interpretable, pope2019explainability, shakerin2019induction, slack2019assessing, topin2019generation, verma2019lirme, wang2019gaining, yoo2019edit, bhatt2020evaluating, chalasani2020concise, molnar2020history, nguyen2020quantitative, panigutti2020doctor, rajapaksha2020lormika, rawal2020beyond, stepin2020generation, warnecke2020evaluating, wu2020regional, liu2021multi, margot2021new, moradi2021post, poppi2021revisiting, rosenfeld2021better, samek2021explaining, dai2022fairness, funke2022zorro, huang2023global,stevens2024explainability}
}
\label{met:explanans_size}
The size of an \gls{explanans} $|e|$ is a common indicator of its complexity.
Smaller or more compact \glspl{explanans} are generally easier to understand and more plausible to human users.
 The exact method to measure size depends on the \gls{explanation} type and context.

A generally applicable method is to compute the file size (in bytes) of the \gls{explanans}, based on the assumption that sparse \glspl{explanans} can be more easily compressed \citep{samek2016evaluating, samek2021explaining}.

For \textbf{\gls{WBS}}, size is typically measured via structural properties of the surrogate model:
tree-based models are assessed by depth, number of nodes, or number of leaves \citenote{\citet{craven1995extracting, alonso2008hilk, guidotti2018survey, hara2018making, wu2018beyond, evans2019s, slack2019assessing, yoo2019edit, molnar2020history, rawal2020beyond, wu2020regional}}, while rule-based systems are evaluated using the number of rules or predicates per rule \citenote{\citet{craven1995extracting, stefanowski2001induction, nauck2003measuring, alonso2008hilk, augasta2012reverse, lakkaraju2016interpretable, zilke2016deepred, lakkaraju2017interpretable, wang2018multi, wu2018sharing, deng2019interpreting, lakkaraju2019faithful, polato2019interpretable, shakerin2019induction, wang2019gaining, panigutti2020doctor, rajapaksha2020lormika, stepin2020generation, margot2021new, moradi2021post, rosenfeld2021better}}.
For explanation-graphs, the number of nodes and edges can serve as a proxy for size \citep{rustamov2018interpretable, topin2019generation}.
Conversely, the (relative) number of instances covered per rule can also express parsimony \citep{deng2019interpreting, guidotti2019factual}.

In \textbf{\gls{FA}}, size is typically based on the number of relevant features.
This may be computed via:
\begin{itemize}
    \item The $L_0$ norm \citenote{\citet{wang2018reinforcement, wang2018interpret, fong2019understanding, polato2019interpretable, verma2019lirme, nguyen2020quantitative, poppi2021revisiting, rosenfeld2021better, stevens2024explainability}},

    \item A count of features exceeding a relevance threshold \citenote{\citet{ignatiev2019abduction, pope2019explainability, warnecke2020evaluating, liu2021multi, dai2022fairness}},

    \item Normalization by input dimensionality, e.g., in graph settings \citep{pope2019explainability},

    \item Integration over multiple thresholds to form a size curve \citep{warnecke2020evaluating}, or

    \item Threshold-free statistics like entropy \citep{samek2016evaluating, bhatt2020evaluating, funke2022zorro} and Gini index \citep{chalasani2020concise}.
\end{itemize}

While not common in the literature,  for \textbf{\gls{CE}}, similar counting mechanisms may be applied.
One may count concepts exceeding a relevance threshold or use the total number of tested concepts (which is not tied to input size).

In \textbf{\gls{ExE}}, size naturally corresponds to the number of examples used in the \gls{explanans} \citep{nguyen2020quantitative, huang2023global}.

For \textbf{\gls{NLE}}, the number of words or sentences provides a straightforward measure of size.

In general, size can be measured per-instance or aggregated over the dataset.
For instance, one can report the number of predicates in the rule explaining a single instance \citep{augasta2012reverse}, or aggregate the  number of predicates across a global rule set with statistics such as the average \citep{lakkaraju2016interpretable, lakkaraju2017interpretable, lakkaraju2019faithful}, sum \citep{margot2021new, moradi2021post}, or maximum \citep{moradi2021post}.
Similarly, rule counts can be aggregated over classes (e.g., average number per class \citep{nauck2003measuring}).
Optional adjustments include applying a tolerance margin (e.g., $\max(0, |e| - k)$ \citep{rosenfeld2021better}) or aggregating over multiple class-specific \glspl{explanans} per instance \citep{pope2019explainability}.

\end{Metric}

\begin{Metric}{}{Overlap}{Parsimony}{\gls{WBS}}{5}{
\citet{lakkaraju2016interpretable, lakkaraju2017interpretable, lakkaraju2019faithful, moradi2021post, hosain2024explainable}}
\label{met:overlap}
The overlap within a rule-based \gls{explanans} (e.g., rule sets or decision trees) can be measured with respect to either the input space or the rules themselves.
A lower degree of overlap is generally preferred, as it implies more distinct, non-redundant rules and enhances interpretability.

\begin{itemize}[topsep=0pt]
\item \textbf{Input Overlap}: Measures how often input instances are covered by multiple rules or decision paths \citep{lakkaraju2016interpretable, lakkaraju2017interpretable, lakkaraju2019faithful, hosain2024explainable}.
High overlap may indicate redundant or conflicting logic. Overlap can also be broken down by class label to distinguish intra-class from inter-class coverage.

\item \textbf{Rule Overlap}: Measures how many rules share identical or highly similar predicates \citep{moradi2021post}, indicating structural redundancy within the rule set.
\end{itemize}
\end{Metric}

\begin{Metric}{}{Explanans Cohesion}{Parsimony, Plausibility}{\gls{FA},(\gls{CE})}{2}{
\citet{fong2017interpretable, saifullah2024privacy}}
\label{met:explanans_cohesion}
In domains where input features are ordered (e.g., images or time series, as opposed to tabular data), we expect relevant information to be located in a smooth and coherent region of the input.
This property can be evaluated in two complementary ways.

First, \citet{fong2017interpretable} assess the locality of the \gls{explanans}: a higher degree of cohesion is achieved when relevant information lies within a small, contiguous region of interest, making the \gls{explanans} more condensed and easier to interpret.
In saliency maps, this can be quantified by computing the area of the smallest bounding box that encloses the thresholded \gls{explanans}.

Second, \citet{saifullah2024privacy} assess the smoothness of the \gls{explanans}: a more continuous attribution map is often easier to understand.
This can be measured by summing the absolute differences in attribution between neighboring features (e.g., in x- and y-directions for images).
A higher score indicates a more fragmented, less interpretable \gls{explanans}.

Both variants are applicable to data with ordered features and extend naturally to temporal domains.
While originally proposed for \glspl{FA}, these metrics can also be applied to \gls{CE} when concept-based saliency maps are available (e.g., as presented by \citet{lucieri2020explaining}).
\end{Metric}

\begin{Metric}{}{Minimality}{Parsimony, Plausibility}{\gls{ExE}}{25}{
    \citet{tolomei2017interpretable, wachter2017counterfactual, karlsson2018explainable, zhang2018interpretingnn, guidotti2019factual, albini2020relation, artelt2020convex, dandl2020multi, kanamori2020dace, karimi2020model, le2020grace, mothilal2020explaining, pawelczyk2020learning, ramon2020comparison, sharma2020certifai, abrate2021counterfactual, hvilshoj2021quantitative, pawelczyk2021carla, rasouli2021analyzing, van2021interpretable, albini2022counterfactual, chou2022counterfactuals, bayrak2023pertcf, huang2023global, verma2024counterfactual}
    }
\label{met:minimality}
To ensure that counterfactuals are understandable and believable, the induced changes should be minimal.
This can be assessed by evaluating both the proximity of the counterfactual to the original instance and the sparsity of the changes.

\textbf{Proximity} is commonly measured as the distance $\delta(x, z)$ between the counterfactual $z$ and the original instance $x$.
A variety of distance measures are used, with the choice significantly impacting the results \citep{wachter2017counterfactual, artelt2020convex, bayrak2023pertcf, verma2024counterfactual}.
Common options include (feature-wise weighted) $L_p$ distances (especially $L_2$) \citenote{\citet{wachter2017counterfactual, artelt2020convex, le2020grace, mothilal2020explaining, sharma2020certifai, rasouli2021analyzing, chou2022counterfactuals, verma2024counterfactual}}, as well as Jaccard or cosine distance \citep{tolomei2017interpretable}, or combinations thereof \citep{karimi2020model, hvilshoj2021quantitative}.
Other notable choices are Mahalanobis distance \citep{artelt2020convex, kanamori2020dace, verma2024counterfactual}, Gower distance \citep{dandl2020multi, karimi2020model}, or feature-wise cumulative density-based distances \citep{pawelczyk2020learning}.
Dataset-specific alternatives include computing the quantile-shift per feature \citep{albini2022counterfactual}.
Different data types require suitable distance measures, e.g.: for graphs, the symmetric difference of adjacency matrices or cosine similarity between node features \citep{abrate2021counterfactual}; for images, the inverse SSIM score \citep{sharma2020certifai}; and for time series, distances measured per time step \citep{karlsson2018explainable}.

\textbf{Sparsity}, in contrast, concerns the number of features changed rather than the extent of change.
It is often quantified via the $L_0$ norm, that is, the number (or fraction) of altered features \citenote{\citet{albini2020relation, dandl2020multi, le2020grace, mothilal2020explaining, ramon2020comparison, pawelczyk2021carla, bayrak2023pertcf, verma2024counterfactual}}.

Bridging sparsity and proximity, measures like $L_1$ \citep{wachter2017counterfactual} or elastic-net regularization \citep{van2021interpretable} are sometimes used.
Depending on the representation, alternative definitions may apply, for example, counting the number of changed rules in rule-based counterfactuals \citep{guidotti2019factual}.
\end{Metric}

\begin{Metric}{}{Autoencoder Plausibility}{Plausibility}{\gls{ExE}}{1}{\citet{van2021interpretable}}
\label{met:autoencoder_plausibility}
Autoencoders are trained to capture the underlying structure of the dataset. 
Based on this idea, \citet{van2021interpretable} propose evaluating the plausibility of counterfactual examples via reconstruction errors from class-specific and general-purpose autoencoders.

Two specific scores are introduced:
\begin{itemize}
   \item \textbf{IM1} (Class-specific reconstruction comparison):
   This metric compares how well a counterfactual $z$ is reconstructed by an autoencoder trained on the target class ($\Phi_{y^*_z}$) versus the originally predicted class ($\Phi_{\hat{y}_x}$): 
   $\frac{\|z - \Phi_{y^*_z}(z)\|_2^2}{\|z - \Phi_{\hat{y}_x}(z)\|_2^2+ \epsilon}$. 

    A low IM1 score implies that the counterfactual is more representative of the target class than of its original class, indicating class-specific plausibility.

    \item \textbf{IM2} (General manifold plausibility):
    This score compares the reconstructions from a general autoencoder $\Phi_\mathcal{Y}$ and a class-specific one:
    $\frac{\|\Phi_\mathcal{Y}(z) - \Phi_{y^*_z}(z)\|_2^2}{\|z\|_1+ \epsilon}$.

    A low IM2 score implies that the counterfactual aligns well with both the general data manifold and the target class distribution, thus indicating higher plausibility.
\end{itemize}
While these metrics provide valuable insight into how realistic or semantically valid counterfactuals are, they come with caveats.
 Training multiple autoencoders (e.g., per class) introduces significant computational overhead.
 Further, reconstruction quality may vary across classes, and \citet{hvilshoj2021quantitative} showed that autoencoders can be sensitive to small perturbations, potentially undermining the consistency of the plausibility assessment.

\end{Metric}

\begin{Metric}{}{Diversity}{Plausibility}{\gls{ExE}}{6}{\citet{wachter2017counterfactual, karimi2020model, mothilal2020explaining, nguyen2020quantitative, stepin2020generation, verma2024counterfactual}}
\label{met:diversity}
In cases where multiple counterfactuals are generated for a single instance, they should be diverse to offer distinct and meaningful alternatives \citep{wachter2017counterfactual, stepin2020generation, verma2024counterfactual}.

Diversity is typically quantified by computing pairwise distances among the set of counterfactuals, either as the average pairwise distance \citep{mothilal2020explaining, nguyen2020quantitative}, or as the number of counterfactual pairs that exceed a predefined distance threshold \citep{karimi2020model}.
\end{Metric}

\begin{Metric}{}{Input Similarity}{Plausibility}{\gls{ExE}}{11}{\citet{laugel2019issues, laugel2019dangers, mahajan2019preserving, singla2019explanation, artelt2020convex, dandl2020multi, kanamori2020dace, delaney2021instance, pawelczyk2021carla, rasouli2021analyzing, smyth2022few}}
\label{met:input_similarity}
    To ensure that counterfactuals remain realistic and trustworthy, they should lie close to the true data manifold.
Counterfactuals that deviate significantly from the distribution of training data are unlikely to be plausible.
 Two principal approaches are used to evaluate this alignment: direct estimation of data conformity and distance-based proximity to the training data.

In the \textbf{direct approach}, several statistical or unsupervised methods determine whether the counterfactual is an outlier.
This includes calculating its likelihood under a kernel density estimator \citep{artelt2020convex} or from known data distributions in synthetic setups \citep{mahajan2019preserving}.
Other methods include Local Outlier Factor \citep{kanamori2020dace, delaney2021instance} and Isolation Forests \citep{delaney2021instance}, which can be applied either across the full dataset or restricted to samples of the counterfactual's target class \citep{artelt2020convex}.
\citet{pawelczyk2021carla} assess how many of a counterfactual's nearest neighbors share its class label.

Alternatively, proximity can be evaluated through \textbf{distance-based} approaches, relying on any suitable distance measure (or inverted similarity function, see \autoref{sec:res_similarity}).
These include distances to the $k$ nearest neighbors in the training data \citep{laugel2019issues, dandl2020multi}, or to the nearest training instances that share the same class label as the counterfactual or the original instance \citep{rasouli2021analyzing, smyth2022few}.
Specific domains may require tailored metrics such as the Fr\'echet Inception Distance (FID) for images \citep{singla2019explanation}.

Raw distance scores are often used directly \citep{dandl2020multi}, but some methods normalize distances, for example, by comparing the counterfactual–input distance $\delta(x,z)$ to average distances between random pairs $\delta(x',x'')$ in the dataset \citep{laugel2019issues, laugel2019dangers}, or to the original instance–counterfactual distance \citep{smyth2022few}.
\end{Metric}

\begin{Metric}{}{Input Contrastivity}{Plausibility, (Fidelity)}{\gls{FA}, (\gls{ExE}, \gls{CE}, \gls{WBS}, \gls{NLE})}{2}{\citet{honegger2018shedding, pornprasit2021pyexplainer}}
\label{met:input_contrastivity}
\Glspl{explanans} should be specific to their corresponding \glspl{explanandum} and not overly generic. That is, distinct inputs should typically yield distinct \glspl{explanans}.

To assess this, we calculate the fraction of instances or instance pairs that result in different \glspl{explanans} \citep{honegger2018shedding, pornprasit2021pyexplainer}.
A higher fraction of distinguishable \glspl{explanans} indicates greater contrastivity and, by extension, higher plausibility.

While this metric has been primarily proposed for \glspl{FA}, it may be applicable to other \gls{explanation} types as well.
\end{Metric}

\begin{Metric}{}{Actionability}{Plausibility, Fidelity}{\gls{ExE}}{7}{\citet{mahajan2019preserving, karimi2020model, le2020grace, pawelczyk2021carla, ma2022clear, smyth2022few, verma2024counterfactual}}
\label{met:actionability}
   For a counterfactual to be helpful, it should only introduce changes that are feasible or allowed. 
    This is often enforced by defining explicit constraints, such as:
    \begin{itemize}
        \item Immutable features (e.g., sex, age) \citep{karimi2020model, pawelczyk2021carla, smyth2022few, verma2024counterfactual}, or
    
        \item Value boundaries for individual features \citep{karimi2020model, le2020grace, ma2022clear, smyth2022few}.
    \end{itemize}
    
The actionability of a counterfactual set can then be assessed by calculating the fraction of counterfactuals that satisfy all constraints \citenote{\citet{karimi2020model, le2020grace, pawelczyk2021carla, ma2022clear, smyth2022few}}, or by computing the harmonic mean of satisfied constraints per counterfactual \citep{mahajan2019preserving}.

\end{Metric}

\begin{Metric}{}{Model-Agnostic Explanation Consistency}{Plausibility, (Consitency)}{\gls{FA}, \gls{ExE}, (\gls{CE}, \gls{WBS}, \gls{NLE})}{4}{\citet{fan2020can, nguyen2020model, hvilshoj2021quantitative, jiang2023formalising}}
\label{met:ma_explanation_consistency}
To assess whether \glspl{explanation} reflect generalizable patterns rather than model-specific artifacts  (such as adversarial shortcuts in counterfactual \glspl{explanation}), several authors evaluate \glspl{explanation} across different models trained on the same dataset.

In general, the approach involves training additional black-box models (oracles) on the same data and task. These oracles are then used to evaluate \glspl{explanation} from the original model.

One strategy is to directly compute the similarity between \glspl{explanans} generated by different models for the same input \citep{fan2020can}. While this has been reported for \glspl{FA}, it is applicable to any \gls{explanation} type, provided a suitable similarity metric is chosen (see \autoref{sec:res_similarity}).

Alternatively, \glspl{explanation} from the original model are evaluated using the oracles:
\begin{itemize}[topsep=0pt]
\item \citet{nguyen2020model} use perturbation-based evaluation (e.g., \metref{met:guided_perturb_F}) on a secondary model to assess whether the \gls{explanation} highlights features that are generally important across models.
\item \citet{hvilshoj2021quantitative} propose that counterfactuals should change the prediction in both the original model and the oracle. Only such counterfactuals are considered plausible.
\item An extension of this approach trains one oracle per class and computes the Jensen-Shannon-Divergence between its predictions on the original and counterfactual input. Ideally, only the target and original class should exhibit strong divergence \citep{hvilshoj2021quantitative}.
\item \citet{jiang2023formalising} define a neighborhood of models via small weight perturbations and count how many counterfactuals remain valid across all neighbors.
\end{itemize}

This approach might be extended to other \gls{explanation} types. 
However, its interpretive strength remains limited: it is unclear whether \glspl{explanation} should be similar across models, as differing model architectures may learn distinct (yet valid) rationales. 
As a result, high or low agreement does not always reflect \gls{explanation} quality, making this metric inherently context-dependent.
\end{Metric}

\begin{Metric}{}{Input Coverage}{Coverage}{\gls{FA}, \gls{ExE}, \gls{WBS}, (\gls{CE},\gls{NLE}}{8}{\citet{lakkaraju2016interpretable, lakkaraju2017interpretable, lakkaraju2019faithful, ribeiro2018anchors, rawal2020beyond, warnecke2020evaluating, moradi2021post, huang2023global}}
\label{met:input_coverage}
An \gls{explanation} method should ideally be capable of generating a valid \gls{explanans} for every input instance, regardless of its position on the data manifold.
To assess this, we calculate the fraction of inputs for which the method provides a valid result.

The notion of ``valid result'' is intentionally broad and depends on the \gls{explanation} type and the use case. A few prominent definitions include:
\begin{itemize}
    \item For perturbation-based \glspl{FA} like LIME (see \citet{ribeiro2016should}), a valid \gls{explanans} can be given if a sufficient number of perturbed samples belong to the target or opposite class, enabling a reliable local approximation \citep{warnecke2020evaluating}.
    \item For rule-based or tree-based \glspl{WBS}, coverage is the number of instances that are captured by at least one rule or decision path \citep{lakkaraju2016interpretable, lakkaraju2017interpretable, lakkaraju2019faithful, ribeiro2018anchors, rawal2020beyond}, optionally restricted to rules associated with the correct class \citep{moradi2021post}.
    \item For global \glspl{ExE}, coverage may be defined as the number of inputs that have at least one sufficiently similar explaining instance within a pre-defined distance \citep{huang2023global}.
\end{itemize}

\end{Metric}

\begin{Metric}{}{Output Coverage}{Coverage}{\gls{WBS}}{1}{\citet{lakkaraju2016interpretable}}
\label{met:output_coverage}
    Inherently interpretable \gls{WBS} models  should be capable of producing predictions for all possible output classes.
\citet{lakkaraju2016interpretable} evaluate the model's class-level coverage by calculating the fraction of target classes that appear in the \gls{explanans}.
This could, for instance, be the number of classes that are assigned to at least one rule in a rule set or to a leaf in a decision tree.

\end{Metric}

\begin{Metric}{}{Output Mutual Information}{Fidelity}{\gls{FA},\gls{CE}}{1}{\citet{nguyen2020quantitative}}
\label{met:output_MI}
When high-level features  (such as concepts or aggregated input features) are provided by an \gls{explanans}, they should capture as much information as possible about the model's output.
 In turn, the prediction should be reflected in the \gls{explanans}, implying a high degree of mutual dependency between both.

To quantify this relationship, \citet{nguyen2020quantitative} leverage the \textit{Mutual Information (MI)} (see \citet{cover1999elements}) between the \gls{explanans} and the output: $\textbf{MI}(e_{\theta,x,\hat{y}}, \theta(x))$.
Since MI is symmetric, a high score indicates that the \gls{explanation} both reflects (i.e. correctness) and encompasses (i.e., completeness) the relevant reasoning of the model.
\end{Metric}

\subsubsection{Model Observation}
\setMetricPrefix{II}
This set of metrics requires access to the model to obtain information about the model's output or its internal activations.

\begin{Metric}{}{Input Mutual Information}{Parsimony}{\gls{FA},\gls{CE}}{1}{\citet{nguyen2020quantitative}}
\label{met:input_MI}
When high-level features are provided by an \gls{explanans}, they should ideally represent an abstracted, human-understandable form of the input, omitting unnecessary detail.
This promotes parsimony by simplifying the input space.
Such features can be obtained either through feature selection in \glspl{FA} or through concept-level representations in \glspl{CE}.

To quantify this abstraction, \citet{nguyen2020quantitative} propose measuring the \textit{Mutual Information (MI)} (see \citep{cover1999elements}) between the input and the \gls{explanans}:  $\textbf{MI}(x,e_{\theta,x,\hat{y}})$
A lower mutual information score suggests that the \gls{explanans} captures less granular input detail and thus constitutes a more compact and parsimonious representation.
\end{Metric}

\begin{Metric}{}{Output Contrastivity}{Plausibility}{\gls{FA}, (\gls{ExE}, \gls{CE}, \gls{WBS}, \gls{NLE})}{5}
{\citet{nie2018theoretical, pope2019explainability, li2020quantitative, rebuffi2020there, sixt2020explanations}}
\label{met:output_contrastivity}
To enhance plausibility, \glspl{explanation} should be class-discriminative, that is, they should differ depending on the target class.
In image classification, for instance, the \glspl{explanans} supporting a ``car'' label should highlight different regions than those supporting ``dog''.
This makes it easier for humans to understand the specific rationale for each class.

Class discriminativeness is commonly measured by comparing \glspl{explanans} generated for different classes on the same input.
Several comparison setups have been proposed: between the most and least likely classes \citep{li2020quantitative, rebuffi2020there}, the top predicted vs.
a randomly chosen class \citep{sixt2020explanations}, or simply between the two classes in binary settings \citep{pope2019explainability}.

Any suitable similarity or distance measure can be used to assess the degree of overlap between \glspl{explanans}, such as the $L_0$ or $L_2$ distance \citep{nie2018theoretical,pope2019explainability}, rank correlation \citep{rebuffi2020there}, or SSIM for image domain \citep{sixt2020explanations}.
A lower similarity implies a clearer distinction between the rationales for each class.

Although the reported implementations target \gls{FA}, the idea of class-discriminative \glspl{explanation} can be naturally extended to other type, wherever \glspl{explanation} can be generated for multiple class hypotheses.
\end{Metric}

\begin{Metric}{}{Output Similarity}{Plausibility}{\gls{ExE}}{1}{\citet{plumb2020explaining}}
\label{met:output_similarity}
    To ensure plausibility, counterfactuals should yield output distribution similar to real instances of the target class.
\citet{plumb2020explaining} evaluate whether each counterfactual $z$ matches the output activation of at least one training sample, i.e.:
 $\exists x' \in \mathcal{X}_{y^*_z} ~:~ \delta\big(\theta(x'), \theta(z)\big) < \epsilon$

This confirms that the counterfactual aligns with typical model behavior for that class.
\end{Metric}

\begin{Metric}{}{Mutual Coherence}{Plausibility, (Fidelity)}{\gls{FA}, (\gls{ExE}, \gls{CE}, \gls{WBS}, \gls{NLE})}{19}{\citet{selvaraju2017grad, guo2018explaining, ancona2019explaining, fernando2019study, fusco2019reconet, jain2019attention, zhang2019towards, marques2020explaining, nguyen2020quantitative, wang2020using, warnecke2020evaluating, graziani2021evaluation, malik2021towards, rajbahadur2021impact, krishna2022disagreement, mercier2022time, jin2023guidelines, duan2024evaluation, tekkesinoglu2024explaining}}
\label{met:mut_coherence}

Several authors propose to evaluate their \glspl{explanation} by comparing them against other XAI methods that are assumed to produce trustworthy or well-understood outputs.
Although only reported for \glspl{FA}, this approach is applicable to any \gls{explanation} type, as long as a suitable reference XAI method is available and an appropriate similarity metric can be defined.

A common reference is the Shapley value framework (see \citet{shapley1953value}), often operationalized via SHAP (see \citet{lundberg2017unified}) due to its solid theoretical grounding.
\glspl{FA} are frequently compared to SHAP estimates using similarity measures \citenote{\citet{ancona2019explaining, zhang2019towards, malik2021towards, jin2023guidelines, tekkesinoglu2024explaining}}.
In the context of saliency maps, Image Occlusion (see \citet{zeiler2013visualizing}) is similarly treated as a proxy ground truth \citep{selvaraju2017grad}.
Comparisons against further \gls{explanation} methods are also reported in literature \citenote{\citet{guo2018explaining, fernando2019study, fusco2019reconet, marques2020explaining, nguyen2020quantitative}}.

Other works compute similarity across multiple \gls{explanation} methods, assuming that high agreement between methods reflects convergence toward a reliable \gls{explanans} \citenote{\citet{jain2019attention, wang2020using, warnecke2020evaluating, graziani2021evaluation, rajbahadur2021impact, krishna2022disagreement, mercier2022time}}.
Any suitable similarity measure may be applied (see \autoref{sec:res_similarity} for an overview).

However, this evaluation approach is not without criticism.
\citet{kumar2020problems} caution that validating one \gls{explanation} method using another may propagate shared biases or assumptions, providing limited evidence for actual correctness.
\end{Metric}

\begin{Metric}{}{Significance Check}{Fidelity}{\gls{FA}, \gls{CE}, (\gls{ExE}, \gls{WBS}, \gls{NLE})}{15}
{\citet{samek2016evaluating, adel2018discovering, chen2018neural, kim2018interpretability, chen2019scalable, deyoung2019eraser, gu2019understanding, wickramanayake2019flex, nam2020relative, hemamou2021multimodal, park2021comparing, pornprasit2021pyexplainer, agarwal2022probing, hameed2022based, bommer2024finding}}
\label{met:significance}
   Statistical significance testing is a common strategy for verifying whether an \gls{explanation} is meaningfully different from random or na\"ive baseline explanations.
This approach is frequently applied to \glspl{FA} \citenote{\citet{samek2016evaluating, chen2018neural, chen2019scalable, deyoung2019eraser, gu2019understanding, wickramanayake2019flex, nam2020relative, hemamou2021multimodal, park2021comparing, pornprasit2021pyexplainer, agarwal2022probing, hameed2022based, bommer2024finding}}.

For \glspl{CE}, statistical tests are typically used to assess whether the extracted concepts carry significantly more information than randomly sampled concepts.
This is done by comparing average relevance or activation scores between the true and random concepts \citep{adel2018discovering, kim2018interpretability}.

In a more advanced variant, \citet{hemamou2021multimodal} train a classifier to distinguish between real and synthetic (random) \glspl{explanation}.
High accuracy in this task indicates that the generated \glspl{explanans} contain statistically meaningful structure.

While commonly reported for \gls{FA} and \gls{CE}, the general idea of statistical significance testing could, in principle, be adapted to other \gls{explanation} types as well.
However, doing so may require \gls{explanation}-specific reformulations.
\end{Metric}

\begin{Metric}{}{(Counter-)Factual Relevance}{Fidelity}{\gls{ExE}}{1}
{\citet{liu2021multi}}
\label{met:CF_relevance}
This metric applies to \glspl{explanation} that generate both a factual and counterfactual \gls{explanans}, aiming to evaluate how well they reflect and contrast the model's reasoning.

Originally proposed by \citet{liu2021multi} for the graph domain, requiring both \glspl{explanans} being subgraphs, it can be generalized to other input types.

The method computes the model's output for the factual and counterfactual \glspl{explanans} and compares them to the original prediction using the negative symmetric Kullback-Leibler divergence.
The difference between these two scores is normalized by the distance between the factual and counterfactual \glspl{explanans}.
A high normalized score indicates that both \gls{explanans} are informative: the factual preserves the original reasoning, while the counterfactual shifts the decision appropriately.

\end{Metric}

\begin{Metric}{}{Prediction Validity}{Fidelity}{\gls{ExE}}{18}{\citet{wachter2017counterfactual, karlsson2018explainable, dhurandhar2019model, guidotti2019factual, mahajan2019preserving, dandl2020multi, le2020grace, molnar2020interpretable, mothilal2020explaining, nguyen2020quantitative, pedapati2020learning, pawelczyk2021carla, rasouli2021analyzing, ma2022clear, tan2022learning, vermeire2022explainable, guidotti2024counterfactual, verma2024counterfactual}}
\label{met:pred_validity}
    By definition, a counterfactual must result in a different prediction from the original input. For untargeted counterfactuals, this simply requires $\hat{y}_z \neq \hat{y}_x$, while targeted counterfactuals must satisfy $\hat{y}_z = y^*_z$ \citep{wachter2017counterfactual, molnar2020interpretable, guidotti2024counterfactual}.

    Most authors assess the fraction of generated counterfactuals that meet this condition \citenote{\citet{karlsson2018explainable, dhurandhar2019model, guidotti2019factual, mahajan2019preserving, le2020grace, mothilal2020explaining, pedapati2020learning, pawelczyk2021carla, ma2022clear, tan2022learning, vermeire2022explainable, verma2024counterfactual}}.
    To capture more nuance, authors also propose measuring the model's confidence in the target class for the counterfactual \citep{rasouli2021analyzing}, or using a continuous loss between the predicted and target class probabilities, such as the $L_1$ distance \citep{dandl2020multi}.
    
    For factual \glspl{explanation}, the class should remain unchanged, i.e., $\hat{y}_z = \hat{y}_x$, which can also be verified either binary \citep{dhurandhar2019model} or using loss-based similarity measures \citep{nguyen2020quantitative}.
\end{Metric}

\begin{Metric}{}{Sufficiency}{Fidelity}{\gls{CE}}{2}{\citet{yeh2020completeness,dasgupta2022framework}}
\label{met:suff}
An \gls{explanation} should be complete enough to serve as a sufficient reason for a given model output.
Ideally, the \gls{explanans} alone should allow us to predict the outcome.
This property reflects the completeness of the \gls{explanans} and can be assessed in different ways.

\citet{yeh2020completeness} train a secondary model that maps the \gls{explanans} back to the black-box's activation space.
The predictive performance using the mapped \gls{explanans} indicates how much information the \gls{explanation} preserves about the original model's decision process.

Alternatively, \citet{dasgupta2022framework} evaluate whether \glspl{explanation} are sufficient for consistent outcomes: Given an \gls{explanans} $e_0$ for an input $x_0$, we identify other instances whose \glspl{explanans} are equivalent (or sufficiently similar) to $e_0$ and compute the fraction that shares the same model prediction as $x_0$.
A higher agreement indicates a more sufficient \gls{explanation}.
\end{Metric}

\begin{Metric}{}{Output Faithfulness}{Fidelity}{\gls{WBS}}{34}{\citet{andrews1995survey, craven1995extracting, stefanowski2001induction, barakat2010intelligible, augasta2012reverse, zilke2016deepred, bastani2017interpreting, krishnan2017palm, lakkaraju2017interpretable, guo2018lemna, laugel2018defining, peake2018explanation, plumb2018model, tan2018distill, wu2018sharing, zhang2018interpreting, chen2019scalable, guidotti2019factual, kanehira2019learning, lakkaraju2019faithful, zhou2019model, anders2020fairwashing, hatwell2020ada, lakkaraju2020robust, panigutti2020doctor, pedapati2020learning, rajapaksha2020lormika, rawal2020beyond, amparore2021trust, chen2021explaining, moradi2021post, pornprasit2021pyexplainer, bayrak2023twin, bo2024incremental}}
\label{met:out_faith}
A surrogate model should closely mimic the behavior of the original black-box model.
Therefore, a common evaluation strategy is to compare the outputs of the surrogate and black-box models using a similarity or performance measure.

This can be assessed with standard measures such as accuracy, $F_1$-score, MSE, or $L_p$ distances \citenote{\citet{andrews1995survey, craven1995extracting, barakat2010intelligible, augasta2012reverse, zilke2016deepred, bastani2017interpreting, krishnan2017palm, lakkaraju2017interpretable, guo2018lemna, laugel2018defining, plumb2018model, tan2018distill, zhang2018interpreting, guidotti2019factual, kanehira2019learning, lakkaraju2019faithful, zhou2019model, anders2020fairwashing, lakkaraju2020robust, panigutti2020doctor, pornprasit2021pyexplainer, bayrak2023twin}},
or with similarity measures such as SSIM, Pearson correlation, or KL divergence \citep{chen2019scalable, anders2020fairwashing}.
Arbitrary loss functions may also be used \citep{amparore2021trust}.

For imbalanced datasets, \citet{moradi2021post} calculate fidelity per class, using either the true labels or the black-box predictions as reference.
For individual rules, fidelity may be expressed as the rule's precision \citep{stefanowski2001induction, hatwell2020ada}.

When evaluating local surrogates, \textit{Local Output Fidelity} is defined by computing fidelity within a neighborhood of the input instance \citep{laugel2018defining, plumb2018model, guidotti2019factual, rajapaksha2020lormika}, or within a synthetic neighborhood \citep{panigutti2020doctor, pornprasit2021pyexplainer}.
For unseen data, \citet{lakkaraju2020robust} validate \glspl{explanation} by comparing outputs to those of the nearest training instances.

\end{Metric}

\begin{Metric}{}{Internal Faithfulness}{\gls{WBS}}{Fidelity}{3}{\citet{messalas2019model, anders2020fairwashing, amparore2021trust}}
\label{met:int_faith}
Since the \gls{WBS} models can achieve similar predictive performance as the original black-box without relying on the same underlying reasoning \citep{messalas2019model, anders2020fairwashing}, it is essential to evaluate their internal fidelity, meaning the similarity in how both models justify their predictions.
This can be achieved by comparing post-hoc \glspl{explanation} of the original and surrogate model for the same inputs. Using feature attribution methods (e.g., SHAP from \citet{lundberg2017unified}), a typical approach is to measure the average overlap of the top-$k$ features between both models' \glspl{explanans} \citep{messalas2019model}.
Other similarity metrics and \gls{explanation} types may also be used.

Alternatively, \citet{amparore2021trust} compare counterfactuals generated from each model, treating their similarity as a proxy for the alignment of decision boundaries. This provides a structural view of how well the surrogate captures the black-box model's rationale beyond mere output agreement.
\end{Metric}

\begin{Metric}{}{Setup Consistency}{Consistency}{\gls{FA}, \gls{ExE}, \gls{WBS}, (\gls{CE}, \gls{NLE}) }{11}{\citet{bastani2017interpreting, honegger2018shedding, guidotti2019factual, rajapaksha2020lormika, warnecke2020evaluating, amparore2021trust, graziani2021evaluation, margot2021new, velmurugan2021evaluating, dai2022fairness, vermeire2022explainable}}
\label{met:setup_consistency}
    This metric evaluates how consistent the \glspl{explanans} remain when generated multiple times for the same input and model.
    This, however, is only a necessary consideration for nondeterministic \gls{explanation} methods.

    Some authors assess this axiomatically, by counting the fraction of identical \glspl{explanans} produced across repeated runs \citep{honegger2018shedding, vermeire2022explainable}. Others calculate distances or similarity scores between different runs and report the aggregated variation, using general or task-specific metrics such as the variance of feature weights or feature presence \citenote{\citet{guidotti2019factual, rajapaksha2020lormika, warnecke2020evaluating, amparore2021trust, graziani2021evaluation, velmurugan2021evaluating, dai2022fairness}}.
    
    Although most work focuses on local instance-wise \glspl{explanation}, the same principle applies to global \glspl{explanation}, where the similarity between globally constructed \glspl{explanans} is measured \citep{bastani2017interpreting, margot2021new}.
    
    While we did not identify examples of this metric applied to every \glspl{explanation} type in the literature, the core idea is general and can easily be extended to any XAI algorithm.
\end{Metric}

\begin{Metric}{}{Hyperparameter Sensitivity}{Consistency}{\gls{FA}, (\gls{ExE}, \gls{CE}, \gls{WBS}, \gls{NLE}) }{6}{\citet{chen2019scalable, verma2019lirme, bansal2020sam, mishra2020reliable, sanchez2020evaluating, graziani2021evaluation}}
\label{met:hyperparam_sens}
This metric evaluates how robust an \gls{explanation} method is to changes in its configuration. Since many XAI methods rely on hyperparameters, high sensitivity may indicate instability, making tuning more difficult and reducing user trust in the method \citep{bansal2020sam}.

A common approach is to generate \glspl{explanans} for the same input across different hyperparameter settings and measure the similarity between them \citenote{\citet{verma2019lirme, bansal2020sam, mishra2020reliable, sanchez2020evaluating, graziani2021evaluation}}.
Alternatively, \citet{chen2019scalable} propose to compare the stability of performance metrics such as fidelity over varying hyperparameters.
Beyond identifying general sensitivity, \citet{graziani2021evaluation} use this technique to guide hyperparameter selection: by progressively adjusting hyperparameters and tracking when the generated \glspl{explanans} converge, one can identify stable regions in the hyperparameter space.

Although most work focuses on \glspl{FA}, the general idea is applicable to any method with tunable hyperparameters.
    
\end{Metric}

\begin{Metric}{}{Execution Time}{Efficiency}{\gls{FA}, \gls{ExE}, \gls{WBS}, (\gls{CE}, \gls{NLE}) }{34}
{\citet{ross2017right, ribeiro2018anchors, zhang2018interpretingnn, chen2019scalable, cheng2019incorporating, fusco2019reconet, ignatiev2019abduction, shakerin2019induction, slack2019assessing, topin2019generation, albini2020relation, guo2020fastif, marques2020explaining, rajapaksha2020lormika, ramon2020comparison, warnecke2020evaluating, abrate2021counterfactual, bajaj2021robust, faber2021comparing, lin2021you, malik2021towards, pawelczyk2021carla, rasouli2021analyzing, van2021interpretable, wang2021probabilistic, amoukou2022accurate, belaid2022we, ma2022clear, mercier2022time, vermeire2022explainable, bayrak2023pertcf, brandt2023precise, jin2023guidelines, verma2024counterfactual}}
\label{met:extime}
    Most authors measure the  time required to generate an \gls{explanans} for a fixed input, model, and dataset, on specified hardware \citenote{\citet{ross2017right, ribeiro2018anchors, zhang2018interpretingnn, cheng2019incorporating, fusco2019reconet, ignatiev2019abduction, shakerin2019induction, guo2020fastif, marques2020explaining, rajapaksha2020lormika, ramon2020comparison, warnecke2020evaluating, bajaj2021robust, faber2021comparing, lin2021you, malik2021towards, pawelczyk2021carla, rasouli2021analyzing, van2021interpretable, wang2021probabilistic, amoukou2022accurate, belaid2022we, ma2022clear, mercier2022time, vermeire2022explainable, bayrak2023pertcf, brandt2023precise, jin2023guidelines, verma2024counterfactual}}.

In addition to empirical runtime, some authors analyze algorithmic complexity using $\mathcal{O}$-notation to characterize the worst-case or average-case number of steps required to run an \gls{explanation} \citenote{\citet{slack2019assessing, topin2019generation, albini2020relation, abrate2021counterfactual, malik2021towards, van2021interpretable}}.
\citet{chen2019scalable} further evaluate parallelizability to understand potential runtime improvements through hardware acceleration or distributed computation.

\end{Metric}

\subsubsection{Input Intervention}
\setMetricPrefix{III}
The metrics listed in this group all require changes to the input, which is then presented to the model again and potentially explained once more.

\begin{Metric}{}{Unguided Perturbation Fidelity}{Fidelity}{\gls{FA}}{10}{\citet{ancona2017towards, shrikumar2017learning, sundararajan2017axiomatic, alvarez2018towards, arya2019one, cheng2019incorporating, yeh2019fidelity, zhang2019interpreting, bhatt2020evaluating, elkhawaga2023evaluating}}
\label{met:unguided_perturb_F}
Some of the most influential XAI metrics evaluate how well an \gls{explanans} aligns with observed model behavior when the input is perturbed. Together, these metrics assess whether the \gls{explanation} faithfully captures how the model responds to its inputs.

This includes \textit{Sensitivity-n} from \citet{ancona2017towards}, \textit{Infidelity} from \citet{yeh2019fidelity}, and \textit{Faithfulness} presented by \citet{alvarez2018towards} and \citet{arya2019one}.
In these approaches, input perturbations are introduced either randomly across all features \citep{yeh2019fidelity}, by zeroing features individually or in small groups \citep{alvarez2018towards, arya2019one, cheng2019incorporating, bhatt2020evaluating}, or by manipulating subsets of fixed size \citep{ancona2017towards}.
The model's change in prediction is then compared to the \gls{explanans}, using different strategies: directly against the raw attribution vector, against a version scaled by perturbation magnitude, or by summing attributions of the changed features.
The deviation is measured using standard metrics like Pearson correlation \citenote{\citet{ancona2017towards, alvarez2018towards, arya2019one, cheng2019incorporating, bhatt2020evaluating}} or mean squared error \citep{yeh2019fidelity}.

Complementing these are completeness-based approaches such as \textit{Completeness} \citep{sundararajan2017axiomatic} and \textit{Summation-to-Delta} \citep{shrikumar2017learning}, which assume that the \gls{explanans} must fully account for the model's behavior.
These compare the difference in output between an instance and a baseline (e.g., a fully perturbed input) with the sum of the attributions across changed features.
Ideally, the two should match exactly.
For practical purposes, however, the relative deviation between attribution sum and output delta can be used as a softer criterion.

%In general this can be expressed as:
% \begin{align*}
%    \mathcal{M}^\text{Cor}_{\text{Sensitivity-}n}(\theta,x,\mathcal{E}) &= \delta\Big(\big[\sum_{j\in \pi} e_{\theta(x,y)}^j\big]_{\pi}~,~\big[\theta(x,y)-\theta(x-\pi,y)\big]_{\pi}   \Big)\\, &\text{perturbations over }n\text{ random features}\\
%    \mathcal{M}^\text{Cor}_\text{Infidelity}(\theta,x,\mathcal{E}) &= \mathbb{E}_\pi\Big[  \delta\big(\pi^T e_{\theta(x,y)}~,~\theta(x,y)-\theta(x-\pi,y) \big)  \Big]\\, & \text{random perturbations  over all features}\\
%    \mathcal{M}^\text{Cor}_\text{Faithfulness}(\theta,x,\mathcal{E}) &=   \delta\Big(e_{\theta(x,y)}~,~ \big[\theta(x,y)-\theta(x-\pi_j,y)\big]_{j=1}^D \Big)\\,&\text{perturbing each feature on its own}\\
%    & \text{where } \delta(\cdot,\cdot) = \text{MSE}(\cdot,\cdot) ~\text{ or } ~-\rho(\cdot, \cdot)
% \end{align*}
    
\end{Metric}

\begin{figure}[ht]
    \centering

    \begin{minipage}{0.49\textwidth}
        \centering
        \includegraphics[width=\linewidth]{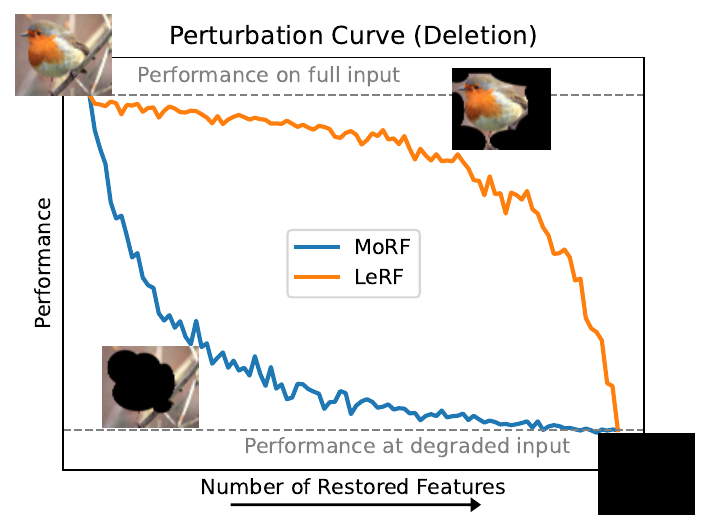} 
        \subcaption{Deletion perturbs features iteratively.}
    \end{minipage}
    \begin{minipage}{0.49\textwidth}
        \centering
        \includegraphics[width=\linewidth]{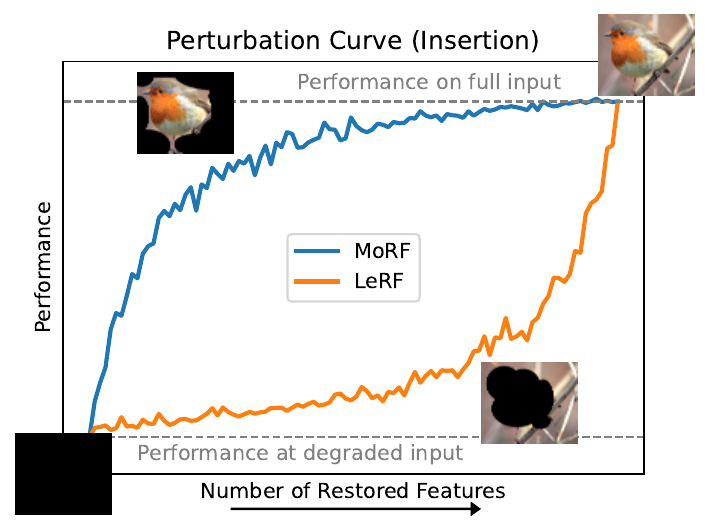} 
        \subcaption{Insertion restores features iteratively.}
    \end{minipage}
    
    \caption{Examples for ideal Perturbation Curves to evaluate feature attributions. Insertion and Deletion approaches are equivalent when switching MoRF for LeRF ordering and vice versa. Deletion MoRF (Insertion LeRF) can be used to evaluate correctness, and Insertion MoRF (Deletion LeRF) to evaluate completeness. }
    \label{fig:iterative_perturbation}
\end{figure}

\begin{Metric}{}{Guided Perturbation Fidelity}{Fidelity}{\gls{FA},\gls{CE}}{75}{\citet{bach2015pixel, samek2016evaluating, ancona2017towards, dabkowski2017real, shrikumar2017learning, chattopadhay2018grad, chu2018exact, guo2018explaining, guo2018lemna, liu2018interpretation, nguyen2018comparing, petsiuk2018rise, wang2018interpret, yang2018towards, annasamy2019towards, arya2019one, brahimi2019deep, deyoung2019eraser, fong2019understanding, kanehira2019multimodal, kapishnikov2019xrai, lin2019explanations, pope2019explainability, schlegel2019towards, serrano2019attention, wagner2019interpretable, yuan2019interpreting, ayush2020generating, brunke2020evaluating, carton2020evaluating, chen2020generating, cong2020exact, fan2020can, hsieh2020evaluations, nguyen2020model, pan2020xgail, rieger2020irof, schlegel2020empirical, schulz2020restricting, singh2020valid, wang2020score, warnecke2020evaluating, el2021towards, ge2021counterfactual, jethani2021have, jung2021towards, liu2021multi, luss2021leveraging, poppi2021revisiting, singh2021extracting, situ2021learning, sun2021preserve, velmurugan2021developing, velmurugan2021evaluating, vu2021c, wang2021probabilistic, yin2021sensitivity, albini2022counterfactual, atanasova2022diagnostics, dai2022fairness, de2022reidentifying, funke2022zorro, hameed2022based, muller2022reshape, ngai2022doctor, rong2022consistent, tan2022learning, veerappa2022validation, vsimic2022perturbation, zou2022ensemble, agarwal2023evaluating, alangari2023intrinsically, jin2023guidelines, schlegel2023deep, awal2024evaluatexai}
}
\label{met:guided_perturb_F}
A common and versatile strategy for evaluating \gls{explanation} quality is to assess how perturbing features based on the \gls{explanans} affects model predictions.
 This approach captures both the necessity (correctness) and sufficiency (completeness) of features highlighted by an \gls{explanans} \citep{carton2020evaluating, deyoung2019eraser, alangari2023intrinsically}.
 Details on perturbation strategies are outlined in \autoref{sec:perturbations}.

\textbf{Perturbation Scope:} Perturbations can be applied either in a fixed-size or iterative manner.
 Fixed-size perturbations change a predefined amount of information at once \citenote{\citet{guo2018lemna, deyoung2019eraser, warnecke2020evaluating, jethani2021have, jung2021towards, velmurugan2021developing, velmurugan2021evaluating, wang2021probabilistic, de2022reidentifying, agarwal2023evaluating}}. 
 The amount of removed features can be selected as a fixed number \citenote{\citet{bach2015pixel, samek2016evaluating, ancona2017towards, chu2018exact, schlegel2019towards, nguyen2020model, alangari2023intrinsically, schlegel2023deep}} or based on a threshold over cumulative importance scores \citenote{\citet{brahimi2019deep, pope2019explainability, schlegel2019towards, ngai2022doctor, schlegel2023deep}}.
  Iterative perturbation gradually removes features one by one or in batches \citenote{\citet{bach2015pixel, samek2016evaluating, ancona2017towards, petsiuk2018rise, deyoung2019eraser, rieger2020irof, jin2023guidelines}}.

\textbf{Perturbation Order:} The order in which features are perturbed is critical. 
Most relevant first (MoRF) evaluates correctness: removing the most important features should lead to a sharp performance drop if the \gls{explanation} correctly highlights necessary features \citenote{\citet{bach2015pixel, samek2016evaluating, chu2018exact, deyoung2019eraser, pope2019explainability, schlegel2019towards, carton2020evaluating, rieger2020irof, ngai2022doctor, alangari2023intrinsically, jin2023guidelines, schlegel2023deep}}.
 Least relevant first (LeRF) tests completeness: performance should remain high if unimportant features are removed first \citenote{\citet{bach2015pixel, ancona2017towards, dabkowski2017real, guo2018lemna, annasamy2019towards, deyoung2019eraser, wagner2019interpretable, carton2020evaluating, singh2020valid, wang2020score, jethani2021have, liu2021multi, singh2021extracting, wang2021probabilistic, dai2022fairness, de2022reidentifying, alangari2023intrinsically}}.
 Alternatively, some authors reverse the entire process by starting from a blank input and adding features incrementally \citenote{\citet{petsiuk2018rise, arya2019one, kapishnikov2019xrai, luss2021leveraging, situ2021learning, sun2021preserve, atanasova2022diagnostics, funke2022zorro, tan2022learning}}, which is functionally equivalent, with MoRF addition corresponding to LeRF deletion, and vice-versa (for an illustration see \autoref{fig:iterative_perturbation}).

\textbf{Measure Score:} The performance change is measured using various scoring mechanisms.
 These include raw prediction difference $\theta(x,\hat{y}) - \theta(\dot{x},\hat{y})$ \citenote{\citet{bach2015pixel, chu2018exact, deyoung2019eraser, kapishnikov2019xrai, chen2020generating, cong2020exact, schulz2020restricting, ngai2022doctor}}, normalized or relative variants \citenote{\citet{chattopadhay2018grad, kapishnikov2019xrai, schulz2020restricting, jung2021towards, velmurugan2021developing, velmurugan2021evaluating, ngai2022doctor}}, or evaluation of loss and accuracy metrics \citenote{\citet{bach2015pixel, chu2018exact, kapishnikov2019xrai, lin2019explanations, schlegel2019towards, serrano2019attention, wagner2019interpretable, warnecke2020evaluating, sun2021preserve, wang2021probabilistic, de2022reidentifying, schlegel2023deep}}.
  Some works use statistical distances such as Kullback-Leibler divergence \citep{liu2021multi, agarwal2023evaluating}, Kendall's Tau \citep{singh2020valid, singh2021extracting}, or Pearson correlation \citep{poppi2021revisiting}.

\textbf{Aggregation:} For iterative perturbations, the results can be aggregated as the average performance change \citep{samek2016evaluating, ancona2017towards}, the area over the perturbation curve (AOPC) \citep{samek2016evaluating, brahimi2019deep, deyoung2019eraser, rieger2020irof}, or the area under it (AUPC) \citep{petsiuk2018rise, ngai2022doctor, vsimic2022perturbation, jin2023guidelines}.
The area between MoRF and LeRF (ABPC) may also be computed \citep{nguyen2020model, schulz2020restricting}, optionally with decay weighting to emphasize early changes \citep{vsimic2022perturbation}.

\textbf{Baselines:} To contextualize results, \gls{explanation} can be compared against various baselines, including random feature selections \citenote{\citet{samek2016evaluating, schlegel2019towards, schlegel2023deep, serrano2019attention, jin2023guidelines}}, zero-inputs \citep{schulz2020restricting}, or na\"ive \gls{explanation} such as edge detectors \citep{hooker2019benchmark}.
Advanced setups may compare against models trained on random labels or with inserted irrelevant features to bound the quality of \gls{explanation} \citep{hameed2022based}.

\textbf{Variants:}
\begin{itemize}
    \item Evaluating \glspl{explanans} across all classes per instance, not only the predicted class \citep{pope2019explainability, awal2024evaluatexai}.

    \item Sanity checks verifying whether the perturbation curve behaves monotonically \citep{arya2019one, fong2019understanding, luss2021leveraging}, or whether MoRF always performs worse than LeRF \citep{vsimic2022perturbation}.

   \item  Normalizing the performance drop by the input difference to mitigate distribution shift: $\frac{\delta\big(\theta(x),\theta(\dot{x})\big)}{\delta(x, \dot{x})}$ \citep{ge2021counterfactual, schlegel2023deep}.

    \item Plotting performance over remaining entropy rather than number of perturbed features \citep{kapishnikov2019xrai}.

    \item Training with randomized feature masking to improve model stability, although this may reduce causality fidelity \citep{jethani2021have}.
\end{itemize}

\textbf{Alternatives:} Further, we may change the perspective through alternative formulations:
\begin{itemize}
   \item Counting the minimum number of features needed to change the model's prediction (in MoRF deletion) \citep{nguyen2018comparing} or using differences between class-specific \glspl{explanans} to guide perturbations \citep{shrikumar2017learning}.

    \item Treating perturbed inputs as counterfactuals and applying metrics such as Minimality (see \metref{met:minimality}) \citep{fan2020can, ge2021counterfactual, albini2022counterfactual}.

    \item Replacing fixed perturbations with adversarial optimization over $k$ features to evaluate minimum change necessary for altering predictions \citep{hsieh2020evaluations, vu2021c}.
\end{itemize}

The approach may be extended to \glspl{CE} by either mapping concepts to features (e.g., concept-based saliency maps by \citet{lucieri2020explaining}), or perturbing internal activations at the concept layer \citep{el2021towards}.

\end{Metric}

\begin{Metric}{}{Counterfactuability}{Fidelity}{\gls{WBS}}{1}{\citet{pornprasit2021pyexplainer}}
\label{met:counterfactuability}
    To assess the expressiveness and impact of rule-based \glspl{explanation} (either generated directly or extracted from surrogate models such as trees), we can use them to guide the generation of counterfactual perturbations. This evaluates whether the rules have predictive leverage and reflect real decision logic, rather than being purely descriptive or spurious.

Specifically, given an input instance that is covered by a rule, \citet{pornprasit2021pyexplainer} perturb the input such that the rule conditions are violated.
If the rule truly captures important rationale, breaking it should influence the black-box model's prediction.
This can be quantified in two ways:
\begin{itemize}
    \item By counting the number of perturbed instances that change the predicted class label.

    \item By measuring the aggregate change in predicted class probability before and after perturbation.
\end{itemize}

\end{Metric}

\begin{Metric}{}{Prediction Neighborhood Continuity}{Fidelity, Continuity}{\gls{WBS}}{1}{\citet{lakkaraju2020robust}}
\label{met:pred_neigborhood}
A \gls{WBS} is only useful if its behavior remains consistent with the black-box even under slight perturbations to the input.

To assess this, \citet{lakkaraju2020robust} propose measuring the difference in Output Faithfulness (see \metref{met:out_faith}) between the original and perturbed inputs.
Specifically, for each input a perturbed variant is created (e.g., through noise or adversarial modification).
The metric then compares the agreement between the black-box model and the surrogate model before and after the perturbations
%$$\underset{x\in\mathcal{X}}{\mathbb{E}}\Big[\mathbbm{1}\big[\hat{y}_x = \hat{y}^e_x\big] - \mathbbm{1}\big[\hat{y}_{\dot{x}} = \hat{y}^e_{\dot{x}}\big]\Big]$$

A lower difference indicates a more robust surrogate, as it preserves faithfulness across perturbations.
High differences may signal that the surrogate captures only superficial model behavior or overfits to specific input patterns.
\end{Metric}

\begin{Metric}{}{Quantification of Unexplainable Features}{Continuity, (Fidelity)}{\gls{FA}, (\gls{CE})}{2}{\citet{zhang2019towards, chen2022can}}
\label{met:quant_unexplainable}
To assess the continuity of \glspl{explanation}, \citet{zhang2019towards} perturb the unimportant features of an input based on the \gls{explanans}, and then generate a new one for the perturbed input.
The similarity between the original and perturbed \glspl{explanans} serves as a measure of continuity: a high similarity suggests that irrelevant changes do not affect the \gls{explanation}, indicating robustness.
In addition, it implies a high completeness, as relevant features must have been captured well enough to maintain the model's behavior despite noise.

A special case of this approach is the Attack Capture Rate proposed by \citet{chen2022can}, which applies to \glspl{FA} in NLP (particularly rationale extraction). Here, insertion attacks introduce distractor phrases to the input, which ideally should not influence the rationale. The metric measures how often the inserted tokens appear in the extracted \gls{explanans}.
This variant relies on artificial attacks, which limits its generalizability beyond NLP.

This metric may be extended to \gls{CE} by perturbing the concept layer corresponding to unimportant concepts, or by mapping concepts back to input features via concept-based localization (e.g., see \citet{lucieri2020explaining}).
\end{Metric}

\begin{Metric}{}{Neighborhood Continuity}{Continuity}{\gls{FA}, \gls{ExE}, \gls{WBS}, \gls{NLE}, (\gls{CE}) }{19}{\citet{alvarez2018robustness, alvarez2018towards, chu2018exact, honegger2018shedding, yeh2019fidelity, zhang2019towards, artelt2020convex, fan2020can, lakkaraju2020robust, artelt2021evaluating, bajaj2021robust, situ2021learning, yin2021sensitivity, agarwal2022rethinking, atanasova2022diagnostics, fouladgar2022metrics, agarwal2023evaluating, bayrak2023pertcf, tekkesinoglu2024explaining}}
\label{met:neighborhood_continutity}
Neighborhood Continuity evaluates whether \glspl{explanans} remain similar for similar \glspl{explanandum}.
The underlying assumption is that small changes in the input should not lead to disproportionately large differences in the output, thereby increasing user trust through Continuity.

The main distinction between proposed metrics lies in how ``similar instances'' are defined.
Some classic metrics include \textit{Local Lipschitz Stability} from \citet{alvarez2018robustness, alvarez2018towards}, \textit{Sensitivity} from \citet{yeh2019fidelity}, and \textit{RIS/RRS/ROS} from \citet{agarwal2022rethinking}.

The \textbf{choice of neighborhood} depends on the domain and \gls{explanation} use case.
Some approaches define neighborhoods using fixed-radius input distances or $k$-nearest neighbors \citep{chu2018exact, situ2021learning, fouladgar2022metrics, tekkesinoglu2024explaining}, while others restrict similarity to instances sharing the same predicted label \citep{honegger2018shedding, fan2020can}.

To \textbf{generate similar inputs}, perturbation-based strategies are often used (see \autoref{sec:perturbations}).
Perturbations may be bounded in magnitude \citenote{\citet{alvarez2018robustness, alvarez2018towards, yeh2019fidelity, bajaj2021robust, agarwal2022rethinking, atanasova2022diagnostics}} or derived from domain-specific semantics \citep{zhang2019towards, yin2021sensitivity, fouladgar2022metrics}.
Some metrics require that perturbed inputs preserve the original prediction $\hat{y}_x$ \citep{alvarez2018robustness, alvarez2018towards, agarwal2022rethinking}, or maintain similar logits $\theta(x)$ \citep{agarwal2023evaluating}.

\textbf{\Gls{explanans} similarity} is typically calculated using distance or correlation-based metrics (see \autoref{sec:res_similarity}) and may be normalized by input distance \citep{alvarez2018robustness, alvarez2018towards, agarwal2022rethinking}, or based on distances in model activation space \citep{agarwal2022rethinking, agarwal2023evaluating}.

\textbf{Variations}: \citet{fan2020can} compare similarity for nearby vs. distant inputs , while \citet{zhang2019towards} restrict comparisons to unchanged features to reduce noise from perturbations.

Although originally proposed for \glspl{FA}, this concept is broadly transferable to other \gls{explanation} types.
For instance, \citet{lakkaraju2020robust} evaluate \glspl{WBS} by comparing surrogate models trained on perturbed data, using model-specific similarity measures such as coefficient mismatch for linear models.

\end{Metric}

\begin{Metric}{}{Adversarial Input Resilience}{Continuity}{\gls{FA}, (\gls{ExE}, \gls{CE}, \gls{WBS}, \gls{NLE})}{10}{\citet{singh2018hierarchical, wang2018interpret, chen2019robust, dombrowski2019explanations, ghorbani2019interpretation, subramanya2019fooling, boopathy2020proper, kuppa2020black, zhang2020interpretable, huang2023safari}}
\label{met:adversarial_input}
    \Glspl{explanation} can be vulnerable to \gls{aa}, where the goal is to manipulate either the \gls{explanans} \citep{ghorbani2019interpretation} or prediction \citep{singh2018hierarchical, wang2018interpret, subramanya2019fooling}.
More sophisticated approaches aim to manipulate one, while additionally restricting changes to the other.
Two main types exist:
\begin{itemize}
    \item \textbf{\Gls{explanans} manipulation}: The \gls{explanans} is altered while the prediction remains fixed \citep{dombrowski2019explanations, boopathy2020proper, kuppa2020black, huang2023safari}.

    \item \textbf{Prediction manipulation}: The prediction changes while the \gls{explanans} remains similar \citep{zhang2020interpretable, huang2023safari}.
\end{itemize}

Evaluation typically measures the distance between the adversarial output (\gls{explanans} or prediction) and either its original or targeted counterpart.
 The input change is usually bounded. 
 Performance is reported either as aggregated distance metrics \citep{dombrowski2019explanations, ghorbani2019interpretation, zhang2020interpretable, huang2023safari} or as attack success rates based on predefined thresholds \citep{kuppa2020black, zhang2020interpretable, huang2023safari}.

While the reported authors focus on \glspl{FA}, the approach can be easily adapted to other \gls{explanation} types by selecting appropriate similarity measures.
    
\end{Metric}

\subsubsection{Model Intervention}
\setMetricPrefix{IV}
For some categories of metrics, it is necessary to change the underlying model, e.g., through retraining or weight randomization.

\begin{Metric}{}{Model Parameter Randomization Test}{Fidelity}{\gls{FA}, (\gls{ExE}, \gls{CE}, \gls{WBS}, \gls{NLE})}{5}{\citet{adebayo2018sanity, kindermans2019reliability, binder2023shortcomings, bommer2024finding, hedstrom2024sanity}}
\label{met:MPRT}
To verify that \glspl{explanation} are truly reflective of the black-box model's learned reasoning, the model's internal parameters are systematically randomized, and the resulting changes in \glspl{explanans} are analyzed.
The rationale is that if an \gls{explanation} remains unchanged under randomization, it is likely generic and not informative of the model's decision logic \citep{adebayo2018sanity}.

Most commonly, the weights of a neural network are randomized either entirely, by layer, or iteratively in top-down or bottom-up order.
The similarity between the original and randomized \glspl{explanans} is then computed \citep{adebayo2018sanity, hedstrom2024sanity}.
Similarity may be evaluated using SSIM and correlation for heatmaps \citep{adebayo2018sanity, binder2023shortcomings, bommer2024finding}, or any suitable similarity metric (see \autoref{sec:res_similarity}).
\citet{hedstrom2024sanity} additionally average the results across noisy inputs (e.g., via input perturbations) to stabilize local \glspl{explanation}.

For gradient-based methods, \citet{sixt2020explanations} propose inserting a random activation vector at a specific layer, to break the causal connection without modifying the network weights.

A complementary strategy proposed by \citet{kindermans2019reliability} keeps the weights fixed but applies a controlled shift to the input distribution, adapting the input-layer biases such that internal computations and outputs remain unchanged. 
Since the model behavior is invariant by design, any change in the \gls{explanans} indicates unwanted sensitivity.
This shift-invariance test can be quantified using standard similarity metrics between pre- and post-shift \glspl{explanans}.

Although originally proposed for \glspl{FA}, the approach is applicable to any \gls{explanation} type for which suitable similarity metrics can be defined.
\end{Metric}

\begin{Metric}{}{Data Randomization Test}{Fidelity}{\gls{FA}, (\gls{ExE}, \gls{CE}, \gls{WBS}, \gls{NLE})}{2}{\citet{adebayo2018sanity, sanchez2020evaluating}}
\label{met:DRT}
\Glspl{explanation} should highlight meaningful structures in the data, not artifacts of memorization.
To verify this, the training data labels are randomized, forcing the model to fit noise rather than learn semantically relevant features \citep{adebayo2018sanity}.
Label randomization can be applied to the full training set \citep{adebayo2018sanity} or a subset only \citep{sanchez2020evaluating}.

This test compares \glspl{explanans} generated by a model trained on the randomized data to those from a model trained on correctly labeled data.
 A strong \gls{explanation} method should yield low similarity between the two, as the latter \glspl{explanans} reflect meaningful decision features, while the former do not.
Similarity can be computed using SSIM, correlation \citep{adebayo2018sanity}, rank-based metrics such as Kendall's Tau \citep{sanchez2020evaluating}, or other suitable measures (see \autoref{sec:res_similarity}).

Although only reported for \glspl{FA}, this approach can be extended to any \gls{explanation} type, provided that a suitable similarity measure between \glspl{explanans} is available.
\end{Metric}

\begin{Metric}{}{Retrained Model Evaluation}{Fidelity}{\gls{FA}}{10}{\citet{guo2019exploring, hooker2019benchmark, cheng2020explaining, han2020explaining, schiller2020relevance, hemamou2021multimodal, shah2021input, li2022face, khalane2023evaluating, raval2023raksha}}
\label{met:roar}
    A key limitation of perturbation-based evaluation methods lies in their potential to introduce distribution shifts though the alteration of features.
To mitigate this, two related strategies retrain models on the perturbed datasets.

The first approach is most famously known as Remove and Retrain (ROAR) by \citet{hooker2019benchmark}, with related variants proposed by \citet{han2020explaining} and \citet{shah2021input}.
It builds on the perturbation strategies of \metref{met:guided_perturb_F}, but ROAR removes features (e.g., the most important ones according to the \gls{explanans}) from the training data and then retrains the model from scratch on the altered dataset.
Performance degradation of this retrained model, compared to the original model, is then used to infer the quality of the \gls{explanation}: a faithful \gls{explanans} should identify features whose removal severely affects performance.

The second strategy leverages \glspl{explanation} for knowledge distillation.
Here, various authors train a new model on a dataset reduced to only the features deemed important by the \gls{explanation} \citenote{\citet{guo2019exploring, cheng2020explaining, schiller2020relevance, hemamou2021multimodal, li2022face, khalane2023evaluating, raval2023raksha}}.

Evaluation follows the general structure of \metref{met:guided_perturb_F}, measuring how much predictive performance is retained when only the relevant features are used.

While both ROAR and distillation-based setups are valuable for benchmarking \gls{explanation} methods in a more robust setting, they do not directly assess the original \gls{explanans} for a specific instance.
Instead, they evaluate the utility of the \gls{explanation} method by measuring the average informativeness of the features it selects.
    
\end{Metric}

\begin{Metric}{}{Influence Fidelity}{Fidelity}{\gls{ExE}}{2}{\citet{krishnan2017palm, guo2020fastif}}
\label{met:influence}
To evaluate the fidelity of \glspl{ExE}, the model is retrained on a dataset modified according to the \gls{explanation}.
The effect of these modifications on model predictions is used to assess the explanatory quality.

\citet{guo2020fastif} remove the identified influential training points from the dataset, retrain the model, and measure the change in loss for the \gls{explanandum}.
 If the removed instances were truly helpful, model performance should degrade.
 Conversely, if they were misleading, removal should lead to an improvement. 
 Thus, a greater change in loss signals a more correct and complete set of influential instances.

In an alternative approach,  \citet{krishnan2017palm} retain the identified influential instances but randomly flips the labels of all remaining training examples before retraining.
If the \gls{explanation} captures all relevant information, the prediction should remain stable.
 Hence, lower prediction variance after retraining indicates a more complete and accurate \gls{explanans}.

\end{Metric}

\begin{Metric}{}{Normalized Movement Rate}{Continuity}{\gls{FA}}{2}{\citet{salih2022investigating, salih2024characterizing}}
\label{met:NMR}
To assess the robustness of \glspl{FA} in the presence of collinear or redundant features, \citet{salih2022investigating} introduce a metric that evaluates the stability of feature rankings as the most important features are iteratively removed and the model is retrained.
 This procedure mirrors the setup of the Remove and Retrain paradigm (\metref{met:roar}), but shifts the focus from prediction performance to the behavior of the \gls{explanation} itself.

After each retraining step, the \gls{explanans} is recomputed, and the ranks of the remaining features are compared to the previous ones.
 A large shift in ranking indicates that redundant or weakly relevant features have taken over the role of more informative ones, suggesting that the attribution method lacks robustness in the face of redundancy and may not reflect the true rationale behind the model's prediction.

To address this issue, \citet{salih2024characterizing} propose the Modified Informative Position (MIP), which aims to stabilize the interpretation by providing a more resilient ranking structure across iterative retraining steps.

\end{Metric}

\begin{Metric}{}{Adversarial Model Resilience}{Continuity}{\gls{FA}, (\gls{ExE}, \gls{CE}, \gls{WBS}, \gls{NLE})}{4}{\citet{heo2019fooling, pruthi2019learning, viering2019manipulate, dimanov2020you}}
\label{met:adversarial_model}
\Gls{explanation} methods should depend meaningfully on the internal parameters of the black-box model.
However, this sensitivity can be exploited to adversarially manipulate them.
Specifically, small but carefully chosen changes to the model's weights can alter the generated \glspl{explanation} without significantly changing predictions.
This manipulation can serve to obfuscate undesirable behaviors or bias within a model.

To assess the robustness of \gls{explanation} methods against such attacks, various strategies perturb the model parameters and observe the resulting changes in \glspl{explanans}. These attacks can be:
\begin{itemize}
    \item \textit{Targeted}, where the modified model is encouraged to produce \glspl{explanans} similar to a predefined target \citep{heo2019fooling, viering2019manipulate}.

    \item \textit{Untargeted}, where the goal is to produce \glspl{explanans} that differ substantially from the original ones \citep{heo2019fooling, pruthi2019learning, dimanov2020you}.
\end{itemize}
To preserve prediction behavior, constraints may be imposed on the model modification. 
\citet{dimanov2020you} bound the change in prediction scores , but other restrictions are thinkable, such as limiting the weight difference norm to ensure the model remains close to the original.

The success of the manipulation is measured depending on the attack setup. 
For targeted attacks, its the similarity between the manipulated \gls{explanans} and the predefined target  \citep{viering2019manipulate}.
Untargeted attacks are evaluated using the dissimilarity between the manipulated and original \gls{explanans} \citep{dimanov2020you, pruthi2019learning}.

All approaches were originally introduced for \gls{FA}, but by applying suitable similarity measures (see \autoref{sec:res_similarity}), they can be extended to other \gls{explanation} types as well.
\end{Metric}

\subsubsection{A Priori Constrained}
\setMetricPrefix{V}
The final group of metrics is those that have additional requirements to the desideratum, namely the necessity for specific annotations in the dataset or using a specific type of models.

\begin{figure}[ht]
    \centering
    \setlength{\fboxsep}{0pt}
    \begin{minipage}{0.30\textwidth}
        \centering
        \fbox{\includegraphics[width=\linewidth, trim = 0 5 5 0, clip]{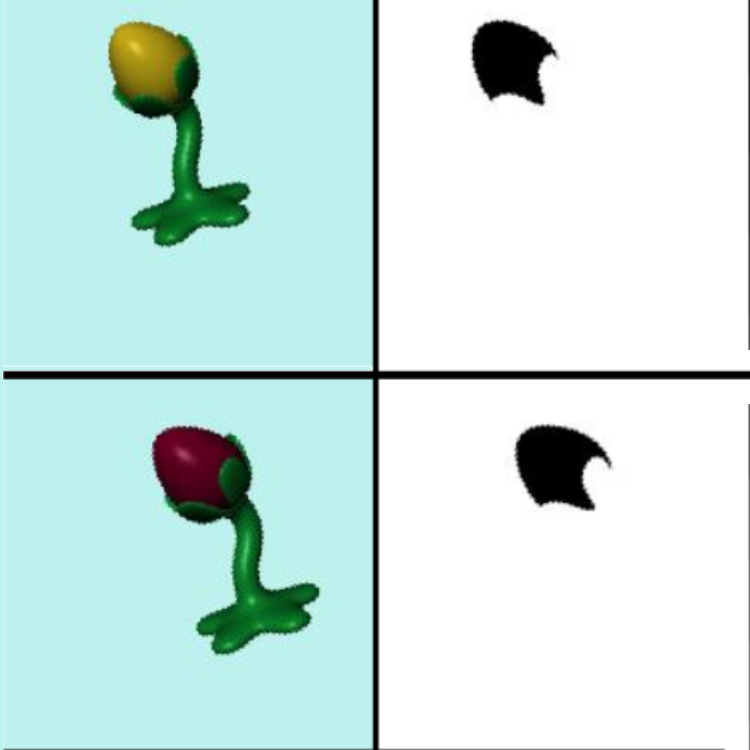}}
        \subcaption{The an8Flower dataset \citep{oramas2017visual}.\\ Left: Input images of flowers, with differing petals. Right: Ground-Truth \gls{explanans} of relevant region.}
    \end{minipage}
    \hspace*{0.5cm}
    \begin{minipage}{0.30\textwidth}
        \centering
        \fbox{\includegraphics[width=\linewidth, trim = -10 0 0 0, clip]{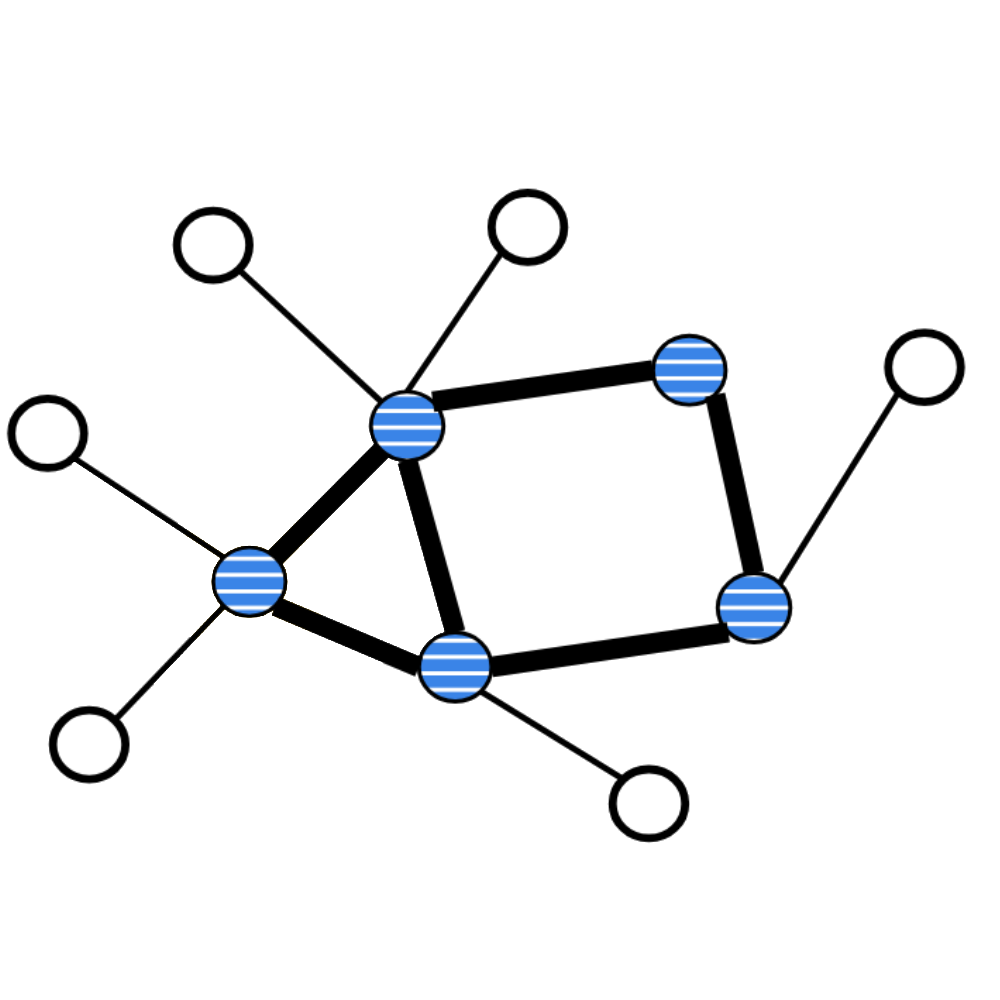}} 
        \subcaption{The ShapeGGen dataset \citep{agarwal2023evaluating}.\\ The graph's label depends on relevant structures or motifs (e.g. house-shape) that serve as ground-truth \gls{explanans}.}
    \end{minipage}
    \hspace*{0.5cm}
    \begin{minipage}{0.30\textwidth}
        \centering
        \fbox{\includegraphics[width=\linewidth, trim = 0 0 0 5, clip]{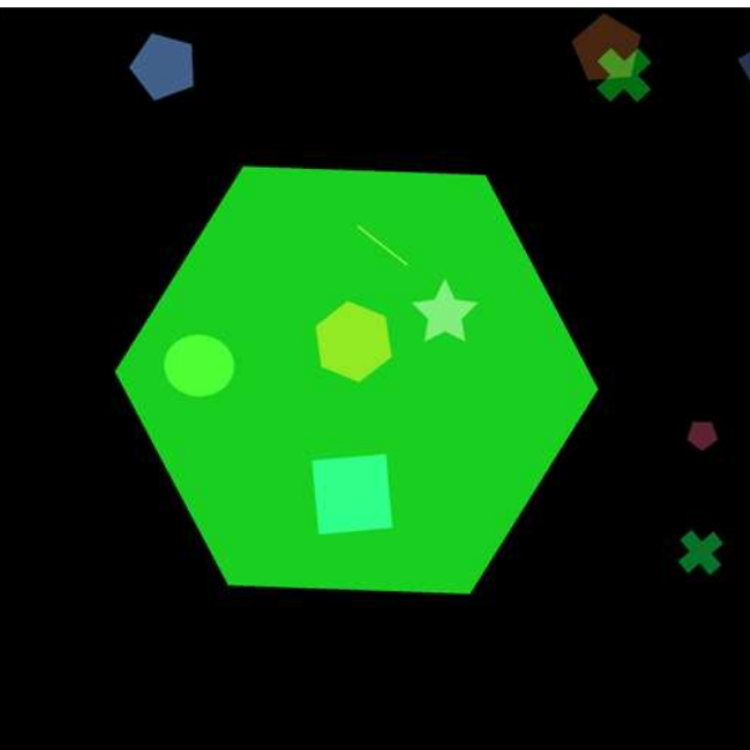}} 
        \subcaption{The SCDB dataset \citep{lucieri2020explaining}.\\ The label is defined by the combination of small objects inside the are of the larger one and serve as ground-truth concepts.}
    \end{minipage}

    \caption{Examples of synthetic datasets containing ground-truth \glspl{explanans} of relevant features.}
    \label{fig:synthetic_datasets}
\end{figure}

\begin{Metric}{nobreak}{GT Dataset Evaluation}{Plausibility, Fidelity}{\gls{FA}, \gls{CE}, \gls{NLE}}{119}%
{\textbf{Human-Annotated Dataset} (69): \citet{simonyan2013deep, cao2015look, lapuschkin2016analyzing, zhou2016learning, das2017human, fong2017interpretable, selvaraju2017grad, bargal2018excitation, baumgartner2018visual, camburu2018snli, chen2018neural, chuang2018learning, jha2018interpretable, liu2018towards, park2018multimodal, poerner2018evaluating, wang2018reinforcement, wu2018faithful, zhang2018top, bastings2019interpretable, chen2019co, chen2019personalized, deyoung2019eraser, fong2019understanding, kanehira2019learning, kanehira2019multimodal, mitsuhara2019embedding, puri2019explain, rajani2019explain, shu2019defend, sydorova2019interpretable, taghanaki2019infomask, trokielewicz2019perception, verma2019lirme, wang2019deliberative, wickramanayake2019flex, zeng2019end, bass2020icam, cheng2020explaining, li2020generate, liu2020towards, nam2020relative, pan2020explainable, rio2020understanding, schulz2020restricting, subramanian2020obtaining, sun2020dual, wang2020scout, xu2020model, xu2020explainable, bany2021eigen, barnett2021interpretable, bykov2021explaining, jang2021training, joshi2021explainable, mathew2021hatexplain, nguyen2021evaluation, wiegreffe2021teach, asokan2022interpretability, cui2022expmrc, du2022care, mucke2022check, theiner2022interpretable, nematzadeh2023ensemble, rasmussen2023machines, ribeiro2023street, tritscher2023evaluating, atanasova2024generating, saifullah2024privacy}
\\
\textbf{Synthetic Dataset} (50): \citet{cortez2013using, chen2017kernel, oramas2017visual, ross2017right, chen2018learning, kim2018interpretability, mascharka2018transparency, yang2018explaining, antwarg2019explaining, arras2019evaluating, camburu2019can, ismail2019input, jia2019improving, lin2019explanations, subramanya2019fooling, takeishi2019shapley, yang2019benchmarking, ying2019gnnexplainer, amiri2020data, ismail2020benchmarking, jia2020exploiting, kohlbrenner2020towards, lucieri2020explaining, luo2020parameterized, nguyen2020quantitative, tritscher2020evaluation, wang2020score, bohle2021convolutional, faber2021comparing, kim2021sanity, lin2021you, liu2021synthetic, shah2021input, yalcin2021evaluating, agarwal2022openxai, amoukou2022accurate, arias2022focus, arras2022clevr, fan2022one, khakzar2022explanations, rao2022towards, tjoa2022quantifying, wilming2022scrutinizing, zhou2022feature, agarwal2023evaluating, hesse2023funnybirds, miro2023novel, sun2023improving, ya2023towards, rao2024better}

}

\label{met:gt_dataset}

To evaluate the quality of \glspl{explanation}, many authors propose comparing them to dataset-based ground truths.
This strategy can follow two distinct paradigms, depending on how the ground truth is derived.
If the rationale is uniquely defined through the data generation process, we may evaluate the \gls{explanation}'s \textbf{Fidelity}. 
If, however, the ground truth stems from human judgment or heuristic annotation, we evaluate the 	\textbf{Plausibility} of the \gls{explanation}.

The former is feasible with synthetic datasets that encode precisely one explanatory rationale. 
The latter is more common in practice but inherently less reliable, as black-box models may learn spurious correlations  as shown by \citet{ribeiro2016should} or identify valid rationales beyond those provided by humans \citep{mucke2022check, ya2023towards}.

\textbf{Synthetic Datasets} are constructed such that the relationship between input features and labels is controlled and known. Ideally, these datasets admit only a single valid rationale. Although this assumption is not always met in practice \citenote{\citet{kim2018interpretability, arras2019evaluating, yang2019benchmarking, tritscher2020evaluation, faber2021comparing}}.
The ground-truth \glspl{explanans} may either be explicitly specified, e.g., by indicating the features responsible for prediction \citenote{\citet{chen2017kernel, oramas2017visual, ross2017right, chen2018learning, yang2018explaining, ying2019gnnexplainer, luo2020parameterized, faber2021comparing, kim2021sanity, tjoa2022quantifying, wilming2022scrutinizing, zhou2022feature}}, or derived from the data-generation process itself, for instance using gradients or Shapley values \citenote{\citet{cortez2013using, jia2019improving, jia2020exploiting, liu2021synthetic, amoukou2022accurate}}.
\autoref{fig:synthetic_datasets} presents selected examples.

Typical strategies for synthetic data include:
\begin{itemize}
    \item Creating inputs from structured primitives (e.g., objects or shapes), where the target labels are deterministic functions of those primitives \citenote{\citet{oramas2017visual, ross2017right, kim2018interpretability, yang2018explaining, ying2019gnnexplainer, lucieri2020explaining, luo2020parameterized, faber2021comparing, kim2021sanity, yalcin2021evaluating, khakzar2022explanations, tjoa2022quantifying, agarwal2023evaluating, hesse2023funnybirds, miro2023novel}}.

    \item Inserting backdoor triggers into inputs, which act as the decisive \gls{explanans} for altered predictions \citenote{\citet{lin2019explanations, lin2021you, fan2022one, sun2023improving, ya2023towards}}.
    \item Performing adversarial attacks restricted to known feature subsets, which then serve as the explanatory evidence for prediction changes \citep{subramanya2019fooling}.

    \item Generating labels as noise-free functions over randomly sampled input features, commonly in tabular domains \citenote{\citet{cortez2013using, chen2017kernel, chen2018learning, ismail2019input, jia2019improving, amiri2020data, ismail2020benchmarking, jia2020exploiting, nguyen2020quantitative, tritscher2020evaluation, liu2021synthetic, agarwal2022openxai, amoukou2022accurate}}.

    \item Constructing mosaic images or input grids where the rationale corresponds to a localized sub-region of the input \citenote{\citet{bohle2021convolutional, shah2021input, arias2022focus, rao2022towards, rao2024better}}.
\end{itemize}

Advantages of synthetic datasets include reduced distribution shift when applying perturbation-based metrics (\metref{met:guided_perturb_F}) \citep{ismail2019input, ismail2020benchmarking, hesse2023funnybirds}, support for concept-level manipulations \citep{yang2019benchmarking, lin2021you, hesse2023funnybirds}, and the inclusion of meaningful counterfactuals or rationale variants \citep{yang2019benchmarking}.

\textbf{Human-Annotated Datasets} instead provide plausible rationales grounded in human intuition or derived from proxy labels.
 In some cases, annotators are explicitly asked to justify their decisions or to highlight which features they consider relevant for a particular prediction \citep{chen2019personalized, xu2020explainable, cui2022expmrc, tritscher2023evaluating}.
  In other settings, rationales are implicit, where existing annotations such as segmentation maps, bounding boxes, or other metadata are repurposed to approximate human reasoning \citenote{\citet{selvaraju2017grad, kanehira2019multimodal, wang2019deliberative, cheng2020explaining, saifullah2024privacy}}.
   These datasets are typically used to evaluate the plausibility of \glspl{explanation} rather than their fidelity, as the ground-truth provided reflects human expectations rather than the actual decision-making logic of the model.

\textbf{Evaluation}:\\
 Across both paradigms, the central assumption is that accurate model predictions imply alignment with the intended rationale, thereby justifying a comparison between the generated \glspl{explanans} and the ground-truth annotations.
 This comparison can be performed using any suitable similarity metric (see \autoref{sec:res_similarity}). 
 While \gls{FA} evaluations often rely on feature-level scoring, \glspl{CE} and \glspl{NLE} require domain-appropriate measures.

In the context of \textbf{\glspl{FA}}, several dedicated evaluation strategies were reported.
 Precision and recall can be computed over truly important features to assess how well the \gls{explanation} captures relevant inputs \citep{bastings2019interpretable}.
  Other techniques evaluate the ranking of important and unimportant features \citep{chen2017kernel, chen2018learning, antwarg2019explaining, camburu2019can}, or quantify the symmetric difference between the sets of selected and annotated features \citep{nguyen2020quantitative}.
   A particularly common metric involves summing the attribution values over known important features, often with normalization or weighting adjustments \citenote{\citet{lapuschkin2016analyzing, yang2019benchmarking, kohlbrenner2020towards, nam2020relative, rio2020understanding, wang2020score, xu2020model, bohle2021convolutional, kim2021sanity, arias2022focus, arras2022clevr, zhou2022feature}}.

For \textbf{visual and natural language domains}, the well-known \textit{Pointing Game} checks whether the most highly attributed features lie within predefined ground-truth regions such as bounding boxes or key tokens \citenote{\citet{bargal2018excitation, poerner2018evaluating, zhang2018top, fong2019understanding, sydorova2019interpretable, taghanaki2019infomask, takeishi2019shapley, schulz2020restricting, barnett2021interpretable, arras2022clevr, theiner2022interpretable}}.
 Closely related is \textit{Weakly Supervised Localization}, where the \gls{explanans} is compared directly to segmentation masks or bounding boxes, typically using Intersection-over-Union (IoU) \citenote{\citet{simonyan2013deep, cao2015look, zhou2016learning, fong2017interpretable, selvaraju2017grad, wickramanayake2019flex, bany2021eigen, nguyen2021evaluation, fan2022one}}. 
 In multi-label or multi-object scenarios, further alignment can be tested by verifying whether the \gls{explanation} focuses on the input features associated with the predicted label \citep{du2019attribution}.

For \textbf{\glspl{CE}}, evaluation typically involves test datasets that explicitly annotate the presence or absence of each concept, enabling precise assessment of concept identification accuracy \citep{kim2018interpretability, lucieri2020explaining, asokan2022interpretability}.

For \textbf{\glspl{NLE}}, comparisons to reference justifications rely on general natural language processing measures such as BLEU or ROUGE \citenote{\citet{camburu2018snli, chuang2018learning, liu2018towards, wu2018faithful, chen2019co, rajani2019explain, wickramanayake2019flex, li2020generate, sun2020dual, jang2021training, wiegreffe2021teach, ribeiro2023street, atanasova2024generating}}, as well as similarity measures specifically designed for \glspl{NLE} \citep{park2018multimodal, du2022care} 

Finally, to ensure that the evaluation isolates \gls{explanation} quality from predictive accuracy,  \citet{fan2022one} recommend to limit evaluation to those samples for which the model's prediction is both correct and sufficiently confident.
\end{Metric}

\begin{Metric}{}{White Box Model Check}{Fidelity}{\gls{FA}, (\gls{ExE}, \gls{NLE})}{12}{\citet{ribeiro2016should, antwarg2019explaining, dhurandhar2019model, jia2019improving, zhou2019model, crabbe2020learning, jia2020exploiting, guidotti2021evaluating, velmurugan2021developing, dai2022fairness, brandt2023precise, carmichael2023well}}
\label{met:WBM}
In white-box models, the internal logic is fully accessible and interpretable, providing a ground-truth rationale against which generated \glspl{explanans} can be directly evaluated.
Typical white-box models used for comparison include linear regressors \citep{crabbe2020learning, dai2022fairness}, feature-additive models \citep{carmichael2023well}, small neural networks with manually set parameters  \citep{antwarg2019explaining, brandt2023precise}, or symbolic models such as decision trees \citep{ribeiro2016should, dhurandhar2019model}.

Given that the model's reasoning is known, \gls{explanation} methods can be assessed by comparing their output against this ground-truth \gls{explanans}.
This can be done using general similarity metrics \citep{guidotti2021evaluating, jia2019improving} or using error metrics like MSE \citep{crabbe2020learning, dai2022fairness, brandt2023precise}.

Further, if the white-box model relies only on a constrained feature subset, one can also measure \gls{explanation} fidelity through accuracy, precision, or recall between \gls{explanans} and the truly influential features \citep{ribeiro2016should, zhou2019model, jia2020exploiting, velmurugan2021developing}.

Although originally proposed for \glspl{FA}, this approach may be extended to \glspl{ExE} and \glspl{NLE}, as the white-box model enables verification of whether the provided \glspl{explanans} are consistent with the known reasoning.
While user inspection is straightforward, adapting this check into an automatic, functionality-grounded evaluation remains challenging but potentially feasible.
\end{Metric}

\renewcommand{\thesubsubsection}{\thesubsection.\arabic{subsubsection}}

\section{Excluded Metrics}
\label{app:excluded_metrics}

We excluded a total of four metrics from our framework. 
While initially identified during our literature review, we later dismissed them after closer inspection.
Some do not evaluate the \gls{explanation} itself, but rather the model or downstream tasks, while others rely on assumptions that are overly narrow or reward properties we consider misleading.
For transparency, we briefly describe each and illustrate our reasons for exclusion below.

\begin{itemize}

    \item \textbf{Alignment}:
    \citet{etmann2019connection} assess plausibility by measuring how well a saliency map aligns with the original input, using correlation as a proxy. 
    However, we argue that input reconstruction does not necessarily enhance interpretability, as simply returning the input or an edge map offers little explanatory value.
    Furthermore, the metric mixes \gls{explanation} with input similarity, rewarding \gls{explanation} that may lack selectivity or meaningful abstraction.
    Therefore, we exclude this metric due to its flawed underlying assumption.

    \item \textbf{Explanatory Power} 
    \citet{arras2017relevant} evaluate the model–\gls{explanation} pair by computing $k$-nearest-neighbor accuracy on document vectors constructed from word-attribution \gls{explanation}. 
     While the score reflects how semantically useful these \glspl{explanation} are for related tasks, it measures downstream utility rather than \gls{explanation} quality. 
     As such, it evaluates a combination of model and \gls{explanation} performance, not the fidelity (or other desideratum) of the \gls{explanation} itself.
     Further, it is not applicable beyond text domain or in scenarios without downstream tasks.

    \item \textbf{Feature Diversity}: 
    \citet{smyth2022few} calculate the fraction of different features altered across all counterfactuals in a dataset, where low diversity implies the same features are consistently changed. 
     However, this appears to be a purely statistical property of the \gls{explanation}, and it remains unclear why higher or lower diversity should indicate better \gls{explanation} quality.
     Without a clear link to any desideratum, its practical value as an evaluation metric is doubtful.

    \item \textbf{Mutual Information}:
    \citet{le2020grace} measure how independent the features changed in counterfactuals are, with the goal of encouraging maximal expressiveness. It uses Symmetrical Uncertainty \citep{press2007numerical} to quantify pairwise dependencies among changed features.
    However, enforcing independence between altered features can easily lead to implausible counterfactuals; especially when high correlations reflect natural dependencies. 
    Breaking these can result in unrealistic or invalid inputs.

\end{itemize}

\newpage
\bibliography{src/references} % 480 references
\bibliographystyle{tmlr}

\end{document}